\documentclass[conference]{IEEEtran}
\usepackage[noadjust]{cite}

\ifCLASSINFOpdf
\else
\fi
\usepackage[cmex10]{amsmath}
\usepackage{amsfonts}

\usepackage{url}

\usepackage{lipsum}
\usepackage[svgnames]{xcolor}
\usepackage{tikz}
\usepackage{amsthm}  

\pgfdeclarelayer{edgelayer}
\pgfdeclarelayer{nodelayer}
\pgfsetlayers{edgelayer,nodelayer,main}

\tikzstyle{none}=[inner sep=0pt]
\tikzstyle{plain}=[inner sep=0pt]
\tikzstyle{every picture}=[baseline=(current bounding box).east,scale=0.3,node distance=5mm]

\newcommand{\ctikzfig}[1]{%
\begin{center}\rm
  
\InputIfFileExists{#1.tikz}{}{\input{.//tikz//#1.tikz}}

\end{center}}

\newcommand{\ket}[1]{|#1\rangle}
\newcommand{\bra}[1]{\langle #1|}
\newcommand{\ten}{\otimes}
\newcommand{\M}{\mathcal}
\newcommand{\Hilb}{\mathbf{FHilb}}
\DeclareMathOperator{\Tr}{Tr}

\newtheorem{definition}{Definition}[section]
\newtheorem{myprop}{Proposition}[section]

\makeatletter
\def\maketag@@@#1{\hbox{\m@th\normalfont\normalsize#1}}
\makeatother

\begin{document}
%

\title{Open System Categorical Quantum Semantics\\in Natural Language Processing}




%
\author{\IEEEauthorblockN{Robin Piedeleu\IEEEauthorrefmark{1},  
Dimitri Kartsaklis\IEEEauthorrefmark{2},
Bob Coecke\IEEEauthorrefmark{1} and
Mehrnoosh Sadrzadeh\IEEEauthorrefmark{2}}
\vspace{0.2cm}\IEEEauthorblockA{\IEEEauthorrefmark{1}University of Oxford, Department of Computer Science\\
Wolfson Building, Parks Road, Oxford OX1 3QD, UK\\
Email: \{robin.piedeleu;bob.coecke\}@cs.ox.ac.uk }
\IEEEauthorblockA{\IEEEauthorrefmark{2}Queen Mary University of London, School of Electronic Engineering and Computer Science\\
Mile End Road, London E1 4NS, UK \\
Email: \{d.kartsaklis;m.sadrzadeh\}@qmul.ac.uk}}  


\maketitle

\begin{abstract}
Originally inspired by categorical quantum mechanics (Abramsky and Coecke, LiCS'04), the categorical compositional distributional model of natural language meaning of Coecke, Sadrzadeh and Clark provides a conceptually motivated procedure to compute the meaning of a sentence, given its grammatical structure within a Lambek pregroup and a vectorial representation of the meaning of its parts. The predictions of this first model have outperformed that of other models in mainstream empirical language processing tasks on large scale data. Moreover, just like CQM allows for varying the model in which we interpret quantum axioms, one can also vary the model in which we interpret word meaning.

In this paper we show that further developments in categorical quantum mechanics are relevant to natural language processing too. Firstly, Selinger's CPM-construction allows for explicitly taking into account lexical ambiguity and distinguishing between the two inherently different notions of homonymy and polysemy. In terms of the model in which we interpret word meaning, this means a passage from the vector space model to density matrices. Despite this change of model, standard empirical methods for comparing meanings can be easily adopted, which we demonstrate by a small-scale experiment on real-world data. This experiment moreover provides preliminary evidence of the validity of our proposed new model for word meaning. 

Secondly, commutative classical structures as well as their non-commutative counterparts that arise in the image of the CPM-construction allow for encoding relative pronouns, verbs and adjectives, and finally, iteration of the CPM-construction, something that has no counterpart in the quantum realm, enables one to accommodate both entailment and ambiguity.
\end{abstract}  


%
\IEEEpeerreviewmaketitle

\section{Introduction}
\label{sec:intro}

Language serves to convey meaning. From this perspective, the ultimate and long-standing goal of any computational linguist is to capture and adequately represent the meaning of an utterance in a computer's memory. At word level, \textit{distributional semantics} offers an effective way to achieve that goal; following the \textit{distributional hypothesis} \cite{harris} which states that the meaning of a word is determined by its context, words are represented as vectors of co-occurrence statistics with all other words in the vocabulary. While models following this paradigm have been found very useful in a number of natural language processing tasks \cite{Schutze,Curran,Landauer}, they do not scale up to the level of phrases or sentences. This is due to the capacity of natural language to generate infinite structures (phases and sentences) from finite means (words); no text corpus, regardless of its size, can provide reliable distributional statistics for a multi-word sentence. On the other hand, type-logical approaches conforming to the tradition of Lambek \cite{lambek1958}, Montague \cite{Mon1} and other pioneers of language, are compositional and deal with the sentence in a more abstract level based on the syntactical rules that hold between the different text constituents, but in principle they do not provide a convincing model for  word meaning.    

The categorical compositional distributional model of Coecke, Sadrzadeh and Clark \cite{Coeckeetal} 
addresses
the challenge of combining these two orthogonal models of meaning in a unified setting. The model is based on the observation that a grammar expressed as a pregroup \cite{Lambek} shares the same structure with the category of finite dimensional vector spaces and linear maps, that of a \textit{compact closed category} \cite{KellyLaplaza}. In principle, this offers a canonical way to express a grammatical derivation as a morphism that defines linear-algebraic manipulations between vector spaces, resulting in a sentence vector. The main characteristic of the model is that the grammatical type of a word determines the vector space in which it lives. Words with atomic types, such as nouns, are represented by vectors living in some basic vector space $N$; on the contrary, relational words such as verbs and adjectives live in tensor product spaces of higher order. An adjective, for example, is an element of $N\ten N$, while a transitive verb lives in $N \ten S \ten N$. The relational tensors act on their argument by \textit{tensor contraction}, a generalization of the familiar notion of matrix multiplication to higher order tensors. 

Ambiguity is a dominant feature of language. At the lexical level, one can distinguish between two broad types of ambiguity: \textit{homonymy} refers to cases in which, due to some historical accident, words that share exactly the same spelling and pronunciation are used to describe completely distinct concepts; such an example is `bank', meaning a financial institution and a land alongside a river. On the other hand, the senses of a \textit{polysemous} word are usually closely related with only small deviations between them; as an example, think of `bank' again as a financial institution and the concrete building where that institution is accommodated. These two notions of ambiguity are inherently different; while a polysemous word still retains a certain level of semantic coherence, a homonymous word can be seen as an incoherent mixing due to coincidence. 
The issue of lexical ambiguity and the different levels of it is currently ignored from almost all attempts that aim to equip distributional models of meaning with compositionality. 




The purpose of this paper is to provide the theoretical foundations for a compositional distributional model of meaning capable of explicitly dealing with lexical ambiguity. At a philosophical level, we define an ambiguous word as a probabilistic mixing of idealistically pure (in the sense of completely unambiguous) concepts. In practice, though, these pure concepts cannot be precisely defined or even expressed by words; no word is completely unambiguous, and its precise meaning can only be defined in relation to a relevant \emph{context} \cite{landauer2001}. Empirically, then, we can approximate these pure concepts by meaning vectors provided by a word sense induction method 
based on clustering the contexts in which a word occurs. We take the set of meaning vectors assigned to a specific word as representing distinct polysemous uses of the word, i.e. relatively self-contained concepts with a certain level of semantic coherence. An ambiguous word then corresponds to a homonymous case, where the same name is used for more than one semantically coherent concepts.

In the proposed model we exploit the observation that the compact closed structure on which the original model of Coecke et al. \cite{Coeckeetal} was based provides an abstraction of the Hilbert space formulation used in the quantum theory, in terms of pure quantum states as vectors, which is known under the umbrella of categorical quantum mechanics \cite{abramsky2004}. In fact, the original model of Coecke et al. was itself greatly inspired by quantum theory, and in particular, by quantum protocols such as quantum teleportation, as explained in \cite{clark2013quantum}. Importantly, vectors in a Hilbert space represent the states of a closed quantum system, also called \textit{pure} states. Selinger's \textit{CPM-construction} \cite{selinger2007dagger}, which maps any dagger compact closed category on another one, then adjoins \textit{open system} states, also called \textit{mixed states}. In the new model, these allow for a lack of knowledge on part of the system under consideration, which may be about an extended part of the quantum system, or uncertainty (read: ambiguity) regarding the preparation procedure. 


The crucial distinction between homonymous and polysemous words is achieved as follows: while a polysemous word corresponds to a \textit{pure} quantum state, a homonymous word is given by a \textit{mixed} state that essentially embodies a probability distribution over all potential meanings of that word. Mathematically, a mixed states is expressed as a \textit{density matrix}: a self-adjoint, positive semi-definite operator with trace one. 
The new formulation offers many opportunities for interesting and novel research. For instance, by exploiting the notion of \textit{Von Neumann entropy} one can measure how ambiguity evolves from individual words to larger text constituents; we would expect that the level of ambiguity in word `bank' is higher than that of the compound `river bank'. 

Furthermore, the richness of the new category in which the meanings of words now live offers interesting alternative design options. In the past, for example, Sadrzadeh, Kartsaklis and colleagues \cite{relpronouns1,kartsaklis2014} enriched the categorical compositional model with elements of classical processing, exploiting the fact that any 
basis of a finite-dimensional vector space induces a \textit{commutative Frobenius algebra} over this space, which allows the uniform copying or deleting of the information relative to this basis \cite{CoeckeVic}. As we will see in Sect. \ref{sec:non-commutative}, the dagger compact closed categories arising from the CPM-construction also accommodate canonical non-commutative Frobenius algebras which have the potential to account for the non-commutativity of language. 

Finally, we  discuss how iterated application of the CPM-construction, which gives rise to states that have no interpretation in quantum theory, does have a natural application  in natural language processing. 
It allows for simultaneous semantic representation of more than one language feature that can be represented by density matrices, for example, lexical entailment in conjunction with ambiguity.  

\vspace{0.2cm}
\textbf{Outline}~~Sect. \ref{sec:background} provides an introduction to categorical compositional distributional semantics; Sect. \ref{sec:understanding} explains the linguistic intuition behind the core ideas of this paper; Sect. \ref{sec:ambiguity} gives the mathematical details for the extension of the original model to the quantum formulation; Sect. \ref{sec:non-commutative} discusses non-commutativity in this new context; finally, Sect. \ref{sec:cpm2} provides the basic intuition for yet another extension of the model that adds the notion of entailment to that of ambiguity. 

    
\vspace{0.2cm}
\textbf{Related work}~~The issue of lexical ambiguity in categorical compositional models of meaning has been previously experimentally investigated by Kartsaklis and Sadrzadeh \cite{kartsaklis:2013:EMNLP}, who present evidence that the introduction of an explicit disambiguation step on the word vectors prior to composition improves the performance of the model in various sentence and phrase similarity tasks.

Furthermore, the research presented here is not the only one that uses density matrices for linguistic purposes. Balk{\i}r \cite{balkir} uses a form of density matrices in order to provide a similarity measure that can be used for evaluating hyponymy-hypernymy relations. In Sect. \ref{sec:cpm2} we indicate how these two uses of density matrices can be merged into one. Blacoe et al. \cite{blacoe-quantum-short} describe a distributional (but not compositional) model of meaning based on density matrices created by grammatical dependencies. At a more generic level not directly associated to density matrices, the application of ideas from quantum theory to language proved to be a very popular field of research---see for example the work of Bruza et al. \cite{bruza2009} and Widdows \cite{widdows2003orthogonal}.   

Finally, the core idea of this paper to represent ambiguous words as mixed states is based on material presented in the MSc thesis of the first author \cite{piedeleu} and the PhD thesis of the second author \cite{kartsaklis-dphil}.

\section{Background}
\label{sec:background}

The field of \textit{category theory} aims at identifying and studying connections between seemingly different forms of mathematical structures. A very representative example of its potency is the compositional categorical framework of Coecke et al. \cite{Coeckeetal}, which shows that a grammatical derivation defining the structure of a sentence is homomorphic to a linear-algebraic formula acting on a semantic space defined by a distributional model. The framework offers a concrete manifestation of the \textit{rule-to-rule hypothesis} \cite{Bach:76}, and a mathematical counterpart to the formal semantics perspective on language. As noted above, the main idea is based on the fact that both the type-logic of the model, a pregroup grammar, and the semantic category, namely $\mathbf{FHilb}$, possess a compact-closed structure. Recall that a \textit{compact closed category} is a monoidal category in which every object $A$ has a left and right adjoint, denoted as $A^l,A^r$ respectively, for which the following special morphisms exist:

\vspace{-0.2cm}
\begin{equation}
\label{equ:eta}
  \eta^l: I \to A \otimes A^l~~~~~~~
  \eta^r:I \to A^r \otimes A
\end{equation}
\vspace{-0.5cm}
\begin{equation}
\label{equ:epsilon}
  \epsilon^l: A^l \otimes A \to I~~~~~~~
  \epsilon^r: A \otimes A^r \to I
\end{equation}

These maps need to satisfy certain conditions (known as \textit{yanking equations}) which ensure that all relevant diagrams commute:

\small
\begin{gather}
\label{equ:yanking}
  (1_A\ten\epsilon^l_A)\circ(\eta^l_A\ten 1_A) = 1_A~~~(\epsilon^r_A\ten 1_A)\circ (1_A\ten \eta^r_A)=1_A  \\ 
  (\epsilon^l_A\ten 1_{A^l})\circ (1_{A^l}\ten \eta^l_A) = 1_{A^l}~~(1_{A^r}\ten\epsilon^r_A)\circ (\eta^r_A\ten 1_{A^r}) = 1_{A^r} \nonumber
\end{gather}
\normalsize

Finally, the passage from syntax to semantics is carried out by a \textit{strong monoidal functor} and, as a result, preserves the compact closed structure. Before we proceed to expand on the above constructions, we briefly introduce the graphical calculus of monoidal categories which will be used throughout our exposition.

\subsection{Graphical calculus}
\label{sec:calculus}

Monoidal categories are complete with regard to a graphical calculus \cite{selinger2011survey} which depicts derivations in their internal language very intuitively, thus simplifying the reading and the analysis. Objects are represented as labelled wires, and morphisms as boxes with input and output wires. The $\eta$- and $\epsilon$-maps are given as half-turns. 

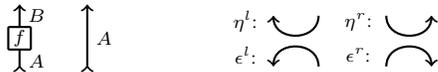
\scriptsize
\ctikzfig{morphism}
\normalsize

Composing morphisms amounts to connecting outputs to inputs, while the tensor product is simply juxtaposition:

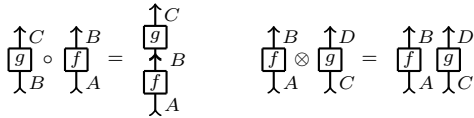
\scriptsize
\ctikzfig{morphism1}
\normalsize

In this language, the yanking equations (\ref{equ:yanking}) get an intuitive visual justification (here for the first two identities):

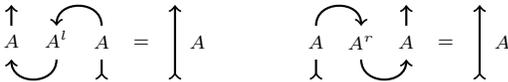
\scriptsize
\ctikzfig{yanking}
\normalsize

For a given object $A$, we define a \textit{state} of $A$ to be a morphism $I\to A$. If $A$ denotes a vector space, we can think of a state as a specific vector living in that space. In our graphical language the unit object $I$ can be omitted, leading to the following representation of states:

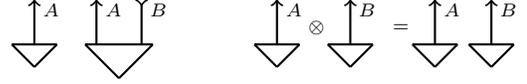
\scriptsize
\ctikzfig{morphism2}
\normalsize

Note that the second diagram from the left depicts an \textit{entangled} state of $A\ten B$; product states (such as the rightmost one) are simple juxtapositions of two states.

\subsection{Pregroup grammars}
\label{sec:pregroups}

A \textit{pregroup algebra}  \cite{Lambek} is a partially ordered monoid with unit 1, whose each element $p$ has a left adjoint $p^l$ and a right adjoint $p^r$, conforming to the following inequalities:

\begin{equation}
 \label{equ:pregroups}
 p^l\cdot p \leq 1 \leq p\cdot  p^l~~~~\text{and}~~~~p\cdot  p^r \leq 1 \leq p^r\cdot  p
\end{equation}

A \textit{pregroup grammar} is a pregroup algebra  freely generated over a set of basic types  ${\cal B}$ including a designated end type and   a type dictionary  that assigns  elements of the pregroup to the vocabulary of a language.  For example, it is usually assumed that ${\cal B} = \{n,s\}$, where $n$ is the type assigned to a noun or a well-formed noun phrase, while $s$ is a designated  type kept for a well-formed sentence. Atomic types can be combined in order to provide types for relational words; for example, an adjective has type $n\cdot n^l$, reflecting the fact that it is something that expects for a noun at its  right-hand side in order to return another noun. Similarly, a transitive verb has type $n^r\cdot s\cdot n^l$, denoting something that expects two nouns (one at each side) in order to return a sentence. Based on (\ref{equ:pregroups}), for this latter case the pregroup derivation gets the following form:

\small
\begin{equation}
  n\cdot (n^r\cdot s \cdot n^l) \cdot n = (n\cdot n^r) \cdot s \cdot (n^l\cdot n) \leq 1\cdot s\cdot 1 \leq s
  \label{equ:derivation}
\end{equation}
\normalsize

Let $\mathbf{C_F}$ denote the \textit{free compact closed category} derived from the pregroup algebra of a pregroup grammar \cite{preller2007free}; then, according to (\ref{equ:eta}) and (\ref{equ:epsilon}), the above type reduction corresponds to the following morphism in $\mathbf{C_F}$:

\begin{equation}
  \epsilon^r_n \cdot 1_s \cdot \epsilon^l_n: n \cdot n^r \cdot s \cdot n^l \cdot n \to s    
\end{equation}

\subsection{From syntax to semantics}

The type-logical approach presented in Sect. \ref{sec:pregroups} is compositional, but unable to distinguish between words of the same type; even more importantly, the only information that a derivation such as the one in (\ref{equ:derivation}) can provide to us is whether the sentence is well-formed or not. Distributional models of meaning offer a solution to the first of these problems, by representing a word in terms of its distributional behaviour in a large corpus of text. While the actual methods for achieving this can vary,\footnote{See Appendix \ref{sec:appendix} for a concrete implementation.} the goal is always the same: to represent words as points of some metric space, where differences in semantic similarity can be detected and precisely quantified. The prime intuition is that words appearing in similar contexts must have a similar meaning \cite{harris}. The word vectors typically live in a highly dimensional semantic space with a fixed orthonormal basis, the elements of which correspond to content-bearing words. The values in the vector of a target word $w_t$ express co-occurrence statistics extracted from some large corpus of text, showing how strongly $w_t$ is associated with each one of the basis words. For a concise introduction to distributional models of meaning see \cite{turney2010}.

We take $(\mathbf{FHilb},\ten)$, the category of finite dimensional Hilbert spaces and linear maps over the scalar field $I$, to be the semantic counterpart of $\mathbf{C_F}$ which, as we saw before, accommodates the grammar. $\mathbf{FHilb}$ is a \textit{dagger compact closed} category (or, $\dagger$-compact closed); that is, a \textit{symmetric} compact closed category (so that $A^r\cong A^l = A^*$ for all $A$) equipped with an involutive contravariant functor $\dagger:\mathbf{FHilb}\to\mathbf{FHilb}$ that is the identity on objects. Concretely, in $\Hilb$, for a morphism $f:A\to B$, its dagger $f^{\dagger}:B \to A$ is simply its adjoint. Furthermore, $\epsilon_A = \eta^{\dagger}_A \circ \sigma_{A^*,A}$ for all $A$. 

Taking $\ket{\psi}$ and $\ket{\phi}$ to be two vectors in a Hilbert space $\mathcal{H}$, $\epsilon_A: A^*\otimes A \rightarrow I$ is the pairing $\epsilon_A(\bra{\psi},\ket{\phi}) = \bra{\psi}(\ket{\phi}) = \bra{\psi}\phi\rangle$ and $\eta_A = \epsilon_A^\dagger$. This yields a categorical definition of the inner product:
\vspace{-0.1cm}
\begin{equation}
  \bra{\psi}\phi\rangle: I \xrightarrow{~\psi~} \mathcal{H} \xrightarrow{~\phi^{\dagger}~} I
  \label{equ:cat-inproduct}
\end{equation}

In practice it is often necessary to normalise in order to obtain the cosine of the angle between vectors as a measure of semantic similarity. This measure has been widely (and successfully, see \cite{Curran}) used in distributional models.

\subsection{Quantizing the grammar}

We now proceed to present a solution to the second problem posed above, that of providing a quantified semantic representation for a sentence by composing the representations of the words therein: in this paper we follow \cite{Preller2010} and \cite{kartsaklis2014} and we achieve the transition from syntax to semantics via a \textit{strong monoidal functor} $Q$:

\begin{equation}
  Q: \mathbf{C_F} \to \mathbf{FHilb}
\end{equation}
\vspace{-0.1cm}

\noindent
which can be shown  to also preserve the compact structure so that $Q(p^l)=Q(p)^l$ and $Q(p^r) = Q(p)^r$ for $p$ an object of $\mathbf{C_F}$. Since each object in $\mathbf{FHilb}$ is its own dual we also have $Q(p^l)\cong Q(p) \cong Q(p^r)$. Moreover, for basic types, we let:

\begin{equation}
  Q(n) = N ~~~~~~ Q(s) = S
\end{equation}

Furthermore, since $Q$ is strongly monoidal, complex types are mapped to tensor product of vector spaces:

\begin{gather}
  Q(n\cdot n^r) = Q(n)\ten Q(n^r) = N\ten N \\
  Q(n^r \cdot s \cdot n^l) = Q(n^r)\ten Q(s) \ten Q(n^l) = N\ten S\ten N \nonumber
\end{gather}

Finally, each morphism in $\mathbf{C_F}$ is mapped to a linear map in $\mathbf{FHilb}$. Equipped with such a functor, we can now define the meaning of a sentence as follows:

\vspace{0.2cm}
\begin{definition}
 Let $\ket{w_i}$ be a vector $I\to Q(p_i)$ corresponding to word $w_i$ with type $p_i$ in a sentence $w_1w_2\hdots w_n$. Given a type-reduction $\alpha: p_1\cdot p_2 \cdot \hdots \cdot p_n \to s$, the meaning of the sentence is defined as:
\begin{equation}
   \ket{w_1w_2\hdots w_n} := Q(\alpha)(\ket{w_1}\ten \hdots \ten \ket{w_n})
 \end{equation}
 \label{def:meaning}
\end{definition}
\vspace{-0.4cm}

Take as an example the sentence ``Trembling shadows play hide-and-seek'', with the standard types $n\cdot n^l$ and $n^r\cdot s\cdot n^l$ assigned to adjectives and verbs, respectively. Then the adjective `trembling' will be a morphism $I\to Q(n\cdot n^l) = I\to N\ten N$, that is, a state in the tensor product space $N\ten N$. Note that this matrix defines a linear map $N\to N$, an interpretation that is fully aligned with the formal semantics perspective: an adjective is a function that takes a noun as input and returns a modified version of it. Similarly, the verb `play'  lives in $N\ten S\ten N$ or, equivalently, is a bi-linear map $N\ten N \to S$ (with a subject and an object as arguments) which returns a sentence. In contrast to those two relational words, the nouns `shadows' and `hide-and-seek' are plain vectors in $N$. The syntax of the sentence conforms to the following type reduction:

\begin{equation}
  \small
  (\epsilon^r_n \cdot 1_s)\circ (1_n \cdot \epsilon^l_n \cdot 1_{n^r} \cdot 1_s \cdot \epsilon^l_n): 
  n \cdot n^l \cdot n \cdot n^r \cdot s \cdot n^l \cdot n \to s 
  \normalsize
\end{equation}

\noindent which, when transferred to $\mathbf{FHilb}$ via $Q$, yields the following diagrammatic derivation:

\begin{equation}
\scriptsize

\begin{tikzpicture}
	\begin{pgfonlayer}{nodelayer}
		\node [style=none, text height=1.5 ex, text depth=0.25 ex] (0) at (2, 3.5) {$S$};
		\node [style=none] (1) at (2, 3) {};
		\node [style=none] (2) at (-5.75, 2.25) {};
		\node [style=none] (3) at (-4.25, 2.25) {};
		\node [style=none] (4) at (-1, 2.25) {};
		\node [style=none] (5) at (1, 2.25) {};
		\node [style=none] (6) at (3, 2.25) {};
		\node [style=none] (7) at (6.25, 2.25) {};
		\node [style=none, text height=1.5 ex, text depth=0.25 ex] (8) at (-5.75, 1.75) {$N$};
		\node [style=none, text height=1.5 ex, text depth=0.25 ex] (9) at (-4.25, 1.75) {$N^l$};
		\node [style=none, text height=1.5 ex, text depth=0.25 ex] (10) at (-1, 1.75) {$N$};
		\node [style=none, text height=1.5 ex, text depth=0.25 ex] (11) at (1, 1.75) {$N^r$};
		\node [style=none, text height=1.5 ex, text depth=0.25 ex] (12) at (3, 1.75) {$N^l$};
		\node [style=none, text height=1.5 ex, text depth=0.25 ex] (13) at (6.25, 1.75) {$N$};
		\node [style=none] (14) at (-5.75, 1.25) {};
		\node [style=none] (15) at (-4.25, 1.25) {};
		\node [style=none] (16) at (-1, 1.25) {};
		\node [style=none] (17) at (1, 1.25) {};
		\node [style=none] (18) at (3, 1.25) {};
		\node [style=none] (19) at (6.25, 1.25) {};
		\node [style=none] (20) at (-6.25, 0.25) {};
		\node [style=none] (21) at (-5.75, 0.25) {};
		\node [style=none] (22) at (-4.25, 0.25) {};
		\node [style=none] (23) at (-3.75, 0.25) {};
		\node [style=none] (24) at (-2, 0.25) {};
		\node [style=none] (25) at (-1, 0.25) {};
		\node [style=none] (26) at (0, 0.25) {};
		\node [style=none] (27) at (0.5, 0.25) {};
		\node [style=none] (28) at (1, 0.25) {};
		\node [style=none] (29) at (2, 0.25) {};
		\node [style=none] (30) at (3, 0.25) {};
		\node [style=none] (31) at (3.5, 0.25) {};
		\node [style=none] (32) at (5.25, 0.25) {};
		\node [style=none] (33) at (6.25, 0.25) {};
		\node [style=none] (34) at (7.25, 0.25) {};
		\node [style=none] (35) at (-1, -0.75) {};
		\node [style=none] (36) at (6.25, -0.75) {};
		\node [style=none] (37) at (-5, -1) {};
		\node [style=none] (38) at (2, -1) {};
		\node [style=none, text height=1.5 ex, text depth=0.25 ex] (39) at (-5, -1.75) {Trembling};
		\node [style=none, text height=1.5 ex, text depth=0.25 ex] (40) at (-1, -1.75) {shadows};
		\node [style=none, text height=1.5 ex, text depth=0.25 ex] (41) at (2, -1.75) {play};
		\node [style=none, text height=1.5 ex, text depth=0.25 ex] (42) at (6.25, -1.75) {hide-and-seek};
	\end{pgfonlayer}
	\begin{pgfonlayer}{edgelayer}
		\draw [thick] (38.center) to (27.center);
		\draw [thick, <-, bend left=90] (6.center) to (7.center);
		\draw [thick] (20.center) to (23.center);
		\draw [thick] (24.center) to (26.center);
		\draw [thick, -<] (28.center) to (17.center);
		\draw [thick, -<] (22.center) to (15.center);
		\draw [thick, ->] (25.center) to (16.center);
		\draw [thick, ->] (29.center) to (1.center);
		\draw [thick, <-, bend left=90] (3.center) to (4.center);
		\draw [thick] (26.center) to (35.center);
		\draw [thick] (32.center) to (34.center);
		\draw [thick] (23.center) to (37.center);
		\draw [thick] (35.center) to (24.center);
		\draw [thick, ->] (33.center) to (19.center);
		\draw [thick, ->] (21.center) to (14.center);
		\draw [thick, -<] (30.center) to (18.center);
		\draw [thick] (34.center) to (36.center);
		\draw [thick] (37.center) to (20.center);
		\draw [thick] (36.center) to (32.center);
		\draw [thick] (31.center) to (38.center);
		\draw [thick] (27.center) to (31.center);
		\draw [thick, ->, bend left=90, looseness=0.75] (2.center) to (5.center);
	\end{pgfonlayer}
\end{tikzpicture}}

\normalsize
\label{equ:shadows}
\end{equation}

Recall that $\epsilon$-morphisms (depicted as caps) in $\mathbf{FHilb}$ denote inner product, so when a relational word of order $m$ is applied on an argument of order $n$, the result is always a tensor of order $n+m-2$; this simply means that the computation in (\ref{equ:shadows}) will be a tensor in $S$, serving as the semantic representation for the sentence.


\subsection{Using Frobenius Algebras in Language}
\label{sec:frobenius}

If distributional models provide a way to build meaning vectors for words with atomic types, the question of how to create a tensor representing a relational or a functional word is much more challenging. Following an approach that resembles an extensional perspective of semantics, Grefenstette and Sadrzadeh \cite{GrefenSadr1} propose the representation of a relational word as the sum of its argument vectors. In other words, an adjective is given as $\sum_i \ket{noun_i}$, where $i$ iterates through all nouns that the specific adjective modifies in a large corpus and $\ket{noun_i}$ is the meaning vector of the $i$th noun; similarly, an intransitive verb is the sum of all its subject nouns, while a transitive verb (a function of two arguments) is defined as follows:

\begin{equation}
  \ket{verb_T} = \sum_i \ket{subj_i}\ten\ket{obj_i}
  \label{equ:tr-verb}
\end{equation}

While this idea is intuitive and fairly easy to implement, it immediately results in a mismatch between the grammatical types of the words and their concrete representations. The type of a transitive verb, for example, is $n^r\cdot s \cdot n^l$, but the method needs a concrete representation which lives in $N\ten N$. In order to provide a solution to this problem, Kartsaklis, Sadrzadeh, Pulman and Coecke \cite{kartsaklis2014} propose to expand this  tensor to $N \otimes N \otimes N$  using the co-monoid part of a  \textit{Frobenius algebra}. Later, Sadrzadeh et al. \cite{relpronouns1,relpronouns2} show how both the monoid and the co-monoid maps of a  Frobenius algebra can be used to model meanings of functional words such as subjective, objective, and possessive relative pronouns.

Recall from \cite{CarboniWal} that a Frobenius algebra in a monoidal category is a quintuple $(A,\Delta,\iota,\mu,\zeta)$ such that:

\begin{itemize}

\item $(A,\mu,\zeta)$ is a monoid, that is we have:

\begin{equation}
  
\begin{tikzpicture}
	\begin{pgfonlayer}{nodelayer}
		\node [style=none] (0) at (-2, 2.25) {};
		\node [draw, circle, minimum size=0.12 cm, fill=white, style=none] (1) at (7.25, 2.25) {};
		\node [style=none] (2) at (-7.5, 1.25) {$\mu=A\otimes A \to A$};
		\node [style=none] (3) at (3.75, 1.25) {$\zeta: I \to A$};
		\node [draw, circle, minimum size=0.12 cm, fill=white, style=none] (4) at (-2, 1) {};
		\node [style=none] (5) at (-3, -0) {};
		\node [style=none] (6) at (-1, -0) {};
		\node [style=none] (7) at (7.25, -0) {};
	\end{pgfonlayer}
	\begin{pgfonlayer}{edgelayer}
		\draw [thick, <-] (0.center) to (4.center);
		\draw [thick, bend left=90, looseness=1.75] (5.center) to (6.center);
		\draw [thick] (7.center) to (1.center);
	\end{pgfonlayer}
\end{tikzpicture}}

\end{equation}

satisfying associativity and unit conditions,

\item $(A,\Delta,\iota)$ is a co-monoid, so that:

\begin{equation}
  
\begin{tikzpicture}
	\begin{pgfonlayer}{nodelayer}
		\node [style=none] (0) at (-3, 2.25) {};
		\node [style=none] (1) at (-1, 2.25) {};
		\node [style=none] (2) at (7.25, 2.25) {};
		\node [style=none] (3) at (-7.5, 1.25) {$\Delta=A \to A \otimes A$};
		\node [draw, circle, minimum size=0.12 cm, fill=white, style=none] (4) at (-2, 1.25) {};
		\node [style=none] (5) at (3.75, 1.25) {$\iota: A \to I$};
		\node [style=none] (6) at (-2, 0) {};
		\node [draw, circle, minimum size=0.12 cm, fill=white, style=none] (7) at (7.25, 0) {};
	\end{pgfonlayer}
	\begin{pgfonlayer}{edgelayer}
		\draw [>-, thick] (6.center) to (4.center);
		\draw [thick] (2.center) to (7.center);
		\draw [thick, bend right=90, looseness=1.75] (0.center) to (1.center);
	\end{pgfonlayer}
\end{tikzpicture}}

\end{equation}

satisfy co-associativity and co-unit conditions; 

\item furthermore, $\Delta$ and $\mu$ adhere to the following \textit{Frobenius condition}:

\begin{equation}
  
\begin{tikzpicture}[baseline=0 pt]
	\begin{pgfonlayer}{nodelayer}
		\node [style=none] (0) at (-4.5, 2) {};
		\node [style=none] (1) at (-2.25, 2) {};
		\node [style=none] (2) at (0.5, 2) {};
		\node [style=none] (3) at (2, 2) {};
		\node [style=none] (4) at (4.75, 2) {};
		\node [style=none] (5) at (7, 2) {};
		\node [draw, style=none, minimum size=0.12 cm, circle, fill=white] (6) at (-2.25, 1.1) {};
		\node [draw, style=none, minimum size=0.12 cm, circle, fill=white] (7) at (1.25, 1.1) {};
		\node [draw, style=none, minimum size=0.12 cm, circle, fill=white] (8) at (4.75, 1.1) {};
		\node [style=none] (9) at (-4.5, 0.25) {};
		\node [style=none] (10) at (-3, 0.25) {};
		\node [style=none] (11) at (-1.5, 0.25) {};
		\node [style=none] (12) at (-0.25, 0.25) {$=$};
		\node [style=none] (13) at (2.75, 0.25) {$=$};
		\node [style=none] (14) at (4, 0.25) {};
		\node [style=none] (15) at (5.5, 0.25) {};
		\node [style=none] (16) at (7, 0.25) {};
		\node [draw, style=none, minimum size=0.12 cm, circle, fill=white] (17) at (-3.75, -0.65) {};
		\node [draw, style=none, minimum size=0.12 cm, circle, fill=white] (18) at (1.25, -0.65) {};
		\node [draw, style=none, minimum size=0.12 cm, circle, fill=white] (19) at (6.25, -0.65) {};
		\node [style=none] (20) at (-3.75, -1.5) {};
		\node [style=none] (21) at (-1.5, -1.5) {};
		\node [style=none] (22) at (0.5, -1.5) {};
		\node [style=none] (23) at (2, -1.5) {};
		\node [style=none] (24) at (4, -1.5) {};
		\node [style=none] (25) at (6.25, -1.5) {};
	\end{pgfonlayer}
	\begin{pgfonlayer}{edgelayer}
		\draw [thick, bend right=90, looseness=2.00] (15.center) to (14.center);
		\draw [thick] (1.center) to (6.center);
		\draw [thick] (0.center) to (9.center);
		\draw [thick, bend left=270, looseness=2.00] (9.center) to (10.center);
		\draw [thick] (19.center) to (25.center);
		\draw [thick] (4.center) to (8.center);
		\draw [thick] (17.center) to (20.center);
		\draw [thick] (14.center) to (24.center);
		\draw [thick, bend left=90, looseness=2.00] (10.center) to (11.center);
		\draw [thick, bend left=90, looseness=2.00] (3.center) to (2.center);
		\draw [thick] (11.center) to (21.center);
		\draw [thick] (7.center) to (18.center);
		\draw [thick] (5.center) to (16.center);
		\draw [thick, bend right=270, looseness=2.00] (16.center) to (15.center);
		\draw [thick, bend left=90, looseness=2.00] (22.center) to (23.center);
	\end{pgfonlayer}
\end{tikzpicture}}

  \label{equ:frobcond}
\end{equation}

\end{itemize}

In a monoidal $\dagger$-category, a $\dagger$-Frobenius algebra is a Frobenius algebra whose co-monoid is adjoint to the monoid. As shown in \cite{CoeckeVic}, every finite dimensional Hilbert space $\mathcal{H}$ with orthonormal basis $\{\ket{i}\}$ has a $\dagger$-Frobenius algebra associated to it, the co-multiplication and multiplication of which corresponds to uniformly \textit{copying} and \textit{uncopying} the basis as follows:

\vspace{-0.5cm}
\begin{gather}
\label{equ:frob}
  \Delta ::\ket{i} \mapsto \ket{i} \ten \ket{i} ~~~~ \iota::\ket{i} \mapsto 1 \\ 
 \small \mu:: \ket{i} \ten \ket{j} \mapsto \delta_{ij}\ket{i} :=  \left\{\begin{array}{c l}
   \ket{i} & i=j \\ 
   \ket{0} & i \neq j
\end{array} \right. ~~~~
  \zeta::1 \mapsto \sum_i \ket{i} \nonumber
\end{gather}

Abstractly, this enables us to copy and delete the (classical) information relative to the given basis. Concretely, the copying $\Delta$-map amounts to encoding faithfully the components of a vector in $\mathcal{H}$ as the diagonal elements of a matrix in $\mathcal{H}\ten \mathcal{H}$, while the ``uncopying'' operation $\mu$ picks out the diagonal elements of a matrix and returns them as a vector in $\mathcal{H}$. Kartsaklis et al. \cite{kartsaklis2014} use the Frobenius co-multiplication in order to faithfully encode tensors of lower order constructed by the argument summing procedure of \cite{GrefenSadr1} to higher order ones, thus restoring the proper functorial relation. The concrete representation of an adjective is given as $\Delta(\sum_i \ket{noun_i})$ which, when substituted to Definition \ref{def:meaning}, gives the composition on the right:\footnote{Notice that the resulting normal form is just a direct application of the Frobenius condition (\ref{equ:frobcond}).}

\begin{equation}
  
\begin{tikzpicture}
	\begin{pgfonlayer}{nodelayer}
		\node [style=none] (0) at (7.5, 3.25) {};
		\node [style=none] (1) at (-0.75, 2.75) {};
		\node [style=none] (2) at (-9, 2.25) {};
		\node [style=none] (3) at (-7, 2.25) {};
		\node [draw, circle, minimum size=0.12 cm, fill=white, style=none] (4) at (7.5, 2.1) {};
		\node [style=none] (5) at (-0.75, 2) {};
		\node [style=none] (6) at (1.25, 2) {};
		\node [style=none] (7) at (2.75, 2) {};
		\node [style=none] (8) at (4.5, 1.5) {$=$};
		\node [draw, circle, minimum size=0.12 cm, fill=white, style=none] (9) at (-8, 1.25) {};
		\node [draw, circle, minimum size=0.12 cm, fill=white, style=none] (10) at (0.25, 1.25) {};
		\node [style=none] (11) at (-10, 1) {{\scriptsize adj:}};
		\node [style=none] (12) at (6.25, 1) {};
		\node [style=none] (13) at (6.25, 1) {};
		\node [style=none] (14) at (8.75, 1) {};
		\node [style=none] (15) at (8.75, 1) {};
		\node [style=none] (16) at (-9, 0.5) {};
		\node [style=none] (17) at (-8, 0.5) {};
		\node [style=none] (18) at (-7, 0.5) {};
		\node [style=none] (19) at (-0.75, 0.5) {};
		\node [style=none] (20) at (0.25, 0.5) {};
		\node [style=none] (21) at (1.25, 0.5) {};
		\node [style=none] (22) at (1.75, 0.5) {};
		\node [style=none] (23) at (2.75, 0.5) {};
		\node [style=none] (24) at (3.75, 0.5) {};
		\node [style=none] (25) at (5.25, 0.5) {};
		\node [style=none] (26) at (6.25, 0.5) {};
		\node [style=none] (27) at (7.25, 0.5) {};
		\node [style=none] (28) at (7.75, 0.5) {};
		\node [style=none] (29) at (8.75, 0.5) {};
		\node [style=none] (30) at (9.75, 0.5) {};
		\node [style=none] (31) at (-8, -0.5) {};
		\node [style=none] (32) at (0.25, -0.5) {};
		\node [style=none] (33) at (2.75, -0.5) {};
		\node [style=none] (34) at (6.25, -0.5) {};
		\node [style=none] (35) at (8.75, -0.5) {};
	\end{pgfonlayer}
	\begin{pgfonlayer}{edgelayer}
		\draw [thick] (19.center) to (32.center);
		\draw [thick] (22.center) to (24.center);
		\draw [thick] (26.center) to (13.center);
		\draw [thick] (24.center) to (33.center);
		\draw [thick, bend left=90, looseness=1.50] (12.center) to (14.center);
		\draw [thick] (16.center) to (31.center);
		\draw [thick] (30.center) to (35.center);
		\draw [thick] (21.center) to (32.center);
		\draw [thick] (22.center) to (33.center);
		\draw [thick, ->] (5.center) to (1.center);
		\draw [thick] (16.center) to (18.center);
		\draw [thick] (20.center) to (10.center);
		\draw [thick] (23.center) to (7.center);
		\draw [thick, <-] (0.center) to (4.center);
		\draw [thick] (25.center) to (34.center);
		\draw [thick] (28.center) to (30.center);
		\draw [thick, bend left=90, looseness=1.75] (6.center) to (7.center);
		\draw [thick, <-<, bend left=270, looseness=1.75] (2.center) to (3.center);
		\draw [thick] (17.center) to (9.center);
		\draw [thick] (28.center) to (35.center);
		\draw [thick] (18.center) to (31.center);
		\draw [thick, bend right=90, looseness=1.25] (5.center) to (6.center);
		\draw [thick] (25.center) to (27.center);
		\draw [thick] (27.center) to (34.center);
		\draw [thick] (29.center) to (15.center);
		\draw [thick] (19.center) to (21.center);
	\end{pgfonlayer}
\end{tikzpicture}}

  \label{equ:frob-adj}
\end{equation}

In order to apply this method to verbs, we first need to notice that since now our sentence space will be essentially produced by copying basis elements of the noun space, our functor $Q$ cannot any more apply different mappings on the two atomic pregroup types $\{s, n\}$; both of these should be mapped onto the same basic vector space, so we get:

\begin{equation}
  Q(n) = W ~~~~~~ Q(s) = W
\end{equation}
 
Given the above limitation, the case of an intransitive verb is quite similar to that of an adjective: we construct a concrete tensor as $\Delta(\sum_i \ket{subj_i})$, the composition of which with a subject noun on its right-hand side proceeds as in (\ref{equ:frob-adj}) due to the commutativity of the algebra. The case of a transitive verb is more interesting, since now the Frobenius structure offers two options: starting from a verb matrix in $W\ten W$ created as in (\ref{equ:tr-verb}), we can encode it to a tensor in $W\ten W\ten W$ by either copying the row dimension (responsible for the interaction of the verb with the subject noun) or the column dimension (responsible for the interaction with the object). For the latter case, referred to by Copy-Object, the composition becomes as follows:

\vspace{-0.5cm}
\begin{equation}
  
\begin{tikzpicture}
	\begin{pgfonlayer}{nodelayer}
		\node [style=none] (0) at (12.5, 3.25) {};
		\node [style=none] (1) at (2, 2.75) {};
		\node [style=none] (2) at (-6.75, 2.25) {};
		\node [style=none] (3) at (-6, 2.25) {};
		\node [style=none] (4) at (-4.5, 2.25) {};
		\node [draw, circle, minimum size=0.12 cm, fill=white, style=none] (5) at (12.5, 2.25) {};
		\node [style=none] (6) at (-0.75, 2) {};
		\node [style=none] (7) at (1.25, 2) {};
		\node [style=none] (8) at (2, 2) {};
		\node [style=none] (9) at (3.5, 2) {};
		\node [style=none] (10) at (5, 2) {};
		\node [draw, circle, minimum size=0.12 cm, fill=white, style=none] (11) at (-5.25, 1.5) {};
		\node [draw, circle, minimum size=0.12 cm, fill=white, style=none] (12) at (2.75, 1.5) {};
		\node [style=none] (13) at (6.5, 1.5) {$=$};
		\node [style=none] (14) at (8, 1.25) {};
		\node [style=none] (15) at (10, 1.25) {};
		\node [style=none] (16) at (11.5, 1.25) {};
		\node [style=none] (17) at (11.5, 1.25) {};
		\node [style=none] (18) at (13.5, 1.25) {};
		\node [style=none] (19) at (13.5, 1.25) {};
		\node [style=none] (20) at (-8.5, 1) {{\scriptsize verb:}};
		\node [style=none] (21) at (-7.25, 0.5) {};
		\node [style=none] (22) at (-6.75, 0.5) {};
		\node [style=none] (23) at (-5.25, 0.5) {};
		\node [style=none] (24) at (-4.75, 0.5) {};
		\node [style=none] (25) at (-1.75, 0.5) {};
		\node [style=none] (26) at (-0.75, 0.5) {};
		\node [style=none] (27) at (0.25, 0.5) {};
		\node [style=none] (28) at (0.75, 0.5) {};
		\node [style=none] (29) at (1.25, 0.5) {};
		\node [style=none] (30) at (2.75, 0.5) {};
		\node [style=none] (31) at (3.25, 0.5) {};
		\node [style=none] (32) at (4, 0.5) {};
		\node [style=none] (33) at (5, 0.5) {};
		\node [style=none] (34) at (6, 0.5) {};
		\node [style=none] (35) at (7, 0.5) {};
		\node [style=none] (36) at (8, 0.5) {};
		\node [style=none] (37) at (9, 0.5) {};
		\node [style=none] (38) at (9.5, 0.5) {};
		\node [style=none] (39) at (10, 0.5) {};
		\node [style=none] (40) at (11.5, 0.5) {};
		\node [style=none] (41) at (12, 0.5) {};
		\node [style=none] (42) at (12.5, 0.5) {};
		\node [style=none] (43) at (13.5, 0.5) {};
		\node [style=none] (44) at (14.5, 0.5) {};
		\node [style=none] (45) at (-0.75, -0.5) {};
		\node [style=none] (46) at (5, -0.5) {};
		\node [style=none] (47) at (8, -0.5) {};
		\node [style=none] (48) at (13.5, -0.5) {};
		\node [style=none] (49) at (-6, -0.75) {};
		\node [style=none] (50) at (2, -0.75) {};
		\node [style=none] (51) at (10.75, -0.75) {};
	\end{pgfonlayer}
	\begin{pgfonlayer}{edgelayer}
		\draw [thick] (36.center) to (14.center);
		\draw [thick, <-] (0.center) to (5.center);
		\draw [thick] (38.center) to (41.center);
		\draw [thick] (23.center) to (11.center);
		\draw [thick] (35.center) to (47.center);
		\draw [thick] (32.center) to (46.center);
		\draw [thick] (30.center) to (12.center);
		\draw [thick, >-] (2.center) to (22.center);
		\draw [thick] (43.center) to (19.center);
		\draw [thick] (15.center) to (39.center);
		\draw [thick] (21.center) to (49.center);
		\draw [thick, bend left=90, looseness=1.75] (14.center) to (15.center);
		\draw [thick] (42.center) to (44.center);
		\draw [thick, <-] (1.center) to (8.center);
		\draw [thick] (31.center) to (50.center);
		\draw [thick] (7.center) to (29.center);
		\draw [thick] (42.center) to (48.center);
		\draw [thick] (37.center) to (47.center);
		\draw [thick, bend right=90, looseness=1.25] (8.center) to (9.center);
		\draw [thick] (24.center) to (49.center);
		\draw [thick, bend left=90, looseness=1.25] (6.center) to (7.center);
		\draw [thick] (25.center) to (45.center);
		\draw [thick] (40.center) to (16.center);
		\draw [thick] (27.center) to (45.center);
		\draw [thick] (28.center) to (31.center);
		\draw [thick] (35.center) to (37.center);
		\draw [thick] (25.center) to (27.center);
		\draw [thick] (41.center) to (51.center);
		\draw [thick] (32.center) to (34.center);
		\draw [thick] (26.center) to (6.center);
		\draw [thick] (38.center) to (51.center);
		\draw [thick] (28.center) to (50.center);
		\draw [thick] (21.center) to (24.center);
		\draw [thick, bend left=90, looseness=1.75] (17.center) to (18.center);
		\draw [thick, bend left=90, looseness=1.75] (9.center) to (10.center);
		\draw [thick] (33.center) to (10.center);
		\draw [thick] (44.center) to (48.center);
		\draw [thick, <-<, bend left=270, looseness=1.75] (3.center) to (4.center);
		\draw [thick] (34.center) to (46.center);
	\end{pgfonlayer}
\end{tikzpicture}}

\end{equation}

\noindent
The composition for the case of copying the subject dimension proceeds similarly on the left-hand side. 
%
In practice,  empirical  work has shown  that objects  have stronger influence on the meaning of a transitive sentence than subjects \cite{kartsaklis2014}, especially when the head verb is ambiguous. This in principle means that the Frobenius structure of the Copy-Object approach is a more effective model of sentential compositionality.

Furthermore, Sadrzadeh et al. \cite{relpronouns1} exploit the  abilities of Frobenius algebras in order to model relative pronouns. Specifically, \textit{copying} is used in conjunction with \textit{deleting} in order to allow the head noun of a relative clause  to interact  with its modifier verb phrase from the far left-hand side of the clause to its right-hand side. For the case of a relative clause modifying a subject this is achieved as follows:

\vspace{-0.5cm}
\begin{equation}
  \scriptsize
  
\begin{tikzpicture}
	\begin{pgfonlayer}{nodelayer}
		\node [style=none] (0) at (13.75, 4.25) {};
		\node [style=none] (1) at (0.5, 3.75) {};
		\node [draw, circle, minimum size=0.12 cm, fill=white, style=none] (2) at (13.75, 3) {};
		\node [style=none] (3) at (-2.75, 2.25) {};
		\node [style=none] (4) at (-0.5, 2.25) {};
		\node [style=none] (5) at (-0.5, 2.25) {};
		\node [style=none] (6) at (0.5, 2.25) {};
		\node [style=none] (7) at (1.5, 2.25) {};
		\node [style=none] (8) at (2.5, 2.25) {};
		\node [style=none] (9) at (4.5, 2.25) {};
		\node [style=none] (10) at (5.5, 2.25) {};
		\node [style=none] (11) at (6.5, 2.25) {};
		\node [style=none] (12) at (8.75, 2.25) {};
		\node [style=none] (13) at (12.5, 2.25) {};
		\node [style=none] (14) at (15, 2.25) {};
		\node [style=none] (15) at (17, 2.25) {};
		\node [style=none] (16) at (19.5, 2.25) {};
		\node [style=none, text height=1.5 ex, text depth=0.25 ex] (17) at (-2.75, 1.75) {\scriptsize{$N$}};
		\node [style=none, text height=1.5 ex, text depth=0.25 ex] (18) at (-0.5, 1.75) {\scriptsize{$N$}};
		\node [style=none, text height=1.5 ex, text depth=0.25 ex] (19) at (0.5, 1.75) {\scriptsize{$N$}};
		\node [style=none, text height=1.5 ex, text depth=0.25 ex] (20) at (1.5, 1.75) {\scriptsize{$S$}};
		\node [style=none, text height=1.5 ex, text depth=0.25 ex] (21) at (2.5, 1.75) {\scriptsize{$N$}};
		\node [style=none, text height=1.5 ex, text depth=0.25 ex] (22) at (4.5, 1.75) {\scriptsize{$N$}};
		\node [style=none, text height=1.5 ex, text depth=0.25 ex] (23) at (5.5, 1.75) {\scriptsize{$S$}};
		\node [style=none, text height=1.5 ex, text depth=0.25 ex] (24) at (6.5, 1.75) {\scriptsize{$N$}};
		\node [style=none, text height=1.5 ex, text depth=0.25 ex] (25) at (8.75, 1.75) {\scriptsize{$N$}};
		\node [style=none, text height=1.5 ex, text depth=0.25 ex] (26) at (12.5, 1.75) {\scriptsize{$N$}};
		\node [style=none, text height=1.5 ex, text depth=0.25 ex] (27) at (15, 1.75) {\scriptsize{$N$}};
		\node [style=none, text height=1.5 ex, text depth=0.25 ex] (28) at (17, 1.75) {\scriptsize{$N$}};
		\node [style=none, text height=1.5 ex, text depth=0.25 ex] (29) at (19.5, 1.75) {\scriptsize{$N$}};
		\node [style=none] (30) at (10.75, 1.5) {$=$};
		\node [style=none] (31) at (-2.75, 1.25) {};
		\node [style=none] (32) at (-0.5, 1.25) {};
		\node [style=none] (33) at (0.5, 1.25) {};
		\node [style=none] (34) at (1.5, 1.25) {};
		\node [style=none] (35) at (2.5, 1.25) {};
		\node [style=none] (36) at (4.5, 1.25) {};
		\node [style=none] (37) at (5.5, 1.25) {};
		\node [style=none] (38) at (6.5, 1.25) {};
		\node [style=none] (39) at (8.75, 1.25) {};
		\node [style=none] (40) at (12.5, 1.25) {};
		\node [style=none] (41) at (15, 1.25) {};
		\node [style=none] (42) at (17, 1.25) {};
		\node [style=none] (43) at (19.5, 1.25) {};
		\node [style=none] (44) at (-3.75, 0.5) {};
		\node [style=none] (45) at (-2.75, 0.5) {};
		\node [style=none] (46) at (-1.75, 0.5) {};
		\node [style=none] (47) at (-0.5, 0.5) {};
		\node [style=none] (48) at (-0.5, 0.5) {};
		\node [style=none] (49) at (-0.5, 0.5) {};
		\node [style=none] (50) at (0.5, 0.5) {};
		\node [style=none] (51) at (0.5, 0.5) {};
		\node [style=none] (52) at (1.5, 0.5) {};
		\node [style=none] (53) at (1.5, 0.5) {};
		\node [style=none] (54) at (2.5, 0.5) {};
		\node [style=none] (55) at (2.5, 0.5) {};
		\node [style=none] (56) at (2.5, 0.5) {};
		\node [style=none] (57) at (3.75, 0.5) {};
		\node [style=none] (58) at (4.5, 0.5) {};
		\node [style=none] (59) at (5.5, 0.5) {};
		\node [style=none] (60) at (6.5, 0.5) {};
		\node [style=none] (61) at (7.25, 0.5) {};
		\node [style=none] (62) at (7.75, 0.5) {};
		\node [style=none] (63) at (8.75, 0.5) {};
		\node [style=none] (64) at (9.75, 0.5) {};
		\node [style=none] (65) at (11.5, 0.5) {};
		\node [style=none] (66) at (12.5, 0.5) {};
		\node [style=none] (67) at (13.5, 0.5) {};
		\node [style=none] (68) at (14.25, 0.5) {};
		\node [style=none] (69) at (15, 0.5) {};
		\node [style=none] (70) at (17, 0.5) {};
		\node [style=none] (71) at (17.75, 0.5) {};
		\node [style=none] (72) at (18.5, 0.5) {};
		\node [style=none] (73) at (19.5, 0.5) {};
		\node [style=none] (74) at (20.5, 0.5) {};
		\node [draw, circle, minimum size=0.12 cm, fill=white, style=none] (75) at (0.5, 0) {};
		\node [draw, circle, minimum size=0.12 cm, fill=white, style=none] (76) at (1.5, 0) {};
		\node [style=none] (77) at (-0.5, -0.25) {};
		\node [style=none] (78) at (0, -0.25) {};
		\node [style=none] (79) at (1, -0.25) {};
		\node [style=none] (80) at (2.5, -0.25) {};
		\node [style=none] (81) at (-2.75, -0.5) {};
		\node [style=none] (82) at (8.75, -0.5) {};
		\node [style=none] (83) at (12.5, -0.5) {};
		\node [style=none] (84) at (19.5, -0.5) {};
		\node [style=none] (85) at (5.5, -1) {};
		\node [style=none] (86) at (16, -1) {};
		\node [style=none, text height=1.5 ex, text depth=0.25 ex] (87) at (-2.75, -1.75) {the~man};
		\node [style=none, text height=1.5 ex, text depth=0.25 ex] (88) at (1, -1.75) {who};
		\node [style=none, text height=1.5 ex, text depth=0.25 ex] (89) at (5.5, -1.75) {likes};
		\node [style=none, text height=1.5 ex, text depth=0.25 ex] (90) at (8.75, -1.75) {Mary};
		\node [style=none, text height=1.5 ex, text depth=0.25 ex] (91) at (12.5, -1.75) {the~man};
		\node [style=none, text height=1.5 ex, text depth=0.25 ex] (92) at (16, -1.75) {likes};
		\node [style=none, text height=1.5 ex, text depth=0.25 ex] (93) at (20, -1.75) {Mary};
	\end{pgfonlayer}
	\begin{pgfonlayer}{edgelayer}
		\draw [thick, looseness=0.00] (75.center) to (51.center);
		\draw [thick, looseness=0.00] (70.center) to (42.center);
		\draw [thick, looseness=0.00] (57.center) to (61.center);
		\draw [thick, looseness=0.00] (65.center) to (83.center);
		\draw [thick, looseness=0.00] (66.center) to (40.center);
		\draw [thick] (77.center) to (47.center);
		\draw [thick, looseness=0.00] (83.center) to (67.center);
		\draw [thick, ->, bend left=90] (8.center) to (9.center);
		\draw [thick, looseness=0.00] (85.center) to (61.center);
		\draw [thick, looseness=0.00] (57.center) to (85.center);
		\draw [thick, looseness=0.00] (47.center) to (48.center);
		\draw [thick, looseness=0.00] (82.center) to (64.center);
		\draw [thick, looseness=0.00] (76.center) to (52.center);
		\draw [thick, ->, bend right=270] (3.center) to (5.center);
		\draw [thick, <-, bend left=90] (15.center) to (16.center);
		\draw [thick, looseness=0.00] (68.center) to (86.center);
		\draw [thick, <-, bend left=90] (7.center) to (10.center);
		\draw [thick, looseness=0.00] (45.center) to (31.center);
		\draw [thick, looseness=0.00] (68.center) to (71.center);
		\draw [thick, looseness=0.00] (44.center) to (46.center);
		\draw [thick, looseness=0.00] (50.center) to (33.center);
		\draw [thick] (80.center) to (54.center);
		\draw [thick, <-, bend left=90] (11.center) to (12.center);
		\draw [thick, looseness=0.00] (54.center) to (56.center);
		\draw [thick, looseness=0.00] (81.center) to (46.center);
		\draw [thick, looseness=0.00] (44.center) to (81.center);
		\draw [thick, looseness=0.00] (65.center) to (67.center);
		\draw [thick, bend left=270] (79.center) to (78.center);
		\draw [thick, looseness=0.00] (72.center) to (84.center);
		\draw [thick, bend left=90] (13.center) to (14.center);
		\draw [thick, bend left=90, looseness=2.75] (78.center) to (77.center);
		\draw [thick, looseness=0.00] (58.center) to (36.center);
		\draw [thick, looseness=0.00] (55.center) to (35.center);
		\draw [thick, ->] (2.center) to (0.center);
		\draw [thick, looseness=0.00] (62.center) to (64.center);
		\draw [thick, looseness=0.00] (73.center) to (43.center);
		\draw [thick, looseness=0.00] (86.center) to (71.center);
		\draw [thick, bend left=90, looseness=1.25] (80.center) to (79.center);
		\draw [thick, looseness=0.00] (53.center) to (34.center);
		\draw [thick, looseness=0.00] (72.center) to (74.center);
		\draw [thick, looseness=0.00] (49.center) to (32.center);
		\draw [thick, looseness=0.00] (59.center) to (37.center);
		\draw [thick, ->, looseness=0.00] (6.center) to (1.center);
		\draw [thick, looseness=0.00] (62.center) to (82.center);
		\draw [thick, looseness=0.00] (84.center) to (74.center);
		\draw [thick, looseness=0.00] (60.center) to (38.center);
		\draw [thick, looseness=0.00] (63.center) to (39.center);
		\draw [thick, looseness=0.00] (69.center) to (41.center);
	\end{pgfonlayer}
\end{tikzpicture}}

  \normalsize
  \label{equ:frob-relpron}
\end{equation}


This concludes the presentation of the categorical compositional model of Coecke et al. \cite{Coeckeetal} and the related research up to today. From the next section we start working towards an extension of this model capable of handling the notions of homonymy and polysemy in a unified manner.

\section{Understanding Ambiguity}
\label{sec:understanding}

In order to deal with lexical ambiguity we firstly need to understand its nature. In other words, we are interested to study in what way an ambiguous word differs from an unambiguous one, and what is the defining quality that makes this distinction clear. On the surface, the answer to these questions seems straightforward: an ambiguous word is one with more than one lexicographic entries in the dictionary. However, this definition fits well only to homonymous cases, in which due to some historical accident words that share the same spelling and pronunciation refer to completely unrelated concepts. Indeed, while the number of meanings of a homonymous word such as `bank' is almost fixed across different dictionaries, the same is not true for the small (and overlapping) variations of senses that might be listed under a word expressing a polysemous case. 

The crucial distinction between homonymy and polysemy is that in the latter case a word still expresses a coherent and self-contained concept. Recall the example of the polysemous use of `bank' as a financial institution and the building where the services of the institution are offered; when we use the sentence `I went to the bank' (with the financial meaning of the word in mind) we essentially refer to \textit{both} of the polysemous meanings of `bank' at the same time---at a higher level, the word `bank' expresses an abstract but concise concept that encompasses all of the available polysemous meanings. On the other hand, the fact that the same name can be used to describe a completely different concept (such as a river bank or a number of objects in a row) is nothing more than an unfortunate coincidence expressing lack of specification. Indeed, a listener of the above sentence can retain a small amount of uncertainty regarding the true intentions of the sayer; although her first guess would be that `bank' refers to the dominant meaning of financial institution (including \textit{all related polysemous meanings}), a small possibility that the sayer has actually visited a river bank still remains. Therefore, in the absence of sufficient context, the meaning of a homonymous word is more reliably expressed as a \textit{probabilistic mixing} of the unrelated individual meanings.

In a distributional model of meaning where a homonymous word is represented by a single vector, the ambiguity in meaning has been collapsed into a convex combination of the relevant sense vectors; the result is a vector that can be seen as the average of all senses, inadequate to reflect the meaning of any of them in a reliable way. We need a way to avoid that. In natural language, ambiguities are resolved with the introduction of context (recall that meaning is use), which means that for a compositional model of meaning the resolving mechanism is the compositional process itself. We would like to retain the ambiguity of a homonymous word when needed (i.e. in the absence of appropriate context) and allow it to collapse only when the context defines the intended sense, during the compositional process. 

In summary, we seek an appropriate model that will allows us: (a) to express homonymous words as probabilistic mixings of their individual meanings; (b) to retain the ambiguity until the presence of sufficient context that will eventually resolve it during composition time; (c) to achieve all the above in the multi-linear setting imposed by the vector space semantics of our original model. 

\section{Encoding Ambiguity}
\label{sec:ambiguity}

The previous compositional model relies on a strong monoidal functor from a compact closed category, representing syntax, to $\Hilb$, modelling a form of distributional semantics. In this section, we will modify the functor to a new codomain category. However, before we start, we establish a few guidelines:

\begin{itemize}
\item our construction needs to retain a compact closed structure in order to carry the grammatical reduction maps to the new category;
\item we wish to be able to compare the meaning of words as in the previous model, i.e., the new category needs to come equipped with a dagger structure that implements this comparison;
\item finally, we need a Frobenius algebra to merge and duplicate information in concrete models.
\end{itemize}

To achieve our goal, we will explore a categorical construction, inspired from quantum physics and originally due to Selinger \cite{selinger2007dagger}, in the context of the categorical model of meaning developed in the previous sections. 

\subsection{Mixing in $\Hilb$}

Although seemingly unrelated, quantum mechanics and linguistics share a common link through the framework of $\dagger$-compact closed categories, an abstraction of the Hilbert space formulation, and have been used in the past \cite{abramsky2004} to provide structural proofs for a class of quantum protocols, essentially recasting the vector space semantics of quantum mechanics in a more abstract way. Shifting the perspective to the field of linguistics, we saw how the same formalism proposes a description of the semantic interactions of words at the sentence level. Here we make the connection between the two fields even more explicit, taking advantage of the fact that the ultimate purpose of quantum mechanics is to deal with ambiguity---and this is exactly what we need to achieve here in the context of language.

We start by observing that, in quantum physics, the Hilbert space model is insufficient to incorporate the epistemic state of the observer in its formalism: what if one does not have knowledge of a quantum system's initial state and can only attribute a probability distribution to a set of possible states? The answer is by considering a \emph{statistical ensemble} of pure states: for example, one may assign a $\frac{1}{2}$ probability that the state vector of a system is $\ket{\psi_1}$ and a $\frac{1}{2}$ probability that it is in state $\ket{\psi_2}$. We say that this system is in a \emph{mixed state}. In the Hilbert space setting, such a state cannot be represented as a vector. In fact, any normalised sum of pure states is again a pure state (by the vector space structure). Note that the state $(\psi_1 + \psi_2)/\sqrt{2}$ is a \textit{quantum superposition} and not the mathematical representation of the mixed state above.

This situation is similar to the issue we face when trying to model ambiguity in distributional semantics: given two different meanings of a homonymous word and their relative weights (given as probabilities), simply looking at the convex composition of the associated vectors collapses the ambiguous meaning to a single vector, thereby fusing together the two senses of the word. This is precisely what was discussed in Sect. \ref{sec:understanding}. The mathematical response to this problem is to move the focus away from states in a Hilbert space to a specific kind of operators on the same space: more specifically, to \emph{density operators}, i.e., positive semi-definite, self-adjoint operators of trace one. The density operator formalism is our means to express a probability distribution over the potential meanings of a homonymous word in a distributional model. We formally define this as follows:

\begin{definition}
\label{def:word}
Let a distributional model be given in the form of a Hilbert space $M$, in which every word $w_t$ is represented by a statistical ensemble $\{(p_i, \ket{w_t^i})\}_i$---where $\ket{w_t^i}$ is a vector corresponding to a specific unambiguous meaning of the word that can occur with probability $p_i$. The distributional meaning of the word is defined as:

\begin{equation}
  \rho(w_t) = \sum_i p_i \ket{w_t^i}\bra{w_t^i}
\end{equation}

\end{definition}

In conceptual terms, mixing is interpreted as ambiguity of meaning: a word $w_t$ with meaning given by $\rho(w_t)$ can have pure meaning $w_t^i$ with probability $p_i$. Note that for the case of a non-homonymous word, the above formula reduces to $\ket{w_t}\bra{w_t}$, with $\ket{w_t}$ corresponding to the state vector assigned to $w_t$. Now, if mixed states are density operators, we need a notion of morphism that preserves this structure, i.e., that maps states to states. In the Hilbert space model, the morphisms were simply linear maps. The corresponding notion in the mixed setting is that of \emph{completely positive map}, that is, positive maps that respect the monoidal structure of the underlying category.





To constitute a compositional model of meaning, our construction also needs to respect our stated goals: specifically, the category of operator spaces and completely positive maps must be a $\dagger$-compact closed category; furthermore, we need to identify the morphism that plays the part of the Frobenius algebra of the previous model. We start working towards these goals by describing a construction that builds a similar category, not only from \textbf{FHilb}, but, more abstractly, from any $\dagger$-compact closed category.

\subsection{Doubling and complete positivity}

The category that we are going to build was originally introduced by Selinger \cite{selinger2007dagger} as a generalisation of the corresponding construction on Hilbert spaces. Conceptually, it corresponds to shifting the focus away from vectors or morphisms of the form $I \rightarrow A$ to operators on the same space or morphisms of type $A \rightarrow A$. We will formalise this idea by first introducing the category $\mathbf{D}(\M{C})$ on a compact closed category $\M{C}$, which can be perhaps better understood in its diagrammatic form as a \textit{doubling} of the wires. In this context, we obtain a duality between states of $\mathbf{D}(\M{C})$ and operators of $\M{C}$, pictured by simple wire manipulations. As we will see, $\mathbf{D}(\M{C})$ retains the compact closedness of $\M{C}$ and is therefore a viable candidate for a semantic category in our compositional model of meaning. However, at this stage, states of $\mathbf{D}(\M{C})$ do not yet admit a clear interpretation in terms of mixing. This is why we need to introduce the notion of completely positive morphisms, of which positive operators on a Hilbert space (mixed states in quantum mechanics) are a special case. This will allow us later to define the subcategory $\mathbf{CPM}(\M{C})$ of $\mathbf{D}(\M{C})$.

\vspace{0.2cm}
\subsubsection{The $\mathbf{D}$ construction---doubling}

First, given a $\dagger$-compact closed category\footnote{The construction works on any monoidal category with a dagger, i.e., an involution, but we will not need the additional generality.} $\M{C}$ we define:

\begin{definition}
The category $\mathbf{D}(\M{C})$ with

\begin{itemize}
\item the same objects as $\M{C}$;
\item morphisms between objects $A$ and $B$ of $\mathbf{D}(\M{C})$ are morphisms $A\otimes A^* \rightarrow B\otimes B^*$ of $\M{C}$.
\item composition and dagger are inherited from $\M{C}$ via the embedding $E: \mathbf{D}(\M{C}) \hookrightarrow \M{C}$ defined by

\begin{equation*}
  \left\{
    \begin{array}{rll}
      A & \mapsto A \otimes A^* & \text{on objects};\\
      f & \mapsto f &  \text{on morphisms}.
    \end{array} \right.
\end{equation*}
\end{itemize}
\end{definition}

In addition, we can endow the category $\mathbf{D}(\M{C})$ of a monoidal structure by defining the tensor $\otimes_\mathbf{D}$ by

\vspace{-0.2cm}
\[A \otimes_\mathbf{D} B = A \otimes B\]

\noindent
on objects $A$ and $B$, and for morphisms $f_1: A \otimes A^* \rightarrow B\otimes B^*$ and $f_2: C \otimes C^* \rightarrow D\otimes D^*$, by:

\vspace{-0.5cm}
\begin{multline}
f_1\otimes_\mathbf{D} f_2: A \otimes C \otimes C^* \otimes A^* \xrightarrow{\cong} A \otimes A^* \otimes C \otimes C^* 
\\ \xrightarrow{f_1\otimes f_2} B \otimes B^* \otimes D \otimes D^* \xrightarrow{\cong} B \otimes D \otimes D^* \otimes B^*
\end{multline}

Or graphically by,

\vspace{-0.3cm}
\begin{equation}
\scriptsize

\begin{tikzpicture}[scale=1.9]
	\begin{pgfonlayer}{nodelayer}
		\node [style=none] (0) at (-7, 1.5) {};
		\node [style=none] (1) at (-5.5, 1.5) {};
		\node [style=none] (2) at (-3, 1.5) {};
		\node [style=none] (3) at (-2, 1.5) {};
		\node [style=none] (4) at (-0.75, 1.5) {};
		\node [style=none] (5) at (0.25, 1.5) {};
		\node [style=none] (6) at (2, 1.5) {};
		\node [style=none] (7) at (3, 1.5) {};
		\node [style=none] (8) at (4, 1.5) {};
		\node [style=none] (9) at (5, 1.5) {};
		\node [style=none] (10) at (2, 0.75) {};
		\node [style=none] (11) at (2.75, 0.75) {};
		\node [style=none] (12) at (3, 0.75) {};
		\node [style=none] (13) at (4, 0.75) {};
		\node [style=none] (14) at (4.25, 0.75) {};
		\node [style=none] (15) at (5, 0.75) {};
		\node [style=none] (16) at (3.5, 0.5) {$f_2$};
		\node [style=none] (17) at (-7.5, 0.25) {};
		\node [style=none] (18) at (-7, 0.25) {};
		\node [style=none] (19) at (-6.5, 0.25) {};
		\node [style=none] (20) at (-6, 0.25) {};
		\node [style=none] (21) at (-5.5, 0.25) {};
		\node [style=none] (22) at (-5, 0.25) {};
		\node [style=none] (23) at (-3.25, 0.25) {};
		\node [style=none] (24) at (-3, 0.25) {};
		\node [style=none] (25) at (-2, 0.25) {};
		\node [style=none] (26) at (-1.75, 0.25) {};
		\node [style=none] (27) at (-1, 0.25) {};
		\node [style=none] (28) at (-0.75, 0.25) {};
		\node [style=none] (29) at (0.25, 0.25) {};
		\node [style=none] (30) at (0.5, 0.25) {};
		\node [style=none] (31) at (2, 0.25) {};
		\node [style=none] (32) at (2.75, 0.25) {};
		\node [style=none] (33) at (3, 0.25) {};
		\node [style=none] (34) at (4, 0.25) {};
		\node [style=none] (35) at (4.25, 0.25) {};
		\node [style=none] (36) at (5, 0.25) {};
		\node [style=none] (37) at (-7, 0) {$f_1$};
		\node [style=none] (38) at (-5.5, 0) {$f_2$};
		\node [style=none] (39) at (-4.25, 0) {$\mapsto$};
		\node [style=none] (40) at (-2.5, 0) {$f_1$};
		\node [style=none] (41) at (-0.25, 0) {$f_2$};
		\node [style=none] (42) at (1.25, 0) {$=$};
		\node [style=none] (43) at (-7.5, -0.25) {};
		\node [style=none] (44) at (-7, -0.25) {};
		\node [style=none] (45) at (-6.5, -0.25) {};
		\node [style=none] (46) at (-6, -0.25) {};
		\node [style=none] (47) at (-5.5, -0.25) {};
		\node [style=none] (48) at (-5, -0.25) {};
		\node [style=none] (49) at (-3.25, -0.25) {};
		\node [style=none] (50) at (-3, -0.25) {};
		\node [style=none] (51) at (-2, -0.25) {};
		\node [style=none] (52) at (-1.75, -0.25) {};
		\node [style=none] (53) at (-1, -0.25) {};
		\node [style=none] (54) at (-0.75, -0.25) {};
		\node [style=none] (55) at (0.25, -0.25) {};
		\node [style=none] (56) at (0.5, -0.25) {};
		\node [style=none] (57) at (2, -0.5) {};
		\node [style=none] (58) at (2.75, -0.5) {};
		\node [style=none] (59) at (3, -0.5) {};
		\node [style=none] (60) at (4, -0.5) {};
		\node [style=none] (61) at (4.25, -0.5) {};
		\node [style=none] (62) at (5, -0.5) {};
		\node [style=none] (63) at (3.5, -0.75) {$f_1$};
		\node [style=none] (64) at (2, -1) {};
		\node [style=none] (65) at (2.75, -1) {};
		\node [style=none] (66) at (3, -1) {};
		\node [style=none] (67) at (4, -1) {};
		\node [style=none] (68) at (4.25, -1) {};
		\node [style=none] (69) at (5, -1) {};
		\node [style=none] (70) at (-7, -1.5) {};
		\node [style=none] (71) at (-5.5, -1.5) {};
		\node [style=none] (72) at (-3, -1.5) {};
		\node [style=none] (73) at (-2, -1.5) {};
		\node [style=none] (74) at (-0.75, -1.5) {};
		\node [style=none] (75) at (0.25, -1.5) {};
		\node [style=none] (76) at (2, -1.75) {};
		\node [style=none] (77) at (3, -1.75) {};
		\node [style=none] (78) at (4, -1.75) {};
		\node [style=none] (79) at (5, -1.75) {};
	\end{pgfonlayer}
	\begin{pgfonlayer}{edgelayer}
		\draw [ultra thick] (19.center) to (17.center);
		\draw [thick, ->, in=-90, out=105] (28.center) to (3.center);
		\draw [thick] (58.center) to (65.center);
		\draw [thick, in=90, out=-90, looseness=0.75] (36.center) to (60.center);
		\draw [thick] (14.center) to (11.center);
		\draw [thick, ->] (24.center) to (2.center);
		\draw [thick] (52.center) to (26.center);
		\draw [ultra thick] (45.center) to (19.center);
		\draw [thick] (27.center) to (53.center);
		\draw [thick, ->, in=90, out=-90] (55.center) to (74.center);
		\draw [thick] (26.center) to (23.center);
		\draw [thick] (61.center) to (58.center);
		\draw [thick, ->, in=90, out=-90] (51.center) to (75.center);
		\draw [thick, ->, in=-90, out=90, looseness=0.75] (12.center) to (6.center);
		\draw [thick, >-, in=-90, out=90, looseness=0.75] (76.center) to (66.center);
		\draw [thick] (65.center) to (68.center);
		\draw [thick] (23.center) to (49.center);
		\draw [thick] (64.center) to (57.center);
		\draw [thick] (31.center) to (10.center);
		\draw [thick] (32.center) to (35.center);
		\draw [thick] (49.center) to (52.center);
		\draw [ultra thick] (17.center) to (43.center);
		\draw [ultra thick] (48.center) to (22.center);
		\draw [thick] (11.center) to (32.center);
		\draw [thick] (68.center) to (61.center);
		\draw [ultra thick] (20.center) to (46.center);
		\draw [thick, >-, in=90, out=-90, looseness=0.75] (8.center) to (15.center);
		\draw [ultra thick] (22.center) to (20.center);
		\draw [thick, in=90, out=-105, looseness=0.75] (34.center) to (62.center);
		\draw [ultra thick, ->] (21.center) to (1.center);
		\draw [ultra thick, ->] (18.center) to (0.center);
		\draw [thick, ->, in=90, out=-90, looseness=0.75] (69.center) to (78.center);
		\draw [thick] (53.center) to (56.center);
		\draw [thick] (15.center) to (36.center);
		\draw [thick, >-, in=90, out=-90, looseness=0.75] (5.center) to (25.center);
		\draw [thick, in=-90, out=90, looseness=0.75] (57.center) to (33.center);
		\draw [thick, >-, in=-90, out=90] (73.center) to (54.center);
		\draw [thick, >-] (72.center) to (50.center);
		\draw [thick] (35.center) to (14.center);
		\draw [thick, ->, in=-90, out=90, looseness=0.75] (10.center) to (7.center);
		\draw [thick] (56.center) to (30.center);
		\draw [thick] (30.center) to (27.center);
		\draw [thick, >-, in=90, out=-90, looseness=0.75] (9.center) to (13.center);
		\draw [thick, >-, in=-90, out=90] (77.center) to (64.center);
		\draw [thick, ->, in=90, out=-90, looseness=0.75] (67.center) to (79.center);
		\draw [ultra thick, >-] (70.center) to (44.center);
		\draw [ultra thick] (46.center) to (48.center);
		\draw [ultra thick, >-] (71.center) to (47.center);
		\draw [thick, >-, in=105, out=-90, looseness=1.25] (4.center) to (29.center);
		\draw [thick] (62.center) to (69.center);
		\draw [thick, in=-90, out=90, looseness=0.75] (59.center) to (31.center);
		\draw [ultra thick] (43.center) to (45.center);
	\end{pgfonlayer}
\end{tikzpicture}}

\normalsize
\end{equation}

\noindent
where the arrow $\mapsto$ represents the functor $E$ and we use the convention of depicting morphisms in $\mathbf{D}(\M{C})$ with thick wires and boxes to avoid confusion. Note that the intuitive alternative of simply juxtaposing the two morphisms as we would in $\M{C}$ fails to produce a completely positive morphism in general, as will become clearer when we define completely positivity in this context.

This category carries all the required structure. We refer the reader to \cite{selinger2007dagger} for a proof of the following:

\begin{myprop}
The category $\mathbf{D}(\M{C})$ inherits a $\dagger$-compact closed structure from $\M{C}$ via the strict monoidal functor $M: \M{C}\rightarrow \mathbf{D}(\M{C})$ defined inductively by
\begin{equation*}
  \left\{
    \begin{array}{rll}
      f_1\otimes f_2 & \mapsto M(f_1)\otimes_\mathbf{D} M(f_2) &;\\
      A & \mapsto A & \text{on objects};\\
      f & \mapsto f\otimes f_* &  \text{on morphisms}.
      
    \end{array} \right.
\end{equation*}
where $f_* = (f^\dagger)^*$ by definition.
\end{myprop}

The functor $M$ shows that we are not losing any expressive power since unambiguous words (represented as maps of $\M{C}$) still admit a faithful representation in doubled form. For reference, the reader can find in Fig. \ref{fig:doubling} a dictionary that translates useful diagrams from one category to the other.

\begin{figure}[b!]
  \scriptsize
  
\InputIfFileExists{table-cpm.tikz}{}{\input{.//tikz//table-cpm.tikz}}

  \normalsize
  \caption{Translation from $\M{C}$ to $\mathbf{D}(\M{C})$.}
  \label{fig:doubling}
\end{figure}

Now, notice that we have a bijective correspondence between states of $\mathbf{D}(\M{C})$, i.e., morphisms $I \rightarrow A$ and operators on $A$ in $\M{C}$. Explicitly, the map $\M{C}(A,A) \rightarrow \M{C}(I,A\otimes A^*)$ is, for an operator $\rho: A \rightarrow A$,

\vspace{-0.2cm}
\begin{equation}

\begin{tikzpicture}[scale=1.9]
	\begin{pgfonlayer}{nodelayer}
		\node [style=none] (0) at (0.75, 0.75) {};
		\node [style=none] (1) at (2, 0.75) {};
		\node [style=none] (2) at (0.75, 0.25) {};
		\node [style=none] (3) at (2, 0.25) {};
		\node [style=none] (4) at (-4.25, 0) {$\rho \mapsto \ulcorner \rho \urcorner = (\rho\otimes 1_{A^*}) \circ \eta_{A^*} = $};
		\node [style=none] (5) at (0.25, 0) {};
		\node [style=none] (6) at (0.75, 0) {};
		\node [style=none] (7) at (1.25, 0) {};
		\node [style=none] (8) at (2, 0) {};
		\node [style=none] (9) at (0.75, -0.25) {{\scriptsize $\rho$}};
		\node [style=none] (10) at (0.25, -0.5) {};
		\node [style=none] (11) at (0.75, -0.5) {};
		\node [style=none] (12) at (1.25, -0.5) {};
		\node [style=none] (13) at (2, -0.5) {};
	\end{pgfonlayer}
	\begin{pgfonlayer}{edgelayer}
		\draw [thick] (5.center) to (10.center);
		\draw [thick, in=90, out=-90, looseness=1.25] (3.center) to (8.center);
		\draw [thick, >-] (1.center) to (3.center);
		\draw [thick] (7.center) to (5.center);
		\draw [thick] (8.center) to (13.center);
		\draw [thick, in=-90, out=90, looseness=1.25] (6.center) to (2.center);
		\draw [thick] (10.center) to (12.center);
		\draw [thick] (12.center) to (7.center);
		\draw [thick, bend right=90, looseness=1.75] (11.center) to (13.center);
		\draw [thick, ->] (2.center) to (0.center);
	\end{pgfonlayer}
\end{tikzpicture}}

\end{equation}
\vspace{-0.2cm}

\noindent
that is easily seen to be an isomorphism by bending back the rightmost wire.\footnote{An application of the yanking equations (\ref{equ:yanking}).} In the special case of states, the generalised inner product generated by the dagger functor can be computed in terms of the canonical trace induced by the compact closed structure (and reduces to the usual inner product on a space of operators in $\Hilb$):
\begin{equation}

\begin{tikzpicture}[scale=1.1]
	\begin{pgfonlayer}{nodelayer}
		\node [style=none] (0) at (-3, 2) {};
		\node [style=none] (1) at (-3, 2) {};
		\node [style=none] (2) at (-1, 2) {};
		\node [style=none] (3) at (0.25, 2) {};
		\node [style=none] (4) at (0.25, 2) {};
		\node [style=none] (5) at (2.25, 2) {};
		\node [style=none] (6) at (2.25, 2) {};
		\node [style=none] (7) at (-1, 1.75) {};
		\node [style=none] (8) at (4.5, 1.75) {};
		\node [style=none] (9) at (5.25, 1.75) {};
		\node [style=none] (10) at (6, 1.75) {};
		\node [style=none] (11) at (7.75, 1.75) {};
		\node [style=none] (12) at (5.25, 1.25) {{\scriptsize $\rho_2^\dagger$}};
		\node [style=none] (13) at (-3.75, 0.5) {};
		\node [style=none] (14) at (-3, 0.5) {};
		\node [style=none] (15) at (-2.25, 0.5) {};
		\node [style=none] (16) at (-1.75, 0.5) {};
		\node [style=none] (17) at (-1, 0.5) {};
		\node [style=none] (18) at (-0.25, 0.5) {};
		\node [style=none] (19) at (4.5, 0.5) {};
		\node [style=none] (20) at (5.25, 0.5) {};
		\node [style=none] (21) at (6, 0.5) {};
		\node [style=none] (22) at (5.25, 0.25) {};
		\node [style=none] (23) at (-9.75, 0) {};
		\node [style=none] (24) at (-9, -0) {};
		\node [style=none] (25) at (-8.75, -0) {};
		\node [style=none] (26) at (-8.25, -0) {};
		\node [style=none] (27) at (-7.75, -0) {};
		\node [style=none] (28) at (-7.5, 0) {};
		\node [style=none] (29) at (-7, 0) {};
		\node [style=none] (30) at (-6.25, -0) {};
		\node [style=none] (31) at (-5.25, -0) {$\mapsto$};
		\node [style=none] (32) at (-3, -0) {{\scriptsize $\rho_1$}};
		\node [style=none] (33) at (-1, -0) {{\scriptsize ${\rho_2}_*$}};
		\node [style=none] (34) at (3.25, -0) {$=$};
		\node [style=none] (35) at (10.5, -0) {$= \Tr(\rho_2^\dagger\rho_1)$};
		\node [style=none] (36) at (4.5, -0.25) {};
		\node [style=none] (37) at (5.25, -0.25) {};
		\node [style=none] (38) at (6, -0.25) {};
		\node [style=none] (39) at (-3.75, -0.5) {};
		\node [style=none] (40) at (-3, -0.5) {};
		\node [style=none] (41) at (-2.25, -0.5) {};
		\node [style=none] (42) at (-1.75, -0.5) {};
		\node [style=none] (43) at (-1, -0.5) {};
		\node [style=none] (44) at (-0.25, -0.5) {};
		\node [style=none] (45) at (5.25, -0.75) {{\scriptsize $\rho_1$}};
		\node [style=none] (46) at (-9, -1) {};
		\node [style=none] (47) at (-7, -1) {};
		\node [style=none] (48) at (-3, -1) {};
		\node [style=none] (49) at (-1, -1) {};
		\node [style=none] (50) at (0.25, -1) {};
		\node [style=none] (51) at (2.25, -1) {};
		\node [style=none] (52) at (4.5, -1.25) {};
		\node [style=none] (53) at (5.25, -1.25) {};
		\node [style=none] (54) at (6, -1.25) {};
		\node [style=none] (55) at (5.25, -2) {};
		\node [style=none] (56) at (7.75, -2) {};
	\end{pgfonlayer}
	\begin{pgfonlayer}{edgelayer}
		\draw [thick, looseness=0.00] (42.center) to (16.center);
		\draw [thick, looseness=0.00] (48.center) to (40.center);
		\draw [thick, looseness=0.00] (39.center) to (13.center);
		\draw [thick, bend left=90, looseness=1.25] (56.center) to (55.center);
		\draw [thick, looseness=0.00] (18.center) to (44.center);
		\draw [thick, looseness=0.00] (38.center) to (36.center);
		\draw [thick, looseness=0.00] (36.center) to (52.center);
		\draw [thick, looseness=0.00] (8.center) to (19.center);
		\draw [ultra thick] (46.center) to (26.center);
		\draw [ultra thick] (27.center) to (28.center);
		\draw [ultra thick] (30.center) to (28.center);
		\draw [thick, bend right=90, looseness=1.50] (11.center) to (9.center);
		\draw [thick, bend right=90, looseness=1.75] (2.center) to (1.center);
		\draw [thick, ->] (37.center) to (22.center);
		\draw [ultra thick] (23.center) to (46.center);
		\draw [thick, looseness=0.00] (41.center) to (39.center);
		\draw [thick, looseness=0.00] (13.center) to (15.center);
		\draw [thick, looseness=0.00] (16.center) to (18.center);
		\draw [ultra thick] (47.center) to (27.center);
		\draw [thick, looseness=0.00] (54.center) to (38.center);
		\draw [thick, <-] (3.center) to (50.center);
		\draw [thick, looseness=0.00] (52.center) to (54.center);
		\draw [thick, bend left=270, looseness=1.75] (5.center) to (4.center);
		\draw [thick, looseness=0.00] (49.center) to (43.center);
		\draw [thick, <-] (56.center) to (11.center);
		\draw [thick, looseness=0.00] (21.center) to (10.center);
		\draw [thick, bend left=90, looseness=1.50] (51.center) to (48.center);
		\draw [ultra thick, bend left=90, looseness=1.75] (24.center) to (29.center);
		\draw [thick, bend left=90, looseness=1.50] (50.center) to (49.center);
		\draw [ultra thick] (25.center) to (23.center);
		\draw [thick, -<] (53.center) to (55.center);
		\draw [thick, looseness=0.00] (44.center) to (42.center);
		\draw [thick, looseness=0.00] (10.center) to (8.center);
		\draw [thick, -<, looseness=0.00] (17.center) to (7.center);
		\draw [thick, looseness=0.00] (19.center) to (21.center);
		\draw [thick, ->, looseness=0.00] (14.center) to (0.center);
		\draw [thick, looseness=0.00] (15.center) to (41.center);
		\draw [thick, >-] (6.center) to (51.center);
		\draw [ultra thick] (26.center) to (25.center);
		\draw [thick] (20.center) to (22.center);
		\draw [ultra thick] (47.center) to (30.center);
	\end{pgfonlayer}
\end{tikzpicture}}

\end{equation} 

We now proceed to introduce complete positivity.

\vspace{0.2cm}
\subsubsection{The $\mathbf{CPM}$ construction---complete positivity  \cite{selinger2007dagger}}  


\begin{definition}\label{def:cp}
A morphism $f: A \rightarrow B$ of $\mathbf{D}(\M{C})$ is completely positive if there exists an object $C$ and a morphism $k: C\otimes A \rightarrow B$, in $\M{C}$, such that $f$ embeds in $\M{C}$ as $(k\otimes k_*) \circ (1_A\otimes \eta_{C^*}\otimes1_{A^*})$ or, pictorially,

\begin{equation}  
\scriptsize

\begin{tikzpicture}[scale=1.3]
	\begin{pgfonlayer}{nodelayer}
		\node [style=none] (0) at (-2, 2.25) {$B$};
		\node [style=none] (1) at (1.75, 2.25) {$B$};
		\node [style=none] (2) at (-2, 1.5) {};
		\node [style=none] (3) at (1.75, 1.5) {};
		\node [style=none] (4) at (5.25, 1.5) {};
		\node [style=none] (5) at (-2.75, 0.5) {};
		\node [style=none] (6) at (-2, 0.5) {};
		\node [style=none] (7) at (-1.25, 0.5) {};
		\node [style=none] (8) at (1.25, 0.5) {};
		\node [style=none] (9) at (1.75, 0.5) {};
		\node [style=none] (10) at (2.75, 0.5) {};
		\node [style=none] (11) at (4.25, 0.5) {};
		\node [style=none] (12) at (5.25, 0.5) {};
		\node [style=none] (13) at (5.75, 0.5) {};
		\node [style=none] (14) at (-2, 0.25) {$f$};
		\node [style=none] (15) at (0, 0.25) {$\mapsto$};
		\node [style=none] (16) at (1.75, 0.25) {$k$};
		\node [style=none, yshift=-0.2mm] (17) at (5.25, 0.25) {$k_*$};
		\node [style=none] (18) at (-2.75, -0.25) {};
		\node [style=none] (19) at (-2, -0.25) {};
		\node [style=none] (20) at (-1.25, -0.25) {};
		\node [style=none] (21) at (-1.25, -0.25) {};
		\node [style=none] (22) at (1.25, -0.25) {};
		\node [style=none] (23) at (1.75, -0.25) {};
		\node [style=none] (24) at (2.25, -0.25) {};
		\node [style=none] (25) at (2.75, -0.25) {};
		\node [style=none] (26) at (4.25, -0.25) {}; 
		\node [style=none] (27) at (4.75, -0.25) {};
		\node [style=none] (28) at (5.25, -0.25) {};
		\node [style=none] (29) at (5.75, -0.25) {};
		\node [style=none] (30) at (-2, -1.25) {};
		\node [style=none] (31) at (1.75, -1.25) {};
		\node [style=none] (32) at (5.25, -1.25) {};
		\node [style=none] (33) at (-2, -1.75) {$A$};
		\node [style=none] (34) at (1.75, -1.75) {$A$};
		\node [style=none] (35) at (3.5, -1.25) {$C$};
		\node [style=none] (36) at (5.25, -1.75) {$\ A^*$};
		\node [style=none] (37) at (5.25, 2.25) {$\ B^*$};
	\end{pgfonlayer}
	\begin{pgfonlayer}{edgelayer}
		\draw [thick, >-] (31.center) to (23.center);
		\draw [thick] (22.center) to (25.center);
		\draw [thick, in=-90, out=-90, looseness=1.00] (27.center) to (24.center);
		\draw [thick] (29.center) to (13.center);
		\draw [thick] (10.center) to (8.center);
		\draw [ultra thick] (18.center) to (21.center);
		\draw [thick, >-] (4.center) to (12.center);
		\draw [ultra thick, ->] (6.center) to (2.center);
		\draw [ultra thick] (5.center) to (18.center);
		\draw [thick, ->] (9.center) to (3.center);
		\draw [ultra thick, >-] (30.center) to (19.center);
		\draw [thick] (13.center) to (11.center);
		\draw [thick, ->] (28.center) to (32.center);
		\draw [thick] (25.center) to (10.center);
		\draw [ultra thick] (21.center) to (7.center);
		\draw [thick] (8.center) to (22.center);
		\draw [thick] (26.center) to (29.center);
		\draw [thick] (11.center) to (26.center);
		\draw [ultra thick] (7.center) to (5.center);
	\end{pgfonlayer}
\end{tikzpicture}}

\normalsize
\end{equation}
\end{definition}

From this last representation, we easily see that the composition of two completely positive maps is completely positive. Similarly, the tensor product of two completely positive maps is completely positive. Therefore, we can define:

\begin{definition}
The category $\mathbf{CPM}(\M{C})$ is the subcategory of $\mathbf{D}(\M{C})$ whose objects are the same and morphisms are completely positive maps.
\end{definition}

$\mathbf{CPM}(\M{C})$ is monoidal and $\otimes_\mathbf{CPM} = \otimes_\mathbf{D}$.
We easily recover the usual notion of positive operator from this definition:  

\vspace{-0.4cm}
\begin{equation}

\begin{tikzpicture}[scale=1.1]
	\begin{pgfonlayer}{nodelayer}
		\node [style=none] (0) at (2.25, 2.5) {};
		\node [style=none] (1) at (4.5, 2.5) {};
		\node [style=none] (2) at (-4.5, 1.5) {};
		\node [style=none] (3) at (-2, 1.5) {};
		\node [style=none] (4) at (1.5, 1.5) {};
		\node [style=none] (5) at (2.25, 1.5) {};
		\node [style=none] (6) at (3, 1.5) {};
		\node [style=none] (7) at (-9, 1) {};
		\node [style=none] (8) at (2.25, 1) {{\scriptsize $k$}};
		\node [style=none] (9) at (-5.25, 0.5) {};
		\node [style=none] (10) at (-4.5, 0.5) {};
		\node [style=none] (11) at (-3.75, 0.5) {};
		\node [style=none] (12) at (-2.75, 0.5) {};
		\node [style=none] (13) at (-2, 0.5) {};
		\node [style=none] (14) at (-1.25, 0.5) {};
		\node [style=none] (15) at (1.5, 0.5) {};
		\node [style=none] (16) at (2.25, 0.5) {};
		\node [style=none] (17) at (3, 0.5) {};
		\node [style=none] (18) at (-10, 0) {};
		\node [style=none] (19) at (-9.75, 0) {};
		\node [style=none] (20) at (-9, 0) {};
		\node [style=none] (21) at (-8, 0) {};
		\node [style=none] (22) at (-6.75, 0) {$\mapsto$};
		\node [style=none] (23) at (-4.5, 0) {{\scriptsize $k$}};
		\node [style=none] (24) at (-2, 0) {{\scriptsize $k_*$}};
		\node [style=none] (25) at (0, 0) {=};
		\node [style=none] (26) at (7.75, 0) {$= \ulcorner k\circ k^\dagger\urcorner$};
		\node [style=none] (27) at (1.5, -0.25) {};
		\node [style=none] (28) at (2.25, -0.25) {};
		\node [style=none] (29) at (3, -0.25) {};
		\node [style=none] (30) at (-5.25, -0.5) {};
		\node [style=none] (31) at (-4.75, -0.5) {};
		\node [style=none] (32) at (-4.5, -0.5) {};
		\node [style=none] (33) at (-3.75, -0.5) {};
		\node [style=none] (34) at (-2.75, -0.5) {};
		\node [style=none] (35) at (-2, -0.5) {};
		\node [style=none] (36) at (-1.75, -0.5) {};
		\node [style=none] (37) at (-1.25, -0.5) {};
		\node [style=none, yshift=0.2 mm] (38) at (2.25, -0.75) {{\scriptsize $\ k^{\dagger}$}};
		\node [style=none] (39) at (-9, -1) {};
		\node [style=none] (40) at (1.5, -1.25) {};
		\node [style=none] (41) at (2.25, -1.25) {};
		\node [style=none] (42) at (3, -1.25) {};
		\node [style=none] (43) at (4.5, -1.25) {};
		\node [style=none] (44) at (4.5, -1.25) {};
	\end{pgfonlayer}
	\begin{pgfonlayer}{edgelayer}
		\draw [thick] (9.center) to (30.center);
		\draw [thick] (14.center) to (12.center);
		\draw [thick] (34.center) to (37.center);
		\draw [thick] (17.center) to (6.center);
		\draw [thick, ->] (5.center) to (0.center);
		\draw [thick] (15.center) to (17.center);
		\draw [thick, >-] (1.center) to (43.center);
		\draw [thick] (30.center) to (33.center);
		\draw [ultra thick] (18.center) to (19.center);
		\draw [thick] (33.center) to (11.center);
		\draw [thick] (37.center) to (14.center);
		\draw [ultra thick] (21.center) to (19.center);
		\draw [thick] (11.center) to (9.center);
		\draw [thick, in=-90, out=-90, looseness=1.75] (44.center) to (41.center);
		\draw [thick] (12.center) to (34.center);
		\draw [thick, >-] (3.center) to (13.center);
		\draw [thick] (28.center) to (16.center);
		\draw [thick] (27.center) to (40.center);
		\draw [ultra thick] (39.center) to (21.center);
		\draw [thick] (42.center) to (29.center);
		\draw [thick] (29.center) to (27.center);
		\draw [thick] (4.center) to (15.center);
		\draw [thick, ->] (10.center) to (2.center);
		\draw [thick, in=-90, out=-90, looseness=1.75] (35.center) to (32.center);
		\draw [ultra thick] (39.center) to (18.center);
		\draw [thick] (40.center) to (42.center);
		\draw [ultra thick, ->] (20.center) to (7.center);
		\draw [thick] (6.center) to (4.center);
	\end{pgfonlayer}
\end{tikzpicture}}

\end{equation}

\noindent
with pure states corresponding to the disconnected case.

Finally, from definition \ref{def:cp} it is clear that, for a morphism $f$ of $\M{C}$, $M(f) = f\otimes f_*$ is completely positive. Thus,

\begin{myprop}
$M$ factors through the embedding $I: \mathbf{CPM}(\M{C}) \hookrightarrow \mathbf{D}(\M{C})$, i.e., there exists a strictly monoidal functor $\tilde{M}: \M{C} \rightarrow \mathbf{CPM}(\M{C})$ such that $M = I \tilde{M}$.
\end{myprop}

\subsubsection{Frobenius algebras}

We are only missing a Frobenius algebra to duplicate and delete information as necessary (Sect. \ref{sec:frobenius}). It is natural to first consider the doubled version of Frobenius algebras in $\M{C}$, i.e. the $\dagger$-Frobenius algebra whose copying map is $M(\Delta)$ and whose deleting map is $M(\iota)$, as doubling preserves both operations. In addition, the monoid operation is clearly completely positive. In more concrete terms, the monoid operation is precisely the point-wise (sometimes called Hadamard) product of matrices. The morphisms of this Frobenius algebra are shown in Fig. \ref{fig:doubling}. 

\subsection{Categorical Model of Meaning: Reprise}
\label{sec:reprise}

We are now ready to put together all the concepts introduced above in the context of a compositional model of meaning. Our aim in this section is to reinterpret the previous model of \cite{Coeckeetal} as a functor from a compact closed grammar to the category $\mathbf{CPM}(\M{C})$, \emph{for any compact closed category} $\M{C}$. Given semantics in the form of a strong monoidal functor $Q: \mathbf{C_F} \rightarrow \M{C}$, our model of meaning is defined by the composition:

\begin{equation}\tilde{M} Q: \mathbf{C_F} \rightarrow \M{C} \rightarrow \mathbf{CPM}(\M{C})
\end{equation} 

Since $\tilde{M}$ sends an object $A$ to the same $A$ in $\mathbf{CPM}(\M{C})$, the mapping of atomic types, their duals and relational types of the grammar occur in exactly the same fashion as in the previous model. Furthermore, note that $Q$ is strongly monoidal and $\tilde{M}$ is strictly monoidal, so the resulting functor is strongly monoidal and, in particular, preserves the compact structure. Thus, we can perform type reductions in $\mathbf{CPM}(\M{C})$ according to the grammatical structure dictated by the category $\mathbf{C_F}$.

Note that we have deliberately abstracted the model to highlight its richness---the category $\M{C}$ could be any compact closed category: $\Hilb$, the category $\mathbf{Rel}$ of sets and relations (in which case we recover a form of Montague semantics \cite{Mon1}) or, as we will see in Sect. \ref{sec:cpm2}, even another iteration of the $\mathbf{CPM}$ construction. 

\begin{definition}


Let $\rho(w_i)$ be a meaning state $I\to \tilde{M}Q(p_i)$ corresponding to word $w_i$ with type $p_i$ in a sentence $w_1\hdots w_n$. Given a type-reduction $\alpha: p_1 \cdot \hdots \cdot p_n \to s$, the meaning of the sentence is defined as:

\vspace{-0.5cm}
\begin{equation*}
  \rho(w_1\hdots w_n) := \tilde{M}Q(\alpha)\big(\rho(w_1) \otimes_\mathbf{CPM} \hdots \otimes_\mathbf{CPM} \rho(w_n)\big)
\end{equation*}

\end{definition}

For example, assigning density matrix representations to the words in the previous example sentence ``trembling shadows play hide and seek'', we obtain the following meaning representation:

\vspace{-5mm}
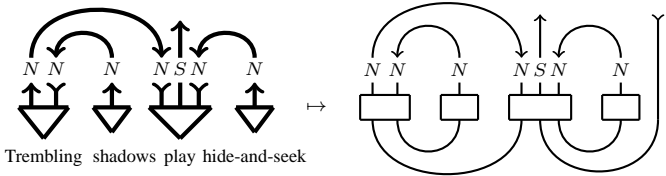
\begin{equation*}
\scriptsize

\begin{tikzpicture}[scale=1.1]
	\begin{pgfonlayer}{nodelayer}
		\node [style=none] (0) at (1.75, 0.5) {};
		\node [style=none] (1) at (1.75, -0.5) {};
		\node [style=none] (2) at (3.75, -0.5) {};
		\node [style=none] (3) at (3.75, 0.5) {};
		\node [style=none] (4) at (6.5, 0.5) {};
		\node [style=none] (5) at (6.5, -0.5) {};
		\node [style=none] (6) at (5, 0.5) {};
		\node [style=none] (7) at (5, -0.5) {};
		\node [style=none] (8) at (3.25, 0.5) {};
		\node [style=none] (9) at (2.25, 0.5) {};
		\node [style=none] (10) at (2.25, -0.5) {};
		\node [style=none] (11) at (3.25, -0.5) {};
		\node [style=none] (12) at (5.75, 0.5) {};
		\node [style=none] (13) at (5.75, -0.5) {};
		\node [style=none] (14) at (3.25, 1) {};
		\node [style=none] (15) at (5.75, 1) {};
		\node [style=none] (16) at (3.25, 2) {};
		\node [style=none] (17) at (5.75, 2) {};
		\node [style=none] (18) at (3.25, 1.5) {$N$};
		\node [style=none] (19) at (5.75, 1.5) {$N$};
		\node [style=none] (20) at (12.25, 1) {};
		\node [style=none] (21) at (13, 0.5) {};
		\node [style=none] (22) at (10.25, -0.5) {};
		\node [style=none] (23) at (13, -0.5) {};
		\node [style=none] (24) at (9.75, 2) {};
		\node [style=none] (25) at (11.5, -0.5) {};
		\node [style=none] (26) at (8.25, -0.5) {};
		\node [style=none] (27) at (9.75, 1.5) {$N$};
		\node [style=none] (28) at (9.75, -0.5) {};
		\node [style=none] (29) at (12.25, 2) {};
		\node [style=none] (30) at (12.25, -0.5) {};
		\node [style=none] (31) at (12.25, 0.5) {};
		\node [style=none] (32) at (8.25, 0.5) {};
		\node [style=none] (33) at (9.75, 0.5) {};
		\node [style=none] (34) at (10.25, 0.5) {};
		\node [style=none] (35) at (12.25, 1.5) {$N$};
		\node [style=none] (36) at (9.75, 1) {};
		\node [style=none] (37) at (7.75, 0.5) {};
		\node [style=none] (38) at (11.5, 0.5) {};
		\node [style=none] (39) at (7.75, -0.5) {};
		\node [style=none] (40) at (9, 2) {};
		\node [style=none] (41) at (9, -0.5) {};
		\node [style=none] (42) at (8.25, 1) {};
		\node [style=none] (43) at (2.25, 1) {};
		\node [style=none] (44) at (2.25, 2) {};
		\node [style=none] (45) at (2.25, 1.5) {$N$};
		\node [style=none] (46) at (8.25, 1.5) {$N$};
		\node [style=none] (47) at (8.25, 2) {};
		\node [style=none] (48) at (13.75, -0.5) {};
		\node [style=none] (49) at (13.75, 3.75) {};
		\node [style=none] (50) at (9, 3.75) {};
		\node [style=none] (51) at (9, 1) {};
		\node [style=none] (52) at (9, 0.5) {};
		\node [style=none] (53) at (9, 1.5) {$S$};
		\node [style=none] (54) at (-8.25, 2) {};
		\node [style=none] (55) at (-3.25, 0) {};
		\node [style=none] (56) at (-4.75, 2) {};
		\node [style=none] (57) at (-5.5, 0) {};
		\node [style=none, text height=1.5 ex, text depth=0.25 ex] (58) at (-4.75, 1.5) {$N$};
		\node [style=none] (59) at (-12, 0) {};
		\node [style=none] (60) at (-2.5, -1) {};
		\node [style=none, text height=1.5 ex, text depth=0.25 ex] (61) at (-10.5, 1.5) {$N$};
		\node [style=none, text height=1.5 ex, text depth=0.25 ex] (62) at (-5.5, 1.5) {$S$};
		\node [style=none] (63) at (-1.75, 0) {};
		\node [style=none, text height=1.5 ex, text depth=0.25 ex] (64) at (-7.75, -2) {shadows};
		\node [style=none] (65) at (-11.5, 1) {};
		\node [style=none, text height=1.5 ex, text depth=0.25 ex] (66) at (-11, -2) {Trembling};
		\node [style=none] (67) at (-4.75, 1) {};
		\node [style=none] (68) at (-5.5, -1.25) {};
		\node [style=none] (69) at (-4.25, 0) {};
		\node [style=none] (70) at (-10.5, 2) {};
		\node [style=none, text height=1.5 ex, text depth=0.25 ex] (71) at (-11.5, 1.5) {$N$};
		\node [style=none] (72) at (-5.5, 1) {};
		\node [style=none] (73) at (-8.25, 1) {};
		\node [style=none] (74) at (-6.75, 0) {};
		\node [style=none] (75) at (-6.25, 0) {};
		\node [style=none] (76) at (-10.5, 1) {};
		\node [style=none] (77) at (-7.5, 0) {};
		\node [style=none] (78) at (-10.5, 0) {};
		\node [style=none] (79) at (-6.25, 2) {};
		\node [style=none] (80) at (-8.25, 0) {};
		\node [style=none, text height=1.5 ex, text depth=0.25 ex] (81) at (-8.25, 1.5) {$N$};
		\node [style=none] (82) at (-6.25, 1) {};
		\node [style=none, text height=1.5 ex, text depth=0.25 ex] (83) at (-6.25, 1.5) {$N$};
		\node [style=none] (84) at (-11.5, 0) {};
		\node [style=none] (85) at (-4.75, 0) {};
		\node [style=none] (86) at (-11.5, 2) {};
		\node [style=none] (87) at (-8.25, -1) {};
		\node [style=none] (88) at (-10, 0) {};
		\node [style=none] (89) at (-9, 0) {};
		\node [style=none, text height=1.5 ex, text depth=0.25 ex] (90) at (-2.5, -2) {hide-and-seek};
		\node [style=none] (91) at (-2.5, 0) {};
		\node [style=none] (92) at (-2.5, 2) {};
		\node [style=none, text height=1.5 ex, text depth=0.25 ex] (93) at (-5.5, -2) {play};
		\node [style=none] (94) at (-11, -1.25) {};
		\node [style=none, text height=1.5 ex, text depth=0.25 ex] (95) at (-2.5, 1.5) {$N$};
		\node [style=none] (96) at (-2.5, 1) {};
		\node [style=none] (97) at (0, 0) {$\mapsto$};
		\node [style=none] (98) at (-5.5, 2) {};
		\node [style=none] (99) at (-5.5, 3.5) {};
	\end{pgfonlayer}
	\begin{pgfonlayer}{edgelayer}
		\draw [thick] (1.center) to (2.center);
		\draw [thick] (2.center) to (3.center);
		\draw [thick] (3.center) to (0.center);
		\draw [thick] (0.center) to (1.center);
		\draw [thick] (7.center) to (5.center);
		\draw [thick] (5.center) to (4.center);
		\draw [thick] (4.center) to (6.center);
		\draw [thick] (6.center) to (7.center);
		\draw [thick, bend right=90, looseness=1.75] (11.center) to (13.center);
		\draw [thick, ->, bend right=90, looseness=1.75] (17.center) to (16.center);
		\draw [thick] (8.center) to (14.center);
		\draw [thick] (12.center) to (15.center);
		\draw [thick] (39.center) to (22.center);
		\draw [thick] (22.center) to (34.center);
		\draw [thick] (34.center) to (37.center);
		\draw [thick] (37.center) to (39.center);
		\draw [thick] (25.center) to (23.center);
		\draw [thick] (23.center) to (21.center);
		\draw [thick] (21.center) to (38.center);
		\draw [thick] (38.center) to (25.center);
		\draw [thick, bend right=90, looseness=1.75] (28.center) to (30.center);
		\draw [thick, ->, bend right=90, looseness=1.75] (29.center) to (24.center);
		\draw [thick] (33.center) to (36.center);
		\draw [thick] (31.center) to (20.center);
		\draw [thick, bend right=90, looseness=1.25] (10.center) to (26.center);
		\draw [thick, ->, bend left=90, looseness=1.25] (44.center) to (47.center);
		\draw [thick] (9.center) to (43.center);
		\draw [thick] (32.center) to (42.center);
		\draw [thick, bend right=90, looseness=1.50] (41.center) to (48.center);
		\draw [thick, >-] (49.center) to (48.center);
		\draw [thick, ->] (40.center) to (50.center);
		\draw [thick] (52.center) to (51.center);
		\draw [ultra thick, -<] (85.center) to (67.center);
		\draw [ultra thick] (89.center) to (77.center);
		\draw [ultra thick] (87.center) to (89.center);
		\draw [ultra thick] (88.center) to (94.center);
		\draw [ultra thick] (55.center) to (63.center);
		\draw [ultra thick] (57.center) to (72.center);
		\draw [ultra thick, ->, bend left=90, looseness=1.25] (86.center) to (79.center);
		\draw [ultra thick] (68.center) to (74.center);
		\draw [ultra thick] (77.center) to (87.center);
		\draw [ultra thick, ->] (80.center) to (73.center);
		\draw [ultra thick] (94.center) to (59.center);
		\draw [ultra thick] (60.center) to (55.center);
		\draw [ultra thick, ->] (84.center) to (65.center);
		\draw [ultra thick] (74.center) to (69.center);
		\draw [ultra thick] (59.center) to (88.center);
		\draw [ultra thick, ->] (91.center) to (96.center);
		\draw [ultra thick, <-, bend left=90, looseness=1.50] (70.center) to (54.center);
		\draw [ultra thick, -<] (78.center) to (76.center);
		\draw [ultra thick] (69.center) to (68.center);
		\draw [ultra thick] (63.center) to (60.center);
		\draw [ultra thick, <-, bend left=90, looseness=1.50] (56.center) to (92.center);
		\draw [ultra thick, -<] (75.center) to (82.center);
		\draw [ultra thick, ->] (98.center) to (99.center);
	\end{pgfonlayer}
\end{tikzpicture}}

\normalsize
\end{equation*}

Diagrammatically, it is clear that in the new setting the partial trace implements meaning composition. Note that diagrams as the above illustrate the flow of ambiguity or information between words. How does ambiguity evolve when composing words to form sentences? This question is very hard to answer precisely in full generality. The key message is that (unambiguous) meaning emerges in the interaction of a word with its context, through the wires. This process of disambiguation is perhaps better understood by studying very simple examples. For instance, it is interesting to examine the interaction of an ambiguous word with a pure meaning word to build intuition---for example the particular interaction of an ambiguous noun with an unambiguous verb. In fact, since density operators are convex sums of pure operators, all interactions are convex combinations of this simple form of word composition. In addition, disambiguation is one of the key NLP tasks on which the previous compositional models were tested and thus constitutes an interesting case study.

\subsection{Introducing ambiguity in formal semantics}

Here, we will work in the category $\mathbf{CPM}(\mathbf{Rel})$. We recall that $\mathbf{Rel}$ is the $\dagger$-compact category of sets and relations. The tensor product is the Cartesian product and the dagger associates to a relation its opposite. Let our sentence set be $S=\{true, false\}$. In $\mathbf{Rel}$, this means that we are only interested in the truth of a sentence, as in Montague semantics. In this context, nouns are subsets of attributes. Given a context to which we pass the meaning of a word,
the meaning of the resulting sentence can be either $\ket{false}$, $\ket{true}$ or $\ket{false} + \ket{true}$, the latter representing superposition, i.e., the case for which the context is insufficient to determine the truth of all the attributes of the word (classically, this can be identified with $false$).

On the other hand, in the internal logic of $\mathbf{CPM}(\mathbf{Rel})$, mixing will add a second dimension that can be interpreted as ambiguous meaning, regardless of truth. The possible values are now:

\begin{minipage}{0.06\textwidth}
        \centering
	\begin{equation*}
		\scriptsize
		
\begin{tikzpicture}[scale=1.1]
	\begin{pgfonlayer}{nodelayer}
		\node [style=none] (0) at (-1, 3) {};
		\node [style=none] (1) at (-4.5, 2) {};
		\node [style=none] (2) at (-2, 2) {};
		\node [style=none] (3) at (-1, 2) {};
		\node [style=none, text height=1.5 ex, text depth=0.25 ex] (4) at (-4.5, 1.5) {$N$};
		\node [style=none, text height=1.5 ex, text depth=0.25 ex] (5) at (-2, 1.5) {$N$};
		\node [style=none] (6) at (-1, 1.5) {$S$};
		\node [style=none] (7) at (-4.5, 1) {};
		\node [style=none] (8) at (-2, 1) {};
		\node [style=none] (9) at (-1, 1) {};
		\node [style=none] (10) at (-5.25, 0) {};
		\node [style=none] (11) at (-4.5, 0) {};
		\node [style=none] (12) at (-3.75, 0) {};
		\node [style=none] (13) at (-2.75, 0) {};
		\node [style=none] (14) at (-2, 0) {};
		\node [style=none] (15) at (-1, 0) {};
		\node [style=none] (16) at (-0.25, 0) {};
		\node [style=none] (17) at (-4.5, -1) {};
		\node [style=none] (18) at (-1.5, -1.25) {};
		\node [style=none, text height=1.5 ex, text depth=0.25 ex] (19) at (-4.5, -1.75) {ambiguous};
		\node [style=none] (20) at (-1.5, -2) {context};
		\node [style=none] (21) at (-4.5, -2.5) {word};
	\end{pgfonlayer}
	\begin{pgfonlayer}{edgelayer}
		\draw [ultra thick, bend left=90, looseness=1.25] (1.center) to (2.center);
		\draw [ultra thick, ->] (3.center) to (0.center);
		\draw [ultra thick] (15.center) to (9.center);
		\draw [ultra thick, -<] (14.center) to (8.center);
		\draw [ultra thick] (16.center) to (18.center);
		\draw [ultra thick] (10.center) to (12.center);
		\draw [ultra thick] (13.center) to (16.center);
		\draw [ultra thick, ->] (11.center) to (7.center);
		\draw [ultra thick] (17.center) to (10.center);
		\draw [ultra thick] (18.center) to (13.center);
		\draw [ultra thick] (12.center) to (17.center);
	\end{pgfonlayer}
\end{tikzpicture}}

		\normalsize
	\end{equation*}
\end{minipage}
\begin{minipage}{0.4\textwidth}
 \footnotesize
\begin{equation*}
  = \left\{
    \begin{array}{l}
      \ket{true}\bra{true},\\
      \ket{false}\bra{false},\\
      (\ket{true}+\ket{false})(\bra{true}+\bra{false}),\\
	  1_S
    \end{array} \right.
\end{equation*}
 \normalsize
\end{minipage}
\vspace{0.05cm}

\noindent
where the identity on $S$ represents ambiguity. 

Note that we use Dirac notation in $\mathbf{Rel}$ rather than set theoretic union and cartesian product, since elements in finite sets can be seen as basis vectors of free modules over the semi-ring of Booleans; a binary relation can be expressed as an adjacency matrix. The trace of a square matrix picks out the elements for which the corresponding relation is reflexive.

Consider the phrase `queen rules'. We allow a few highly simplifying assumptions: first, we restrict our set of nouns to the rather peculiar `Freddy Mercury', `Brian May', `Elisabeth II', `chess', `England' and the empty word $\epsilon$. Moreover, we consider the verb `rule', supposed to have the following unambiguous meaning:

\vspace{-0.3cm}
\small
\begin{eqnarray*}
\ket{rule} &=& \ket{band}\otimes\ket{true}\otimes\ket{\epsilon} + \ket{chess}\otimes\ket{false}\otimes\ket{\epsilon} \\
& & +\ket{elisabeth}\otimes\ket{true}\otimes\ket{england}
\end{eqnarray*}
\normalsize

\noindent
with the obvious $\ket{band}= \ket{freddy}+\ket{brian}$. This definition reflects the fact that a band can rule (understand ``be the best") as well as a monarch. Finally, the ambiguous meaning of `queen' is represented by the following operator:

\vspace{-0.3cm}
\small
\begin{eqnarray*}
\rho(queen) &=& \ket{elisabeth}\bra{elisabeth} 
	\\&&+ \ket{band}\bra{band} + \ket{chess}\bra{chess}
\end{eqnarray*}
\normalsize

\noindent
With the self-evident grammatical structure we can compute the meaning of the sentence diagrammatically:

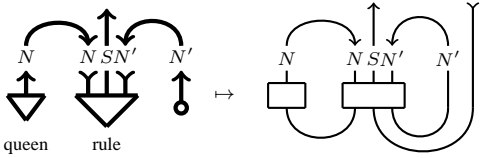
\begin{equation*}
\scriptsize

\begin{tikzpicture}[scale=1.1]
	\begin{pgfonlayer}{nodelayer}
		\node [style=none] (0) at (6, 3.75) {};
		\node [style=none] (1) at (10, 3.75) {};
		\node [style=none] (2) at (-4.75, 3.5) {};
		\node [style=none] (3) at (-8, 2) {};
		\node [style=none] (4) at (-5.5, 2) {};
		\node [style=none] (5) at (-4.75, 2) {};
		\node [style=none] (6) at (-4, 2) {};
		\node [style=none] (7) at (-1.75, 2) {};
		\node [style=none] (8) at (2.5, 2) {};
		\node [style=none] (9) at (5.25, 2) {};
		\node [style=none] (10) at (6, 2) {};
		\node [style=none] (11) at (6.75, 2) {};
		\node [style=none] (12) at (9, 2) {};
		\node [style=none, text height=1.5 ex, text depth=0.25 ex] (13) at (-8, 1.5) {$N$};
		\node [style=none, text height=1.5 ex, text depth=0.25 ex] (14) at (-5.5, 1.5) {$N$};
		\node [style=none, text height=1.5 ex, text depth=0.25 ex] (15) at (-4.75, 1.5) {$S$};
		\node [style=none, text height=1.5 ex, text depth=0.25 ex] (16) at (-4, 1.5) {$N'$};
		\node [style=none, text height=1.5 ex, text depth=0.25 ex] (17) at (-1.75, 1.5) {$N'$};
		\node [style=none] (18) at (2.5, 1.5) {$N$};
		\node [style=none] (19) at (5.25, 1.5) {$N$};
		\node [style=none] (20) at (6, 1.5) {$S$};
		\node [style=none] (21) at (6.75, 1.5) {$N'$};
		\node [style=none] (22) at (9, 1.5) {$N'$};
		\node [style=none] (23) at (-8, 1) {};
		\node [style=none] (24) at (-5.5, 1) {};
		\node [style=none] (25) at (-4.75, 1) {};
		\node [style=none] (26) at (-4, 1) {};
		\node [style=none] (27) at (-1.75, 1) {};
		\node [style=none] (28) at (2.5, 1) {};
		\node [style=none] (29) at (5.25, 1) {};
		\node [style=none] (30) at (6, 1) {};
		\node [style=none] (31) at (6.75, 1) {};
		\node [style=none] (32) at (9, 1) {};
		\node [style=none] (33) at (1.75, 0.5) {};
		\node [style=none] (34) at (2.5, 0.5) {};
		\node [style=none] (35) at (3.25, 0.5) {};
		\node [style=none] (36) at (4.75, 0.5) {};
		\node [style=none] (37) at (5.25, 0.5) {};
		\node [style=none] (38) at (6, 0.5) {};
		\node [style=none] (39) at (6.75, 0.5) {};
		\node [style=none] (40) at (7.25, 0.5) {};
		\node [style=none] (41) at (-8.75, 0) {};
		\node [style=none] (42) at (-8, 0) {};
		\node [style=none] (43) at (-7.25, 0) {};
		\node [style=none] (44) at (-6, 0) {};
		\node [style=none] (45) at (-5.5, 0) {};
		\node [style=none] (46) at (-4.75, 0) {};
		\node [style=none] (47) at (-4, 0) {};
		\node [style=none] (48) at (-3.5, 0) {};
		\node [style=none] (49) at (0, 0) {$\mapsto$};
		\node [style=none, draw, circle, fill=white, minimum size=0.15 cm, ultra thick] (50) at (-1.75, -0.5) {};
		\node [style=none] (51) at (1.75, -0.5) {};
		\node [style=none] (52) at (2.5, -0.5) {};
		\node [style=none] (53) at (3.25, -0.5) {};
		\node [style=none] (54) at (4.75, -0.5) {};
		\node [style=none] (55) at (5.25, -0.5) {};
		\node [style=none] (56) at (6, -0.5) {};
		\node [style=none] (57) at (6.75, -0.5) {};
		\node [style=none] (58) at (7.25, -0.5) {};
		\node [style=none] (59) at (9, -0.5) {};
		\node [style=none] (60) at (9, -0.5) {};
		\node [style=none] (61) at (9, -0.5) {};
		\node [style=none] (62) at (10, -0.5) {};
		\node [style=none] (63) at (-8, -1) {};
		\node [style=none] (64) at (-4.75, -1.25) {};
		\node [style=none, text height=1.5 ex, text depth=0.25 ex] (65) at (-8, -2) {queen};
		\node [style=none, text height=1.5 ex, text depth=0.25 ex] (66) at (-4.75, -2) {rule};
	\end{pgfonlayer}
	\begin{pgfonlayer}{edgelayer}
		\draw [thick] (38.center) to (30.center);
		\draw [ultra thick, <-, bend left=90, looseness=1.50] (6.center) to (7.center);
		\draw [thick] (37.center) to (29.center);
		\draw [ultra thick, -<] (47.center) to (26.center);
		\draw [thick] (51.center) to (53.center);
		\draw [ultra thick, -<] (45.center) to (24.center);
		\draw [thick] (33.center) to (51.center);
		\draw [thick] (58.center) to (40.center);
		\draw [thick, ->] (10.center) to (0.center);
		\draw [thick] (36.center) to (54.center);
		\draw [thick] (35.center) to (33.center);
		\draw [thick] (59.center) to (32.center);
		\draw [ultra thick, ->] (5.center) to (2.center);
		\draw [ultra thick, ->] (50.center) to (27.center);
		\draw [thick, bend right=90, looseness=1.75] (57.center) to (61.center);
		\draw [thick] (53.center) to (35.center);
		\draw [ultra thick, ->, bend left=90, looseness=1.25] (3.center) to (4.center);
		\draw [thick] (34.center) to (28.center);
		\draw [ultra thick] (41.center) to (43.center);
		\draw [thick] (39.center) to (31.center);
		\draw [ultra thick] (46.center) to (25.center);
		\draw [thick] (54.center) to (58.center);
		\draw [ultra thick, ->] (42.center) to (23.center);
		\draw [thick, bend right=90, looseness=1.50] (56.center) to (62.center);
		\draw [thick, bend right=90, looseness=1.50] (52.center) to (55.center);
		\draw [ultra thick] (43.center) to (63.center);
		\draw [ultra thick] (64.center) to (44.center);
		\draw [thick, ->, bend left=90, looseness=1.25] (8.center) to (9.center);
		\draw [thick] (40.center) to (36.center);
		\draw [ultra thick] (63.center) to (41.center);
		\draw [thick, ->, bend right=90, looseness=1.75] (12.center) to (11.center);
		\draw [thick, >-] (1.center) to (62.center);
		\draw [ultra thick] (44.center) to (48.center);
		\draw [ultra thick] (48.center) to (64.center);
	\end{pgfonlayer}
\end{tikzpicture}}

\normalsize
\end{equation*}

\noindent
which, in algebraic form, yields, $\Tr_N(\ket{rule}\bra{rule}\circ (\Tr_{N'}(\rho(queen)) \otimes 1_N')) = 1_S$
\footnote{Note that we delete the subject dimension of the verb in order to reflect the fact that it is used intransitively.}.
In other words, the meaning of the sentence is neither $true$ nor $false$ but still ambiguous. This is because the context that we pass to `queen' is insufficient to disambiguate it (the band \emph{or} the monarch can rule).

Now, if we consider `queen rules England', the only matching pattern in the definition of $\ket{rule}$ is $\ket{elisabeth}$ which corresponds to a \emph{unique} and therefore unambiguous meaning of $\rho(queen)$. Hence, a similar calculation yields $\Tr_N(\ket{rule}\bra{rule}\circ (\Tr_{N'}(\rho(queen)) \otimes \ket{england}\bra{england})) = \ket{true}\bra{true}$ and the sentence is not only true but unambiguous. In this case, the context was sufficient to disambiguate the meaning of the word `queen'.

\subsection{Flow of information with $\dagger$-Frobenius algebras}

In the above examples we used the assumption that a verb tensor had been faithfully constructed according to its grammatical type. However, as we saw in section \ref{sec:frobenius}, concrete constructions might yield operators on a space of tensor order lower than the space to which the functor $\tilde{M}Q$ maps their grammatical type. As before, $\dagger$-Frobenius algebras can be used to solve this type mismatch and encode the information carried by an operator into tensors of higher order.

Assume that we have a distributional model in the form of a vector space $W$ with a distinguished basis and density matrices on $W$ and $W\otimes W$ to represent the meaning of our nouns and verbs, respectively. Using the doubled version of the $\dagger$-Frobenius algebra induced by the basis (as well as the proven empirical method of copying the object) our example sentence is given by:

\vspace{-0.4cm}
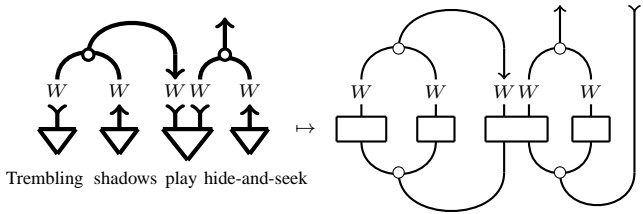
\begin{equation*}
\scriptsize

\begin{tikzpicture}[scale=1.1]
	\begin{pgfonlayer}{nodelayer}
		\node [style=none] (0) at (10.5, 5) {};
		\node [style=none] (1) at (13.5, 5) {};
		\node [style=none] (2) at (-3, 4.5) {};
		\node [style=none, draw, circle, fill=white, minimum size=0.15 cm] (3) at (4, 3.25) {};
		\node [style=none] (4) at (8.25, 3.25) {};
		\node [style=none, draw, circle, fill=white, minimum size=0.15 cm] (5) at (10.5, 3.25) {};
		\node [style=none, draw, circle, fill=white, minimum size=0.15 cm, very thick] (6) at (-8.5, 3) {};
		\node [style=none] (7) at (-5, 3) {};
		\node [style=none] (8) at (-5, 3) {};
		\node [style=none, draw, circle, fill=white, minimum size=0.15 cm, very thick] (9) at (-3, 3) {};
		\node [style=none] (10) at (-9.75, 2) {};
		\node [style=none] (11) at (-7.25, 2) {};
		\node [style=none] (12) at (-5, 2) {};
		\node [style=none] (13) at (-4, 2) {};
		\node [style=none] (14) at (-2, 2) {};
		\node [style=none] (15) at (2.5, 2) {};
		\node [style=none] (16) at (5.5, 2) {};
		\node [style=none] (17) at (8.25, 2) {};
		\node [style=none] (18) at (9.25, 2) {};
		\node [style=none] (19) at (11.75, 2) {};
		\node [style=none, text height=1.5 ex, text depth=0.25 ex] (20) at (-9.75, 1.5) {$W$};
		\node [style=none, text height=1.5 ex, text depth=0.25 ex] (21) at (-7.25, 1.5) {$W$};
		\node [style=none, text height=1.5 ex, text depth=0.25 ex] (22) at (-5, 1.5) {$W$};
		\node [style=none, text height=1.5 ex, text depth=0.25 ex] (23) at (-4, 1.5) {$W$};
		\node [style=none, text height=1.5 ex, text depth=0.25 ex] (24) at (-2, 1.5) {$W$};
		\node [style=none] (25) at (2.5, 1.5) {$W$};
		\node [style=none] (26) at (5.5, 1.5) {$W$};
		\node [style=none] (27) at (8.25, 1.5) {$W$};
		\node [style=none] (28) at (9.25, 1.5) {$W$};
		\node [style=none] (29) at (11.75, 1.5) {$W$};
		\node [style=none] (30) at (-9.75, 1) {};
		\node [style=none] (31) at (-7.25, 1) {};
		\node [style=none] (32) at (-5, 1) {};
		\node [style=none] (33) at (-4, 1) {};
		\node [style=none] (34) at (-2, 1) {};
		\node [style=none] (35) at (2.5, 1) {};
		\node [style=none] (36) at (5.5, 1) {};
		\node [style=none] (37) at (8.25, 1) {};
		\node [style=none] (38) at (9.25, 1) {};
		\node [style=none] (39) at (11.75, 1) {};
		\node [style=none] (40) at (1.5, 0.5) {};
		\node [style=none] (41) at (2.5, 0.5) {};
		\node [style=none] (42) at (3.5, 0.5) {};
		\node [style=none] (43) at (4.75, 0.5) {};
		\node [style=none] (44) at (5.5, 0.5) {};
		\node [style=none] (45) at (5.5, 0.5) {};
		\node [style=none] (46) at (6.25, 0.5) {};
		\node [style=none] (47) at (7.5, 0.5) {};
		\node [style=none] (48) at (8.25, 0.5) {};
		\node [style=none] (49) at (9.25, 0.5) {};
		\node [style=none] (50) at (10, 0.5) {};
		\node [style=none] (51) at (11, 0.5) {};
		\node [style=none] (52) at (11.75, 0.5) {};
		\node [style=none] (53) at (12.5, 0.5) {};
		\node [style=none] (54) at (-10.5, 0) {};
		\node [style=none] (55) at (-9.75, 0) {};
		\node [style=none] (56) at (-9, 0) {};
		\node [style=none] (57) at (-8, 0) {};
		\node [style=none] (58) at (-7.25, 0) {};
		\node [style=none] (59) at (-6.5, 0) {};
		\node [style=none] (60) at (-5.5, 0) {};
		\node [style=none] (61) at (-5, 0) {};
		\node [style=none] (62) at (-4, 0) {};
		\node [style=none] (63) at (-3.5, 0) {};
		\node [style=none] (64) at (-2.75, -0) {};
		\node [style=none] (65) at (-2, -0) {};
		\node [style=none] (66) at (-1.25, -0) {};
		\node [style=none] (67) at (0.25, 0) {$\mapsto$};
		\node [style=none] (68) at (1.5, -0.5) {};
		\node [style=none] (69) at (2.5, -0.5) {};
		\node [style=none] (70) at (3.5, -0.5) {};
		\node [style=none] (71) at (4.75, -0.5) {};
		\node [style=none] (72) at (5.5, -0.5) {};
		\node [style=none] (73) at (6.25, -0.5) {};
		\node [style=none] (74) at (7.5, -0.5) {};
		\node [style=none] (75) at (8.25, -0.5) {};
		\node [style=none] (76) at (9.25, -0.5) {};
		\node [style=none] (77) at (10, -0.5) {};
		\node [style=none] (78) at (11, -0.5) {};
		\node [style=none] (79) at (11.75, -0.5) {};
		\node [style=none] (80) at (12.5, -0.5) {};
		\node [style=none] (81) at (-9.75, -1) {};
		\node [style=none] (82) at (-7.25, -1) {};
		\node [style=none] (83) at (-2, -1) {};
		\node [style=none] (84) at (-4.5, -1.25) {};
		\node [style=none, draw, circle, fill=white, minimum size=0.15 cm] (85) at (4, -1.75) {};
		\node [style=none] (86) at (8.25, -1.75) {};
		\node [style=none, draw, circle, fill=white, minimum size=0.15 cm] (87) at (10.5, -1.75) {};
		\node [style=none] (88) at (13.5, -1.75) {};
		\node [style=none, text height=1.5 ex, text depth=0.25 ex] (89) at (-10.25, -2) {Trembling};
		\node [style=none, text height=1.5 ex, text depth=0.25 ex] (90) at (-7, -2) {shadows};
		\node [style=none, text height=1.5 ex, text depth=0.25 ex] (91) at (-4.75, -2) {play};
		\node [style=none, text height=1.5 ex, text depth=0.25 ex] (92) at (-1.75, -2) {hide-and-seek};
	\end{pgfonlayer}
	\begin{pgfonlayer}{edgelayer}
		\draw [ultra thick, bend left=90, looseness=1.50] (10.center) to (11.center);
		\draw [thick, bend right=90, looseness=1.50] (69.center) to (72.center);
		\draw [thick] (45.center) to (36.center);
		\draw [ultra thick] (84.center) to (60.center);
		\draw [thick, >-] (1.center) to (88.center);
		\draw [ultra thick] (81.center) to (54.center);
		\draw [thick, bend right=90, looseness=1.50] (87.center) to (88.center);
		\draw [thick] (46.center) to (43.center);
		\draw [thick] (80.center) to (53.center);
		\draw [thick] (70.center) to (42.center);
		\draw [thick, ->] (5.center) to (0.center);
		\draw [thick, ->] (4.center) to (17.center);
		\draw [thick] (68.center) to (70.center);
		\draw [ultra thick, ->] (9.center) to (2.center);
		\draw [thick] (43.center) to (71.center);
		\draw [ultra thick, ->] (58.center) to (31.center);
		\draw [thick] (41.center) to (35.center);
		\draw [ultra thick] (60.center) to (63.center);
		\draw [thick] (78.center) to (80.center);
		\draw [thick] (75.center) to (86.center);
		\draw [ultra thick, ->] (65.center) to (34.center);
		\draw [thick] (48.center) to (37.center);
		\draw [thick, bend left=90, looseness=1.75] (18.center) to (19.center);
		\draw [ultra thick, ->] (7.center) to (12.center);
		\draw [thick] (53.center) to (51.center);
		\draw [thick] (71.center) to (73.center);
		\draw [thick, bend right=90, looseness=1.75] (76.center) to (79.center);
		\draw [thick, bend right=90, looseness=1.25] (85.center) to (86.center);
		\draw [ultra thick] (56.center) to (81.center);
		\draw [thick, bend left=90, looseness=1.25] (3.center) to (4.center);
		\draw [thick] (73.center) to (46.center);
		\draw [ultra thick, -<] (62.center) to (33.center);
		\draw [ultra thick] (66.center) to (83.center);
		\draw [thick] (51.center) to (78.center);
		\draw [thick, bend left=90, looseness=1.50] (15.center) to (16.center);
		\draw [thick] (47.center) to (74.center);
		\draw [ultra thick] (57.center) to (59.center);
		\draw [thick] (74.center) to (77.center);
		\draw [ultra thick, bend left=90, looseness=1.25] (6.center) to (8.center);
		\draw [thick] (42.center) to (40.center);
		\draw [ultra thick] (64.center) to (66.center);
		\draw [ultra thick] (83.center) to (64.center);
		\draw [thick] (52.center) to (39.center);
		\draw [ultra thick] (82.center) to (57.center);
		\draw [ultra thick, -<] (61.center) to (32.center);
		\draw [thick] (50.center) to (47.center);
		\draw [thick] (77.center) to (50.center);
		\draw [ultra thick, -<] (55.center) to (30.center);
		\draw [thick] (49.center) to (38.center);
		\draw [thick] (40.center) to (68.center);
		\draw [ultra thick] (54.center) to (56.center);
		\draw [ultra thick] (63.center) to (84.center);
		\draw [ultra thick, bend left=90, looseness=1.50] (13.center) to (14.center);
		\draw [ultra thick] (59.center) to (82.center);
	\end{pgfonlayer}
\end{tikzpicture}}

\normalsize
\end{equation*}

In addition to being a convenient way for creating verb tensors, the application of Frobenius algebras in the new model has another very important practical advantage: it results a significant reduction in the dimensionality of the density matrices, mitigating space complexity problems that might be created from the imposed doubling in a practical implementation.

\subsection{Measuring ambiguity with real data}
\label{sec:experiment}

While a large-scale experiment is out of the scope of this paper, in this section we present some preliminary witnessing results that showcase the potential of the model. Using 2000-dimensional meaning vectors created by the procedure described in Appendix \ref{sec:appendix}, we show how ambiguity evolves for five ambiguous nouns when they are modified by an adjective or a relative clause. For example, `nail' can appear as `rusty nail' or `nail that grows'; in both cases the modifier resolves part of the ambiguity, so we expect that the entropy of the larger compound would be lower than that of the original ambiguous noun. Both types of composition use the Frobenius framework described in Sect. \ref{sec:frobenius}; specifically, composing an adjective with a noun follows (\ref{equ:frob-adj}), while for the relative pronoun case we use (\ref{equ:frob-relpron}). We further remind that for a density matrix $\rho$ with eigen-decomposition $\rho = \sum e_i \ket{e_i}\bra{e_i}$, Von Neumann entropy is given as:

\begin{equation}
  S(\rho) = -\Tr(\rho \ln\rho) = -\sum_i e_i \ln e_i
  \label{equ:entropy}
\end{equation}

The results are presented in Table \ref{tbl:exp}. Note that the entropy of the compounds are always lower than that of the ambiguous noun. Even more interestingly, for some cases (e.g `vessel that sails') the context is so strong that is capable to almost \textit{purify} the meaning of the noun. This demonstrates an important aspect of the proposed model: \textit{disambiguation = purification}. 

\begin{table}[t]
\begin{center}
\begin{tabular}{l|ccc}
  \hline
  \multicolumn{4}{c}{\textbf{Relative Clauses}} \\
  \hline
  \textit{noun}: \textit{verb}$_1$/\textit{verb}$_2$ & \textit{noun} & \textit{noun} that \textit{verb}$_1$ & \textit{noun} that \textit{verb}$_2$ \\
  \hline
  \textit{organ}: enchant/ache & 0.18 & 0.11 & 0.08 \\
  \textit{vessel}: swell/sail  & 0.25 & 0.16 & \textbf{0.01} \\
  \textit{queen}: fly/rule     & 0.28 & 0.14 & 0.16 \\
  \textit{nail}: gleam/grow    & 0.19 & 0.06 & 0.14 \\
  \textit{bank}: overflow/loan & 0.21 & 0.19 & 0.18 \\
  \hline
  \hline
  \multicolumn{4}{c}{\textbf{Adjectives}} \\
  \hline
  \textit{noun}: \textit{adj}$_1$/\textit{adj}$_2$ & \textit{noun} & \textit{adj}$_1$ \textit{noun} & \textit{adj}$_2$ \textit{noun} \\
  \hline
  \textit{organ}: music/body     & 0.18 & 0.10 & 0.13 \\
  \textit{vessel}: blood/naval   & 0.25 & 0.05 & 0.07 \\
  \textit{queen}: fair/chess     & 0.28 & 0.05 & 0.16 \\
  \textit{nail}: rusty/finger    & 0.19 & \textbf{0.04} & 0.11 \\
  \textit{bank}: water/financial & 0.21 & 0.20 & 0.16 \\
  \hline
\end{tabular}
\end{center}
\caption{Computing Entropy for Nouns Modified by Relative Clauses and Adjectives.}
\label{tbl:exp}
\end{table}

Finally, the fact that the composite semantic representations reflect indeed their intended meaning has been verified by performing a number of informal comparisons; for example, `queen that flies' was close to `bee', but `queen that rules' was closer to `palace'; `water bank' was closer to `fish', but `financial bank' was closer to `money', and so on.

\section{Non-commutativity}
\label{sec:non-commutative}

If the last section was concerned with applications of the CPM-construction to model ambiguity, here we discuss the role of the D-construction for the same purpose. Frobenius algebras on objects of $\mathbf{D}(\M{C})$ are not necessarily commutative and thus their associated monoid is not a completely positive morphism. In the quantum physical literature, non-completely positive maps are not usually considered since they are not physically realisable. However, in linguistics, free from these constraints, we could theoretically venture outside of the subcategory $\mathbf{CPM}(\M{C})$, deep into $\mathbf{D}(\M{C})$. More general states could appear as a result of combining mixed states according to the reduction rules of our compositional model. There is no reason for such an operator to be a mixed state itself since there is no constraint in our model that requires sentences to decompose into a mixture of atomic concepts. 

\subsection{Non-commutativity and complete positivity}

Coecke, Heunen and Kissinger \cite{coecke2013channels} introduced the category $\mathbf{CP}^*(\M{C})$ of $\dagger$-Frobenius algebras (with additional technical conditions) and completely positive maps, over an arbitrary $\dagger$-compact category $\M{C}$, in order to study the interaction of classical and quantum systems in a single categorical setting: classical systems are precisely the commutative algebras and completely positive maps are quantum channels, that is, physically realisable processes between systems. Interestingly, in accordance with the content of the no-broadcasting theorem for quantum systems the multiplication of a commutative algebra is a completely positive morphism while the multiplication of a non-commutative algebra is not. It is clear that the meaning composition of words in a sentence is only commutative in exceptional cases. The non commutativity of the grammatical structure reflects this. However, in earlier methods of composition, this complexity was lost in translation when passing to semantics.


With linguistic applications in mind, the $\mathbf{CP}^*$ construction suggests various ways of composing the meaning of words, each corresponding to a specific Frobenius algebra operation. Conceptually, this idea makes sense since a verb does not compose with its subject in the same way that an adjective composes with the noun phrase to which it applies. The various ways of composing words may also offer a theoretical base for the introduction of logic in distributional models of natural language.  

\subsection{A purely quantum algebra}

This is where the richness of $\mathbf{D}(\M{C})$ reveals itself: algebras in this category are more complex and, in particular, allow us to study the action of non-commutative structures---a topic of great interest to formal linguistics where the interaction of words is highly non-commutative. Hereafter we introduce a non-commutative $\dagger$-Frobenius algebra that is not the doubled image of any algebra in $\M{C}$.

\begin{definition}
For every object $A$ of $D(\M{C})$, the morphisms of $D(\M{C})$, $\mu: A \otimes_\mathbf{D} A \rightarrow A$ defined by the following diagram in $\M{C}$:

\begin{equation*}

\begin{tikzpicture}
	\begin{pgfonlayer}{nodelayer}
		\node [style=none] (0) at (-1.75, 2) {};
		\node [style=none] (1) at (-0.75, 2) {};
		\node [style=none] (2) at (-6.25, 1.75) {};
		\node [style=none, draw, circle, fill=black, minimum size=0.15 cm, ultra thick] (3) at (-6.25, 0.25) {};
		\node [style=none] (4) at (10, 0.25) {$=(1_A\otimes \epsilon_A \otimes 1_{A^*})\circ (1_{A\otimes A}\otimes \sigma_{A,A^*})$};
		\node [style=none] (5) at (-4, 0) {$\mapsto$};
		\node [style=none] (6) at (-2.75, 0) {};
		\node [style=none] (7) at (-1.75, 0) {};
		\node [style=none] (8) at (-0.75, 0) {};
		\node [style=none] (9) at (0.25, 0) {};
		\node [style=none] (10) at (-7.5, -0.75) {};
		\node [style=none] (11) at (-5, -0.75) {};
		\node [style=none] (12) at (-2.75, -1) {};
		\node [style=none] (13) at (-1.75, -1) {};
		\node [style=none] (14) at (-0.75, -1) {};
		\node [style=none] (15) at (0.25, -1) {};
	\end{pgfonlayer}
	\begin{pgfonlayer}{edgelayer}
		\draw [thick, ->, in=-90, out=90, looseness=1.25] (6.center) to (0.center);
		\draw [thick, ->, in=90, out=-90, looseness=1.50] (9.center) to (14.center);
		\draw [thick, in=90, out=90, looseness=1.50] (7.center) to (8.center);
		\draw [thick, >-, in=90, out=-90] (1.center) to (9.center);
		\draw [thick, >-] (13.center) to (7.center);
		\draw [thick, ->, in=90, out=-90, looseness=1.25] (8.center) to (15.center);
		\draw [thick, >-] (12.center) to (6.center);
		\draw [ultra thick, ->] (3.center) to (2.center);
		\draw [ultra thick, bend left=90, looseness=1.50] (10.center) to (11.center);
	\end{pgfonlayer}
\end{tikzpicture}}

\end{equation*}

\noindent
and $\iota : I \rightarrow A$ with the following definition in $\M{C}$:

\begin{equation*}

\begin{tikzpicture}
	\begin{pgfonlayer}{nodelayer}
		\node [style=none] (0) at (-3.5, 0.5) {};
		\node [style=none] (1) at (-0.5, 0) {};
		\node [style=none] (2) at (0.75, 0) {};
		\node [style=none] (3) at (-2, -0.5) {$\mapsto$};
		\node [style=none] (4) at (2.75, -0.5) {$ = \eta_{A^*}$};
		\node [style=none, draw, circle, fill=black, minimum size=0.15 cm, ultra thick] (5) at (-3.5, -0.75) {};
	\end{pgfonlayer}
	\begin{pgfonlayer}{edgelayer}
		\draw [ultra thick] (5.center) to (0.center);
		\draw [thick, ->, in=-90, out=-90, looseness=1.75] (2.center) to (1.center);
	\end{pgfonlayer}
\end{tikzpicture}}

\end{equation*}

\noindent
are the multiplication and unit of a $\dagger$-Frobenius algebra $\M{F}_\mathbf{D}$---where $\sigma$ is the natural swap isomorphism in $\M{C}$.
\end{definition}

Proof that the above construction is indeed a $\dagger$-Frobenius algebra can be found in \cite{coecke2012picturing}. The action of the Frobenius multiplication $\mu$ on states $I \rightarrow A$ of $\mathbf{D}(\M{C})$ is particularly interesting; in fact, it implements the composition of operators of $\M{C}$, in $\mathbf{D}(\M{C})$, as evidenced by the next diagram:

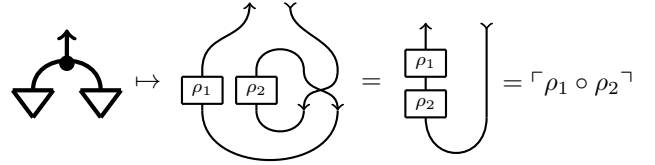
\vspace{-0.3cm}
\begin{equation*}

\begin{tikzpicture}[scale=1.2]
	\begin{pgfonlayer}{nodelayer}
		\node [style=none] (0) at (-4, 3) {};
		\node [style=none] (1) at (-2.5, 3) {};
		\node [style=none] (2) at (2.5, 2.25) {};
		\node [style=none] (3) at (4.75, 2.25) {};
		\node [style=none] (4) at (-10.75, 2) {};
		\node [style=none] (5) at (1.75, 1.25) {};
		\node [style=none] (6) at (2.5, 1.25) {};
		\node [style=none] (7) at (3.25, 1.25) {};
		\node [style=none, draw, circle, fill=black, minimum size=0.15 cm, ultra thick] (8) at (-10.75, 0.75) {};
		\node [style=none] (9) at (2.5, 0.75) {{\scriptsize $\rho_1$}};
		\node [style=none] (10) at (-5.75, 0.5) {};
		\node [style=none] (11) at (-3.75, 0.5) {};
		\node [style=none] (12) at (-2, 0.5) {};
		\node [style=none] (13) at (-0.75, 0.5) {};
		\node [style=none] (14) at (-6.5, 0.25) {};
		\node [style=none] (15) at (-5.75, 0.25) {};
		\node [style=none] (16) at (-5, 0.25) {};
		\node [style=none] (17) at (-4.5, 0.25) {};
		\node [style=none] (18) at (-3.75, 0.25) {};
		\node [style=none] (19) at (-3, 0.25) {};
		\node [style=none] (20) at (1.75, 0.25) {};
		\node [style=none] (21) at (2.5, 0.25) {};
		\node [style=none] (22) at (3.25, 0.25) {};
		\node [style=none] (23) at (-7.75, 0) {$\mapsto$};
		\node [style=none] (24) at (0.5, 0) {$=$};
		\node [style=none] (25) at (7.75, 0) {$= \ulcorner\rho_1 \circ \rho_2\urcorner$};
		\node [style=none] (26) at (-12.75, -0.25) {};
		\node [style=none] (27) at (-12, -0.25) {};
		\node [style=none] (28) at (-11.25, -0.25) {};
		\node [style=none] (29) at (-10.25, -0.25) {};
		\node [style=none] (30) at (-9.5, -0.25) {};
		\node [style=none] (31) at (-8.75, -0.25) {};
		\node [style=none] (32) at (-5.75, -0.25) {{\scriptsize $\rho_1$}};
		\node [style=none] (33) at (-3.75, -0.25) {{\scriptsize $\rho_2$}};
		\node [style=none] (34) at (1.75, -0.25) {};
		\node [style=none] (35) at (2.5, -0.25) {};
		\node [style=none] (36) at (3.25, -0.25) {};
		\node [style=none] (37) at (-6.5, -0.75) {};
		\node [style=none] (38) at (-5.75, -0.75) {};
		\node [style=none] (39) at (-5, -0.75) {};
		\node [style=none] (40) at (-4.5, -0.75) {};
		\node [style=none] (41) at (-3.75, -0.75) {};
		\node [style=none] (42) at (-3, -0.75) {};
		\node [style=none] (43) at (2.5, -0.75) {{\scriptsize $\rho_2$}};
		\node [style=none] (44) at (-5.75, -1) {};
		\node [style=none] (45) at (-3.75, -1) {};
		\node [style=none] (46) at (-2, -1) {};
		\node [style=none] (47) at (-0.75, -1) {};
		\node [style=none] (48) at (-12, -1.25) {};
		\node [style=none] (49) at (-9.5, -1.25) {};
		\node [style=none] (50) at (1.75, -1.25) {};
		\node [style=none] (51) at (2.5, -1.25) {};
		\node [style=none] (52) at (3.25, -1.25) {};
		\node [style=none] (53) at (4.75, -1.25) {};
	\end{pgfonlayer}
	\begin{pgfonlayer}{edgelayer}
		\draw [ultra thick] (31.center) to (49.center);
		\draw [ultra thick] (29.center) to (31.center);
		\draw [thick] (45.center) to (41.center);
		\draw [thick, in=-90, out=-90, looseness=2.00] (53.center) to (51.center);
		\draw [thick, in=-90, out=-90, looseness=1.25] (47.center) to (44.center);
		\draw [thick, ->, in=-90, out=90, looseness=1.25] (10.center) to (0.center);
		\draw [thick, >-] (3.center) to (53.center);
		\draw [thick] (42.center) to (19.center);
		\draw [ultra thick, ->] (8.center) to (4.center);
		\draw [thick] (44.center) to (38.center);
		\draw [ultra thick] (49.center) to (29.center);
		\draw [thick] (50.center) to (52.center);
		\draw [thick, ->, in=90, out=-90, looseness=1.25] (12.center) to (47.center);
		\draw [thick] (10.center) to (15.center);
		\draw [ultra thick, bend left=90, looseness=1.50] (27.center) to (30.center);
		\draw [thick] (5.center) to (20.center);
		\draw [ultra thick] (48.center) to (26.center);
		\draw [thick] (52.center) to (36.center);
		\draw [thick] (35.center) to (21.center);
		\draw [thick] (11.center) to (18.center);
		\draw [thick] (36.center) to (34.center);
		\draw [thick] (37.center) to (39.center);
		\draw [thick] (20.center) to (22.center);
		\draw [thick] (34.center) to (50.center);
		\draw [thick] (16.center) to (14.center);
		\draw [thick] (39.center) to (16.center);
		\draw [thick] (17.center) to (40.center);
		\draw [thick, in=90, out=90, looseness=1.50] (11.center) to (12.center);
		\draw [ultra thick] (28.center) to (48.center);
		\draw [thick, in=-90, out=-90, looseness=1.50] (46.center) to (45.center);
		\draw [thick] (14.center) to (37.center);
		\draw [thick] (22.center) to (7.center);
		\draw [->, thick] (6.center) to (2.center);
		\draw [thick, ->, in=90, out=-90, looseness=1.50] (13.center) to (46.center);
		\draw [thick] (7.center) to (5.center);
		\draw [thick, >-, in=90, out=-90] (1.center) to (13.center);
		\draw [thick] (40.center) to (42.center);
		\draw [ultra thick] (26.center) to (28.center);
		\draw [thick] (19.center) to (17.center);
	\end{pgfonlayer}
\end{tikzpicture}}

\end{equation*}

The meaning of the ``trembling shadows...'' sentence using the algebra $\M{F}_\mathbf{D}$ becomes:

\vspace{-0.5cm}
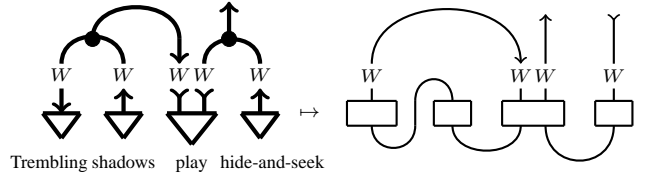
\begin{equation*}
\scriptsize

\begin{tikzpicture}[scale=1.1]
	\begin{pgfonlayer}{nodelayer}
		\node [style=none] (0) at (-3, 4.5) {};
		\node [style=none] (1) at (9.75, 4) {};
		\node [style=none] (2) at (12.5, 4) {};
		\node [style=none, draw, circle, fill=white, minimum size=0.15 cm] (3) at (-8.5, 3) {};
		\node [style=none, draw, circle, fill=black, minimum size=0.20 cm] (4) at (-8.5, 3) {};
		\node [style=none] (5) at (-5, 3) {};
		\node [style=none] (6) at (-5, 3) {};
		\node [style=none, draw, circle, fill=white, minimum size=0.15 cm] (7) at (-3, 3) {};
		\node [style=none, draw, circle, fill=black, minimum size=0.20 cm] (8) at (-3, 3) {};
		\node [style=none] (9) at (-9.75, 2) {};
		\node [style=none] (10) at (-7.25, 2) {};
		\node [style=none] (11) at (-5, 2) {};
		\node [style=none] (12) at (-4, 2) {};
		\node [style=none] (13) at (-1.75, 2) {};
		\node [style=none] (14) at (2.75, 2) {};
		\node [style=none] (15) at (8.75, 2) {};
		\node [style=none] (16) at (9.75, 2) {};
		\node [style=none] (17) at (12.5, 2) {};
		\node [style=none, text height=1.5 ex, text depth=0.25 ex] (18) at (-9.75, 1.5) {$W$};
		\node [style=none, text height=1.5 ex, text depth=0.25 ex] (19) at (-7.25, 1.5) {$W$};
		\node [style=none, text height=1.5 ex, text depth=0.25 ex] (20) at (-5, 1.5) {$W$};
		\node [style=none, text height=1.5 ex, text depth=0.25 ex] (21) at (-4, 1.5) {$W$};
		\node [style=none, text height=1.5 ex, text depth=0.25 ex] (22) at (-1.75, 1.5) {$W$};
		\node [style=none] (23) at (2.75, 1.5) {$W$};
		\node [style=none] (24) at (8.75, 1.5) {$W$};
		\node [style=none] (25) at (9.75, 1.5) {$W$};
		\node [style=none] (26) at (12.5, 1.5) {$W$};
		\node [style=none] (27) at (-9.75, 1) {};
		\node [style=none] (28) at (-7.25, 1) {};
		\node [style=none] (29) at (-5, 1) {};
		\node [style=none] (30) at (-4, 1) {};
		\node [style=none] (31) at (-1.75, 1) {};
		\node [style=none] (32) at (2.75, 1) {};
		\node [style=none] (33) at (8.75, 1) {};
		\node [style=none] (34) at (9.75, 1) {};
		\node [style=none] (35) at (12.5, 1) {};
		\node [style=none] (36) at (1.75, 0.5) {};
		\node [style=none] (37) at (2.75, 0.5) {};
		\node [style=none] (38) at (3.75, 0.5) {};
		\node [style=none] (39) at (4.5, 0.5) {};
		\node [style=none] (40) at (5.25, 0.5) {};
		\node [style=none] (41) at (6, 0.5) {};
		\node [style=none] (42) at (6.75, 0.5) {};
		\node [style=none] (43) at (8, 0.5) {};
		\node [style=none] (44) at (8.75, 0.5) {};
		\node [style=none] (45) at (9.75, 0.5) {};
		\node [style=none] (46) at (10.5, 0.5) {};
		\node [style=none] (47) at (11.75, 0.5) {};
		\node [style=none] (48) at (12.5, 0.5) {};
		\node [style=none] (49) at (13.25, 0.5) {};
		\node [style=none] (50) at (-10.5, 0) {};
		\node [style=none] (51) at (-9.75, 0) {};
		\node [style=none] (52) at (-9, 0) {};
		\node [style=none] (53) at (-8, 0) {};
		\node [style=none] (54) at (-7.25, 0) {};
		\node [style=none] (55) at (-6.5, 0) {};
		\node [style=none] (56) at (-5.5, 0) {};
		\node [style=none] (57) at (-5, 0) {};
		\node [style=none] (58) at (-4, 0) {};
		\node [style=none] (59) at (-3.5, 0) {};
		\node [style=none] (60) at (-2.5, 0) {};
		\node [style=none] (61) at (-1.75, 0) {};
		\node [style=none] (62) at (-1, 0) {};
		\node [style=none] (63) at (0.25, 0) {$\mapsto$};
		\node [style=none] (64) at (1.75, -0.5) {};
		\node [style=none] (65) at (2.75, -0.5) {};
		\node [style=none] (66) at (3.75, -0.5) {};
		\node [style=none] (67) at (4.5, -0.5) {};
		\node [style=none] (68) at (5.25, -0.5) {};
		\node [style=none] (69) at (6, -0.5) {};
		\node [style=none] (70) at (6.75, -0.5) {};
		\node [style=none] (71) at (8, -0.5) {};
		\node [style=none] (72) at (8.75, -0.5) {};
		\node [style=none] (73) at (9.75, -0.5) {};
		\node [style=none] (74) at (10.5, -0.5) {};
		\node [style=none] (75) at (11.75, -0.5) {};
		\node [style=none] (76) at (12.5, -0.5) {};
		\node [style=none] (77) at (13.25, -0.5) {};
		\node [style=none] (78) at (-9.75, -1) {};
		\node [style=none] (79) at (-7.25, -1) {};
		\node [style=none] (80) at (-1.75, -1) {};
		\node [style=none] (81) at (-4.5, -1.25) {};
		\node [style=none, text height=1.5 ex, text depth=0.25 ex] (82) at (-10.25, -2) {Trembling};
		\node [style=none, text height=1.5 ex, text depth=0.25 ex] (83) at (-7.25, -2) {shadows};
		\node [style=none, text height=1.5 ex, text depth=0.25 ex] (84) at (-4.5, -2) {play};
		\node [style=none, text height=1.5 ex, text depth=0.25 ex] (85) at (-1.25, -2) {hide-and-seek};
	\end{pgfonlayer}
	\begin{pgfonlayer}{edgelayer}
		\draw [thick] (64.center) to (66.center);
		\draw [thick] (74.center) to (46.center);
		\draw [thick, bend right=90, looseness=2.00] (41.center) to (39.center);
		\draw [thick] (66.center) to (38.center);
		\draw [ultra thick] (52.center) to (78.center);
		\draw [thick] (38.center) to (36.center);
		\draw [thick] (48.center) to (35.center);
		\draw [thick] (39.center) to (67.center);
		\draw [thick, bend right=90, looseness=1.75] (73.center) to (76.center);
		\draw [ultra thick, ->] (61.center) to (31.center);
		\draw [ultra thick] (80.center) to (60.center);
		\draw [thick, bend left=90, looseness=1.75] (67.center) to (65.center);
		\draw [ultra thick] (50.center) to (52.center);
		\draw [ultra thick, ->] (7.center) to (0.center);
		\draw [ultra thick] (60.center) to (62.center);
		\draw [thick] (40.center) to (68.center);
		\draw [thick] (42.center) to (40.center);
		\draw [thick] (45.center) to (34.center);
		\draw [ultra thick] (62.center) to (80.center);
		\draw [ultra thick, bend left=90, looseness=1.50] (12.center) to (13.center);
		\draw [thick] (46.center) to (43.center);
		\draw [ultra thick, ->] (54.center) to (28.center);
		\draw [ultra thick] (81.center) to (56.center);
		\draw [thick] (36.center) to (64.center);
		\draw [thick] (37.center) to (32.center);
		\draw [thick, >-] (2.center) to (17.center);
		\draw [ultra thick] (78.center) to (50.center);
		\draw [thick] (75.center) to (77.center);
		\draw [thick] (44.center) to (33.center);
		\draw [thick] (68.center) to (70.center);
		\draw [ultra thick] (59.center) to (81.center);
		\draw [thick, ->, bend left=90, looseness=1.25] (14.center) to (15.center);
		\draw [ultra thick] (53.center) to (55.center);
		\draw [ultra thick, ->] (27.center) to (51.center);
		\draw [thick] (70.center) to (42.center);
		\draw [thick, bend right=90, looseness=1.25] (69.center) to (72.center);
		\draw [ultra thick, -<] (58.center) to (30.center);
		\draw [ultra thick, ->] (6.center) to (11.center);
		\draw [thick] (49.center) to (47.center);
		\draw [thick] (47.center) to (75.center);
		\draw [thick] (77.center) to (49.center);
		\draw [ultra thick, -<] (57.center) to (29.center);
		\draw [thick] (71.center) to (74.center);
		\draw [ultra thick, bend left=90, looseness=1.50] (9.center) to (10.center);
		\draw [thick, ->, in=-90, out=90, looseness=1.50] (16.center) to (1.center);
		\draw [thick] (43.center) to (71.center);
		\draw [ultra thick] (56.center) to (59.center);
		\draw [ultra thick] (55.center) to (79.center);
		\draw [ultra thick, bend left=90, looseness=1.25] (3.center) to (5.center);
		\draw [ultra thick] (79.center) to (53.center);
	\end{pgfonlayer}
\end{tikzpicture}}

\normalsize
\end{equation*}

How does composition with the new algebra affect the flow of ambiguity in the simple case of an ambiguous word to which we pass an unambiguous context? Given a projection onto a one-dimensional subspace $|w\rangle\langle w|$ and a density operator $\rho$, the composition $|w\rangle\langle w|\rho$ is a (\emph{not necessarily orthogonal}) projection. In a sense, the meaning of the pure word determined that of the ambiguous word as evidenced by the disconnected topology of the following diagram:

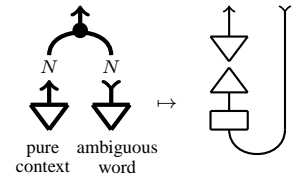
\begin{equation*}
\scriptsize

\begin{tikzpicture}[scale=1.1]
	\begin{pgfonlayer}{nodelayer}
		\node [style=none] (0) at (-3.25, 4) {};
		\node [style=none] (1) at (2.75, 4) {};
		\node [style=none] (2) at (5, 4) {};
		\node [style=none, draw, circle, fill=black, minimum size=0.15 cm, ultra thick] (3) at (-3.25, 3) {};
		\node [style=none] (4) at (2, 2.75) {};
		\node [style=none] (5) at (2.75, 2.75) {};
		\node [style=none] (6) at (3.5, 2.75) {};
		\node [style=none] (7) at (-4.5, 2) {};
		\node [style=none] (8) at (-2, 2) {};
		\node [style=none] (9) at (2.75, 1.75) {};
		\node [style=none, text height=1.5 ex, text depth=0.25 ex] (10) at (-4.5, 1.5) {$N$};
		\node [style=none, text height=1.5 ex, text depth=0.25 ex] (11) at (-2, 1.5) {$N$};
		\node [style=none] (12) at (2.75, 1.5) {};
		\node [style=none] (13) at (-4.5, 1) {};
		\node [style=none] (14) at (-2, 1) {};
		\node [style=none] (15) at (2, 0.5) {};
		\node [style=none] (16) at (2.75, 0.5) {};
		\node [style=none] (17) at (3.5, 0.5) {};
		\node [style=none] (18) at (-5.25, 0) {};
		\node [style=none] (19) at (-4.5, 0) {};
		\node [style=none] (20) at (-3.75, 0) {};
		\node [style=none] (21) at (-2.75, 0) {};
		\node [style=none] (22) at (-2, 0) {};
		\node [style=none] (23) at (-1.25, 0) {};
		\node [style=none] (24) at (0.25, 0) {$\mapsto$};
		\node [style=none] (25) at (2, -0.25) {};
		\node [style=none] (26) at (2.75, -0.25) {};
		\node [style=none] (27) at (3.5, -0.25) {};
		\node [style=none] (28) at (-4.5, -1) {};
		\node [style=none] (29) at (-2, -1) {};
		\node [style=none] (30) at (2, -1) {};
		\node [style=none] (31) at (2.75, -1) {};
		\node [style=none] (32) at (3.5, -1) {};
		\node [style=none] (33) at (5, -1) {};
		\node [style=none, text height=1.5 ex, text depth=0.25 ex] (34) at (-4.75, -1.75) {pure};
		\node [style=none, text height=1.5 ex, text depth=0.25 ex] (35) at (-1.75, -1.75) {ambiguous};
		\node [style=none] (36) at (-4.75, -2.5) {context};
		\node [style=none] (37) at (-1.75, -2.5) {word};
	\end{pgfonlayer}
	\begin{pgfonlayer}{edgelayer}
		\draw [ultra thick, ->] (3.center) to (0.center);
		\draw [ultra thick, bend left=90, looseness=1.25] (7.center) to (8.center);
		\draw [ultra thick] (21.center) to (23.center);
		\draw [thick] (12.center) to (15.center);
		\draw [thick] (25.center) to (27.center);
		\draw [ultra thick] (18.center) to (20.center);
		\draw [ultra thick] (23.center) to (29.center);
		\draw [thick, bend left=90, looseness=1.50] (33.center) to (31.center);
		\draw [thick, ->] (5.center) to (1.center);
		\draw [thick] (15.center) to (17.center);
		\draw [thick] (30.center) to (25.center);
		\draw [ultra thick] (28.center) to (18.center);
		\draw [ultra thick] (20.center) to (28.center);
		\draw [ultra thick] (29.center) to (21.center);
		\draw [thick] (27.center) to (32.center);
		\draw [ultra thick, ->] (19.center) to (13.center);
		\draw [thick] (4.center) to (6.center);
		\draw [thick] (9.center) to (4.center);
		\draw [thick, >-] (2.center) to (33.center);
		\draw [thick] (6.center) to (9.center);
		\draw [thick] (26.center) to (16.center);
		\draw [thick] (32.center) to (30.center);
		\draw [ultra thick, -<] (22.center) to (14.center);
		\draw [thick] (17.center) to (12.center);
	\end{pgfonlayer}
\end{tikzpicture}}

\normalsize
\end{equation*}



\vspace{0.05cm}
\section{Adding lexical entailment}    
\label{sec:cpm2}  

We now demonstrate the advantage of fact that the CPM-construction is an abstract construction, and hence can be applied to any suitable (i.e.~living in a $\dagger$-compact closed category) model of word meaning.

Besides ambiguity, another feature of language which is not captured by  the distributional model is  the fact that the meaning of one word  (= \em hypernym\em) generalises  that of another word   (= \em hyponym\em).  This points at a partial ordering of word meanings.  For example, `painter' generalises `Brueghel'. Density matrices can be endowed with a partial ordering which could play that role, e.g.~the \em Bayesian ordering \em \cite{CoeckeMartin}. This raises the question of how to accommodate both features together in a model of natural language meaning.

Since $\mathbf{CPM}(\M{C})$ is always $\dagger$-compact closed, a canonical solution is obtained by iterating the CPM-construction:
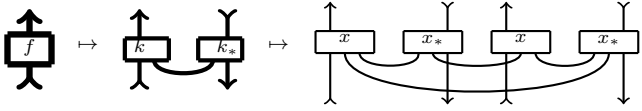
\[
\scriptsize

\begin{tikzpicture}[scale=1.3]
	\begin{pgfonlayer}{nodelayer}
		\node [style=none] (0) at (-4.25, 1.5) {};
		\node [style=none, yshift=1.5pt] (1) at (-5, 0.5) {};
		\node [style=none, yshift=0.2mm] (2) at (-4.25, 0.5) {};
		\node [style=none, yshift=1.5pt] (3) at (-3.5, 0.5) {};
		\node [style=none, yshift=-0.2mm] (4) at (-4.25, 0.25) {$f$};
		\node [style=none] (5) at (-2.25, 0.25) {$\mapsto$};
		\node [style=none, yshift=-1.5pt] (6) at (-5, -0.25) {};
		\node [style=none, yshift=-0.2mm] (7) at (-4.25, -0.25) {};
		\node [style=none, yshift=-1.5pt] (8) at (-3.5, -0.25) {};
		\node [style=none] (9) at (-4.25, -1.25) {};
		\node [style=none] (10) at (4.25, 0.25) {$\mapsto$};
		\node [style=none, yshift=-0.2mm] (11) at (15.5, 0.5) {$x_*$};
		\node [style=none] (12) at (10.5, 0) {};
		\node [style=none] (13) at (12, -1.75) {};
		\node [style=none] (14) at (13, 0) {};
		\node [style=none] (15) at (16, 0) {};
		\node [style=none] (16) at (10, 0) {};
		\node [style=none] (17) at (10.5, 0.75) {};
		\node [style=none] (18) at (8.5, 0) {};
		\node [style=none] (19) at (8.5, 0.75) {};
		\node [style=none] (20) at (15, 0) {};
		\node [style=none] (21) at (6, 0.75) {};
		\node [style=none] (22) at (12, 0) {};
		\node [style=none] (23) at (12.5, 0.5) {$x$};
		\node [style=none] (24) at (11.5, 0.75) {};
		\node [style=none] (25) at (11.5, 0) {};
		\node [style=none] (26) at (7.5, 0.75) {};
		\node [style=none] (27) at (12, 1.75) {};
		\node [style=none] (28) at (16, -1.75) {};
		\node [style=none] (29) at (9, 0) {};
		\node [style=none] (30) at (6, 0) {};
		\node [style=none] (31) at (6, 1.75) {};
		\node [style=none] (32) at (7, 0) {};
		\node [style=none, yshift=-0.2mm] (33) at (9.5, 0.5) {$x_*$};
		\node [style=none] (34) at (14.5, 0) {};
		\node [style=none] (35) at (10, 1.75) {};
		\node [style=none] (36) at (5.5, 0) {};
		\node [style=none] (37) at (10, -1.75) {};
		\node [style=none] (38) at (14.5, 0.75) {};
		\node [style=none] (39) at (13.5, 0.75) {};
		\node [style=none] (40) at (6.5, 0.5) {$x$};
		\node [style=none] (41) at (5.5, 0.75) {};
		\node [style=none] (42) at (12, 0.75) {};
		\node [style=none] (43) at (10, 0.75) {};
		\node [style=none] (44) at (16.5, 0.75) {};
		\node [style=none] (45) at (16, 1.75) {};
		\node [style=none] (46) at (6, -1.75) {};
		\node [style=none] (47) at (16, 0.75) {};
		\node [style=none] (48) at (13.5, 0) {};
		\node [style=none] (49) at (16.5, 0) {};
		\node [style=none] (50) at (7.5, 0) {};
		\node [style=none] (51) at (12.5, 0) {};
		\node [style=none] (52) at (9.5, 0) {};
		\node [style=none] (53) at (15.5, 0) {};
		\node [style=none] (54) at (6.5, 0) {};
		\node [style=none] (55) at (-0.5, 0.5) {};
		\node [style=none] (56) at (0.5, 0.5) {};
		\node [style=none] (57) at (2, -0.25) {};
		\node [style=none] (58) at (-0.5, 1.5) {};
		\node [style=none] (59) at (1.5, 0.5) {};
		\node [style=none] (60) at (0.5, -0.25) {};
		\node [style=none] (61) at (-0.5, -0.25) {};
		\node [style=none] (62) at (2.5, 0.5) {};
		\node [style=none] (63) at (-1, 0.5) {};
		\node [style=none, yshift=-0.2mm] (64) at (2.5, 0.25) {$k_*$};
		\node [style=none] (65) at (2.5, -0.25) {};
		\node [style=none] (66) at (-0.5, 0.25) {$k$};
		\node [style=none] (67) at (-0.5, -1.25) {};
		\node [style=none] (68) at (1.5, -0.25) {};
		\node [style=none] (69) at (3, 0.5) {};
		\node [style=none] (70) at (3, -0.25) {};
		\node [style=none] (71) at (0, -0.25) {};
		\node [style=none] (72) at (2.5, 1.5) {};
		\node [style=none] (73) at (-1, -0.25) {};
		\node [style=none] (74) at (2.5, -1.25) {};
	\end{pgfonlayer}
	\begin{pgfonlayer}{edgelayer}
		\draw [line width=2.4pt] (6.center) to (8.center);
		\draw [line width=2.4pt, ->] (2.center) to (0.center);
		\draw [line width=2.4pt] (1.center) to (6.center);
		\draw [line width=2.4pt, >-] (9.center) to (7.center);
		\draw [line width=2.4pt] (8.center) to (3.center);
		\draw [line width=2.4pt] (3.center) to (1.center);
		\draw [thick=2.4pt, >-] (46.center) to (30.center);
		\draw [thick] (36.center) to (50.center);
		\draw [thick, in=-90, out=-75, looseness=0.75] (29.center) to (32.center);
		\draw [thick] (12.center) to (17.center);
		\draw [thick] (26.center) to (41.center);
		\draw [thick, >-] (35.center) to (43.center);
		\draw [thick, ->] (21.center) to (31.center);
		\draw [thick] (17.center) to (19.center);
		\draw [thick, ->] (16.center) to (37.center);
		\draw [thick] (50.center) to (26.center);
		\draw [thick] (41.center) to (36.center);
		\draw [thick] (18.center) to (12.center);
		\draw [thick] (19.center) to (18.center);
		\draw [thick=2.4pt, >-] (13.center) to (22.center);
		\draw [thick] (25.center) to (48.center);
		\draw [thick, in=-90, out=-90, looseness=0.75] (20.center) to (14.center);
		\draw [thick] (49.center) to (44.center);
		\draw [thick] (39.center) to (24.center);
		\draw [thick, >-] (45.center) to (47.center);
		\draw [thick, ->] (42.center) to (27.center);
		\draw [thick] (44.center) to (38.center);
		\draw [thick, ->] (15.center) to (28.center);
		\draw [thick] (48.center) to (39.center);
		\draw [thick] (24.center) to (25.center);
		\draw [thick] (34.center) to (49.center);
		\draw [thick] (38.center) to (34.center);
		\draw [thick, in=-90, out=-90, looseness=0.50] (51.center) to (52.center);
		\draw [thick, in=-90, out=-90, looseness=0.50] (53.center) to (54.center);
		\draw [ultra thick, >-] (67.center) to (61.center);
		\draw [ultra thick] (73.center) to (60.center);
		\draw [ultra thick, in=-90, out=-90, looseness=0.75] (57.center) to (71.center);
		\draw [ultra thick] (70.center) to (69.center);
		\draw [ultra thick] (56.center) to (63.center);
		\draw [ultra thick, >-] (72.center) to (62.center);
		\draw [ultra thick, ->] (55.center) to (58.center);
		\draw [ultra thick] (69.center) to (59.center);
		\draw [ultra thick, ->] (65.center) to (74.center);
		\draw [ultra thick] (60.center) to (56.center);
		\draw [ultra thick] (63.center) to (73.center);
		\draw [ultra thick] (68.center) to (70.center);
		\draw [ultra thick] (59.center) to (68.center);
	\end{pgfonlayer}
\end{tikzpicture}}

\normalsize  
\]   
Given a word/phrase/sentence meaning:      
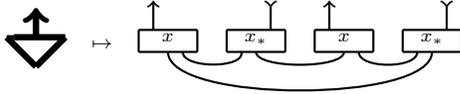
\[
\scriptsize

\begin{tikzpicture}[scale=1.3]
	\begin{pgfonlayer}{nodelayer}
		\node [style=none] (0) at (4.25, 0.25) {$\mapsto$};
		\node [style=none, yshift=-0.2mm] (1) at (15.5, 0.5) {$x_*$};
		\node [style=none] (2) at (10.5, 0) {};
		\node [style=none] (3) at (13, 0) {};
		\node [style=none] (4) at (16, 0) {};
		\node [style=none] (5) at (10, 0) {};
		\node [style=none] (6) at (10.5, 0.75) {};
		\node [style=none] (7) at (8.5, 0) {};
		\node [style=none] (8) at (8.5, 0.75) {};
		\node [style=none] (9) at (15, 0) {};
		\node [style=none] (10) at (6, 0.75) {};
		\node [style=none] (11) at (12, 0) {};
		\node [style=none] (12) at (12.5, 0.5) {$x$};
		\node [style=none] (13) at (11.5, 0.75) {};
		\node [style=none] (14) at (11.5, 0) {};
		\node [style=none] (15) at (7.5, 0.75) {};
		\node [style=none] (16) at (12, 1.75) {};
		\node [style=none] (17) at (9, 0) {};
		\node [style=none] (18) at (6, 0) {};
		\node [style=none] (19) at (6, 1.75) {};
		\node [style=none] (20) at (7, 0) {};
		\node [style=none, yshift=-0.2mm] (21) at (9.5, 0.5) {$x_*$};
		\node [style=none] (22) at (14.5, 0) {};
		\node [style=none] (23) at (10, 1.75) {};
		\node [style=none] (24) at (5.5, 0) {};
		\node [style=none] (25) at (14.5, 0.75) {};
		\node [style=none] (26) at (13.5, 0.75) {};
		\node [style=none] (27) at (6.5, 0.5) {$x$};
		\node [style=none] (28) at (5.5, 0.75) {};
		\node [style=none] (29) at (12, 0.75) {};
		\node [style=none] (30) at (10, 0.75) {};
		\node [style=none] (31) at (16.5, 0.75) {};
		\node [style=none] (32) at (16, 1.75) {};
		\node [style=none] (33) at (16, 0.75) {};
		\node [style=none] (34) at (13.5, 0) {};
		\node [style=none] (35) at (16.5, 0) {};
		\node [style=none] (36) at (7.5, 0) {};
		\node [style=none] (37) at (12.5, 0) {};
		\node [style=none] (38) at (9.5, 0) {};
		\node [style=none] (39) at (15.5, 0) {};
		\node [style=none] (40) at (6.5, 0) {};
		\node [style=none] (41) at (1, 0.5) {};
		\node [style=none] (42) at (2, -0.5) {};
		\node [style=none] (43) at (3, 0.5) {};
		\node [style=none] (44) at (1.25, 0.5) {};
		\node [style=none] (45) at (2, 1.5) {};
		\node [style=none] (46) at (2, 0.5) {};
	\end{pgfonlayer}
	\begin{pgfonlayer}{edgelayer}
		\draw [thick] (24.center) to (36.center);
		\draw [thick, in=-90, out=-75, looseness=0.75] (17.center) to (20.center);
		\draw [thick] (2.center) to (6.center);
		\draw [thick] (15.center) to (28.center);
		\draw [thick, >-] (23.center) to (30.center);
		\draw [thick, ->] (10.center) to (19.center);
		\draw [thick] (6.center) to (8.center);
		\draw [thick] (36.center) to (15.center);
		\draw [thick] (28.center) to (24.center);
		\draw [thick] (7.center) to (2.center);
		\draw [thick] (8.center) to (7.center);
		\draw [thick] (14.center) to (34.center);
		\draw [thick, in=-90, out=-90, looseness=0.75] (9.center) to (3.center);
		\draw [thick] (35.center) to (31.center);
		\draw [thick] (26.center) to (13.center);
		\draw [thick, >-] (32.center) to (33.center);
		\draw [thick, ->] (29.center) to (16.center);
		\draw [thick] (31.center) to (25.center);
		\draw [thick] (34.center) to (26.center);
		\draw [thick] (13.center) to (14.center);
		\draw [thick] (22.center) to (35.center);
		\draw [thick] (25.center) to (22.center);
		\draw [thick, in=-90, out=-90, looseness=0.50] (37.center) to (38.center);
		\draw [thick, in=-90, out=-90, looseness=0.50] (39.center) to (40.center);
		\draw [ultra thick] (41.center) to (44.center);
		\draw [line width=2.4pt] (42.center) to (41.center);
		\draw [line width=2.4pt] (43.center) to (44.center);
		\draw [line width=2.4pt] (42.center) to (43.center);
		\draw [line width=2.4pt, ->] (46.center) to (45.center);
	\end{pgfonlayer}
\end{tikzpicture}}

\normalsize  
\]
lack of any ambiguity or generality correspond to distinct diagrams, respectively:
 \[
\scriptsize

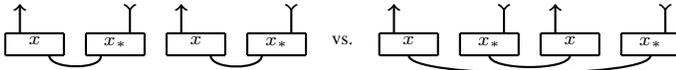
\begin{tikzpicture}[scale=1.3]
	\begin{pgfonlayer}{nodelayer}
		\node [style=none, yshift=-0.2mm] (0) at (15.5, 0) {$x_*$};
		\node [style=none] (1) at (11, -0.5) {};
		\node [style=none] (2) at (13.25, -0.5) {};
		\node [style=none] (3) at (16, -0.5) {};
		\node [style=none] (4) at (10.5, -0.5) {};
		\node [style=none] (5) at (11, 0.25) {};
		\node [style=none] (6) at (9, -0.5) {};
		\node [style=none] (7) at (9, 0.25) {};
		\node [style=none] (8) at (15, -0.5) {};
		\node [style=none] (9) at (6.75, 0.25) {};
		\node [style=none] (10) at (12.25, -0.5) {};
		\node [style=none] (11) at (12.75, 0) {$x$};
		\node [style=none] (12) at (11.75, 0.25) {};
		\node [style=none] (13) at (11.75, -0.5) {};
		\node [style=none] (14) at (8.25, 0.25) {};
		\node [style=none] (15) at (12.25, 1.25) {};
		\node [style=none] (16) at (9.5, -0.5) {};
		\node [style=none] (17) at (6.75, -0.5) {};
		\node [style=none] (18) at (6.75, 1.25) {};
		\node [style=none] (19) at (7.75, -0.5) {};
		\node [style=none, yshift=-0.2mm] (20) at (10, 0) {$x_*$};
		\node [style=none] (21) at (14.5, -0.5) {};
		\node [style=none] (22) at (10.5, 1.25) {};
		\node [style=none] (23) at (6.25, -0.5) {};
		\node [style=none] (24) at (14.5, 0.25) {};
		\node [style=none] (25) at (13.75, 0.25) {};
		\node [style=none] (26) at (7.25, 0) {$x$};
		\node [style=none] (27) at (6.25, 0.25) {};
		\node [style=none] (28) at (12.25, 0.25) {};
		\node [style=none] (29) at (10.5, 0.25) {};
		\node [style=none] (30) at (16.5, 0.25) {};
		\node [style=none] (31) at (16, 1.25) {};
		\node [style=none] (32) at (16, 0.25) {};
		\node [style=none] (33) at (13.75, -0.5) {};
		\node [style=none] (34) at (16.5, -0.5) {};
		\node [style=none] (35) at (8.25, -0.5) {};
		\node [style=none] (36) at (12.75, -0.5) {};
		\node [style=none] (37) at (10, -0.5) {};
		\node [style=none] (38) at (15.5, -0.5) {};
		\node [style=none] (39) at (7.25, -0.5) {};
		\node [style=none] (40) at (-6.5, -0.5) {};
		\node [style=none] (41) at (-4.5, -0.5) {};
		\node [style=none] (42) at (1.75, -0.5) {};
		\node [style=none] (43) at (1, 0.25) {};
		\node [style=none] (44) at (-0.5, -0.5) {};
		\node [style=none] (45) at (0, -0.5) {};
		\node [style=none] (46) at (0.5, -0.5) {};
		\node [style=none] (47) at (-5.5, -0.5) {};
		\node [style=none] (48) at (-0.5, 1.25) {};
		\node [style=none, yshift=-0.2mm] (49) at (-2.75, 0) {$x_*$};
		\node [style=none] (50) at (-6, -0.5) {};
		\node [style=none] (51) at (-0.5, 0.25) {};
		\node [style=none] (52) at (-3.75, 0.25) {};
		\node [style=none] (53) at (0, 0) {$x$};
		\node [style=none] (54) at (-1.75, -0.5) {};
		\node [style=none] (55) at (2.75, -0.5) {};
		\node [style=none] (56) at (3.25, 1.25) {};
		\node [style=none] (57) at (-5, -0.5) {};
		\node [style=none] (58) at (-3.75, -0.5) {};
		\node [style=none] (59) at (3.25, -0.5) {};
		\node [style=none] (60) at (1, -0.5) {};
		\node [style=none] (61) at (2.25, -0.5) {};
		\node [style=none] (62) at (-1, -0.5) {};
		\node [style=none] (63) at (3.75, -0.5) {};
		\node [style=none] (64) at (-1, 0.25) {};
		\node [style=none] (65) at (-6, 0.25) {};
		\node [style=none] (66) at (-2.25, -0.5) {};
		\node [style=none] (67) at (1.75, 0.25) {};
		\node [style=none, yshift=-0.2mm] (68) at (2.75, 0) {$x_*$};
		\node [style=none] (69) at (-1.75, 0.25) {};
		\node [style=none] (70) at (-2.25, 0.25) {};
		\node [style=none] (71) at (-2.75, -0.5) {};
		\node [style=none] (72) at (-6, 1.25) {};
		\node [style=none] (73) at (-3.25, -0.5) {};
		\node [style=none] (74) at (-4.5, 0.25) {};
		\node [style=none] (75) at (-5.5, 0) {$x$};
		\node [style=none] (76) at (-2.25, 1.25) {};
		\node [style=none] (77) at (3.75, 0.25) {};
		\node [style=none] (78) at (3.25, 0.25) {};
		\node [style=none] (79) at (-6.5, 0.25) {};
		\node [style=none] (80) at (5, 0) {vs.};
	\end{pgfonlayer}
	\begin{pgfonlayer}{edgelayer}
		\draw [thick] (23.center) to (35.center);
		\draw [thick] (1.center) to (5.center);
		\draw [thick] (14.center) to (27.center);
		\draw [thick, >-] (22.center) to (29.center);
		\draw [thick, ->] (9.center) to (18.center);
		\draw [thick] (5.center) to (7.center);
		\draw [thick] (35.center) to (14.center);
		\draw [thick] (27.center) to (23.center);
		\draw [thick] (6.center) to (1.center);
		\draw [thick] (7.center) to (6.center);
		\draw [thick] (13.center) to (33.center);
		\draw [thick] (34.center) to (30.center);
		\draw [thick] (25.center) to (12.center);
		\draw [thick, >-] (31.center) to (32.center);
		\draw [thick, ->] (28.center) to (15.center);
		\draw [thick] (30.center) to (24.center);
		\draw [thick] (33.center) to (25.center);
		\draw [thick] (12.center) to (13.center);
		\draw [thick] (21.center) to (34.center);
		\draw [thick] (24.center) to (21.center);
		\draw [thick, in=-90, out=-90, looseness=0.50] (36.center) to (37.center);
		\draw [thick, in=-90, out=-90, looseness=0.25] (38.center) to (39.center);
		\draw [thick] (40.center) to (41.center);
		\draw [thick, in=-90, out=-75, looseness=0.75] (73.center) to (57.center);
		\draw [thick] (54.center) to (69.center);
		\draw [thick] (74.center) to (79.center);
		\draw [thick, >-] (76.center) to (70.center);
		\draw [thick, ->] (65.center) to (72.center);
		\draw [thick] (69.center) to (52.center);
		\draw [thick] (41.center) to (74.center);
		\draw [thick] (79.center) to (40.center);
		\draw [thick] (58.center) to (54.center);
		\draw [thick] (52.center) to (58.center);
		\draw [thick] (62.center) to (60.center);
		\draw [thick, in=-90, out=-90, looseness=0.75] (61.center) to (46.center);
		\draw [thick] (63.center) to (77.center);
		\draw [thick] (43.center) to (64.center);
		\draw [thick, >-] (56.center) to (78.center);
		\draw [thick, ->] (51.center) to (48.center);
		\draw [thick] (77.center) to (67.center);
		\draw [thick] (60.center) to (43.center);
		\draw [thick] (64.center) to (62.center);
		\draw [thick] (42.center) to (63.center);
		\draw [thick] (67.center) to (42.center);
	\end{pgfonlayer}
\end{tikzpicture}}
  
\normalsize  
\]

Ambiguity or generality can be then measured by taking the von Neumann entropy of the following operators respectively:

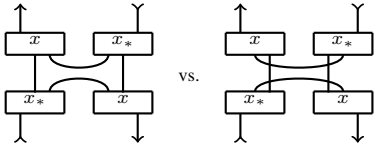
\[
\scriptsize

\begin{tikzpicture}[scale=1.3]
	\begin{pgfonlayer}{nodelayer}
		\node [style=none, yshift=-0.2mm] (0) at (3.75, -0.75) {$x_*$};
		\node [style=none] (1) at (7.75, 0.75) {};
		\node [style=none] (2) at (6.25, -0.5) {};
		\node [style=none] (3) at (3.25, -0.5) {};
		\node [style=none] (4) at (7.25, 0.75) {};
		\node [style=none] (5) at (7.75, 1.5) {};
		\node [style=none] (6) at (5.75, 0.75) {};
		\node [style=none] (7) at (5.75, 1.5) {};
		\node [style=none] (8) at (4.25, -0.5) {};
		\node [style=none] (9) at (3.25, 1.5) {};
		\node [style=none] (10) at (7.25, -0.5) {};
		\node [style=none] (11) at (6.75, -0.75) {$x$};
		\node [style=none] (12) at (7.75, -1.25) {};
		\node [style=none] (13) at (7.75, -0.5) {};
		\node [style=none] (14) at (4.75, 1.5) {};
		\node [style=none] (15) at (7.25, -2.25) {};
		\node [style=none] (16) at (6.25, 0.75) {};
		\node [style=none] (17) at (3.25, 0.75) {};
		\node [style=none] (18) at (3.25, 2.5) {};
		\node [style=none] (19) at (4.25, 0.75) {};
		\node [style=none, yshift=-0.2mm] (20) at (6.75, 1.25) {$x_*$};
		\node [style=none] (21) at (4.75, -0.5) {};
		\node [style=none] (22) at (7.25, 2.5) {};
		\node [style=none] (23) at (2.75, 0.75) {};
		\node [style=none] (24) at (4.75, -1.25) {};
		\node [style=none] (25) at (5.75, -1.25) {};
		\node [style=none] (26) at (3.75, 1.25) {$x$};
		\node [style=none] (27) at (2.75, 1.5) {};
		\node [style=none] (28) at (7.25, -1.25) {};
		\node [style=none] (29) at (7.25, 1.5) {};
		\node [style=none] (30) at (2.75, -1.25) {};
		\node [style=none] (31) at (3.25, -2.25) {};
		\node [style=none] (32) at (3.25, -1.25) {};
		\node [style=none] (33) at (5.75, -0.5) {};
		\node [style=none] (34) at (2.75, -0.5) {};
		\node [style=none] (35) at (4.75, 0.75) {};
		\node [style=none] (36) at (6.75, -0.5) {};
		\node [style=none] (37) at (6.75, 0.75) {};
		\node [style=none] (38) at (3.75, -0.5) {};
		\node [style=none] (39) at (3.75, 0.75) {};
		\node [style=none] (40) at (14.75, 2.5) {};
		\node [style=none] (41) at (15.25, 0.75) {};
		\node [style=none] (42) at (12.25, 1.5) {};
		\node [style=none] (43) at (13.25, 1.5) {};
		\node [style=none] (44) at (14.75, -2.25) {};
		\node [style=none] (45) at (10.25, 0.75) {};
		\node [style=none] (46) at (10.75, -1.25) {};
		\node [style=none] (47) at (10.75, -2.25) {};
		\node [style=none] (48) at (14.75, 1.5) {};
		\node [style=none] (49) at (10.75, 0.75) {};
		\node [style=none] (50) at (11.25, 0.75) {};
		\node [style=none] (51) at (13.25, -1.25) {};
		\node [style=none] (52) at (10.25, -1.25) {};
		\node [style=none] (53) at (15.25, -1.25) {};
		\node [style=none] (54) at (14.75, -0.5) {};
		\node [style=none] (55) at (11.75, 0.75) {};
		\node [style=none] (56) at (11.25, 1.25) {$x$};
		\node [style=none] (57) at (14.75, -1.25) {};
		\node [style=none, yshift=-0.2mm] (58) at (14.25, 1.25) {$x_*$};
		\node [style=none] (59) at (11.25, -0.5) {};
		\node [style=none] (60) at (13.25, 0.75) {};
		\node [style=none] (61) at (12.25, -0.5) {};
		\node [style=none] (62) at (12.25, -1.25) {};
		\node [style=none, yshift=-0.2mm] (63) at (11.25, -0.75) {$x_*$};
		\node [style=none] (64) at (10.25, 1.5) {};
		\node [style=none] (65) at (13.25, -0.5) {};
		\node [style=none] (66) at (12.25, 0.75) {};
		\node [style=none] (67) at (10.75, 1.5) {};
		\node [style=none] (68) at (11.75, -0.5) {};
		\node [style=none] (69) at (14.25, -0.75) {$x$};
		\node [style=none] (70) at (10.75, -0.5) {};
		\node [style=none] (71) at (10.25, -0.5) {};
		\node [style=none] (72) at (15.25, -0.5) {};
		\node [style=none] (73) at (15.25, 1.5) {};
		\node [style=none] (74) at (10.75, 2.5) {};
		\node [style=none] (75) at (14.75, 0.75) {};
		\node [style=none] (76) at (14.25, -0.5) {};
		\node [style=none] (77) at (14.25, 0.75) {};
		\node [style=none] (78) at (13.75, -0.5) {};
		\node [style=none] (79) at (13.75, 0.75) {};
		\node [style=none] (80) at (9, 0) {vs.};
	\end{pgfonlayer}
	\begin{pgfonlayer}{edgelayer}
		\draw [thick] (23.center) to (35.center);
		\draw [thick, in=-90, out=-90, looseness=0.75] (16.center) to (19.center);
		\draw [thick] (1.center) to (5.center);
		\draw [thick] (14.center) to (27.center);
		\draw [thick, >-] (22.center) to (29.center);
		\draw [thick, ->] (9.center) to (18.center);
		\draw [thick] (5.center) to (7.center);
		\draw [thick] (35.center) to (14.center);
		\draw [thick] (27.center) to (23.center);
		\draw [thick] (6.center) to (1.center);
		\draw [thick] (7.center) to (6.center);
		\draw [thick] (13.center) to (33.center);
		\draw [thick, in=90, out=90, looseness=0.75] (8.center) to (2.center);
		\draw [thick] (34.center) to (30.center);
		\draw [thick] (25.center) to (12.center);
		\draw [thick, >-] (31.center) to (32.center);
		\draw [thick, ->] (28.center) to (15.center);
		\draw [thick] (30.center) to (24.center);
		\draw [thick] (33.center) to (25.center);
		\draw [thick] (12.center) to (13.center);
		\draw [thick] (21.center) to (34.center);
		\draw [thick] (24.center) to (21.center);
		\draw [thick, in=-90, out=-90, looseness=0.50] (36.center) to (37.center);
		\draw [thick, in=-90, out=-90, looseness=0.50] (38.center) to (39.center);
		\draw [thick] (45.center) to (66.center);
		\draw [thick, in=-90, out=-90, looseness=0.50] (77.center) to (50.center);
		\draw [thick] (41.center) to (73.center);
		\draw [thick] (42.center) to (64.center);
		\draw [thick, >-] (40.center) to (48.center);
		\draw [thick, ->] (67.center) to (74.center);
		\draw [thick] (73.center) to (43.center);
		\draw [thick] (66.center) to (42.center);
		\draw [thick] (64.center) to (45.center);
		\draw [thick] (60.center) to (41.center);
		\draw [thick] (43.center) to (60.center);
		\draw [thick] (72.center) to (65.center);
		\draw [thick, in=90, out=90, looseness=0.50] (59.center) to (76.center);
		\draw [thick] (71.center) to (52.center);
		\draw [thick] (51.center) to (53.center);
		\draw [thick, >-] (47.center) to (46.center);
		\draw [thick, ->] (57.center) to (44.center);
		\draw [thick] (52.center) to (62.center);
		\draw [thick] (65.center) to (51.center);
		\draw [thick] (53.center) to (72.center);
		\draw [thick] (61.center) to (71.center);
		\draw [thick] (62.center) to (61.center);
		\draw [thick, in=-90, out=-90, looseness=0.50] (68.center) to (55.center);
		\draw [thick, in=-90, out=-90, looseness=0.50] (78.center) to (79.center);
	\end{pgfonlayer}
\end{tikzpicture}}

\normalsize  
\]

\section{Conclusion and Future Work}
\label{sec:conclusion}

In this paper we detailed a compositional distributional model of meaning capable of explicitly handling lexical ambiguity. We discussed its theoretical properties and demonstrated its potential for real-world natural language processing tasks by a small-scale experiment. A large-scale evaluation will be our challenging next step, aiming to provide empirical evidence regarding the effectiveness of the model in general and the performance of the different Frobenius algebras in particular. On the theoretical side, the logic of ambiguity in $\mathbf{CPM}(\mathbf{Rel})$, the non-commutative features of the D-construction as well as further exploration of nested levels of CPM, each deserve a separate treatment. In addition one important weakness of distributional models is the representation of words that serve a purely logical role, like logical connectives or negation. Density operators support a form of logic whose distributional and compositional properties could be examined, potentially providing a solution to this long-standing problem of compositional distributional models. 


\bibliographystyle{IEEEtran}
\bibliography{refs}


\appendices
\section{From Theory to Practice}
\label{sec:appendix}

The purpose of this appendix is to show how the theoretical ideas presented in this paper can take a concrete form using standard natural language processing techniques. The setting we present below has been used for the mini-experiments in Sect. \ref{sec:experiment}. We approach the creation of density matrices as a three-step process: (a) we first produce an ambiguous semantic space; (b) we apply a word sense induction method on it in order to associate each word with a set of sense vectors; and finally (c) we use the sense vectors in order to create a density matrix for each word. These steps are described in separate sections below.

\subsection{Creating a Concrete Semantic Space}
\label{sec:semspace}

We train our basic vector space using ukWaC, a corpus of English text with 2 billion words (100 million sentences). The basis of the vector space consists of the 2,000 most frequent content words (nouns, verbs, adjectives, and adverbs), excluding a list of \textit{stop words}.\footnote{That is, very common words with low information content, such as the verbs `get' and `take' or adverbs like `really' and `always'.} Furthermore, the vector space is lemmatized and unambiguous regarding syntactic information; in other words, each vector is uniquely identified by a (\textit{lemma},\textit{pos-tag}) pair, which means for example that `book' as a noun and `book' as a verb are represented by different meaning vectors. The weights of each vector are set to the ratio of the probability of the context word $c_i$ given the target word $t$ to the probability of the context word overall, as follows:

\begin{equation*}
   v_i(t) = \frac{p(c_i|t)}{p(c_i)} = \frac{\text{count}(c_i,t) \cdot \text{count}(total)}{\text{count}(t) \cdot \text{count}(c_i)}
\label{equ:weights}
\end{equation*}

\noindent where $\text{count}(c_i,t)$ refers to how many times $c_i$ appears in the context of $t$ (that is, in a 5-word window at either side of $t$) and $\text{count}(total)$ is the total number of word tokens in the corpus. 


\subsection{Word Sense Induction}
\label{sec:wsi}

The notion of word sense induction, that is, the task of detecting the different meanings under which a word appears in a text, is intimately connected with that of distributional hypothesis---that the meaning of a word is always context-dependent. If we had a way to create a vectorial representation for the contexts in which a specific word occurs, then, a clustering algorithm could be applied in order to create groupings of these contexts that hopefully reveal different usages of the word---\textit{different meanings}---in the training corpus. 

This intuitive idea was first presented by Sch\"utze \cite{Schutze} in 1998, and more or less is the cornerstone of every unsupervised word sense induction and disambiguation method based on semantic word spaces up to today. The approach we use is a direct variation of this standard technique. For what follows, we assume that each word in the vocabulary has already been assigned to an \textit{ambiguous} semantic vector by following typical distributional procedures, for example similar to the setting described in Sect. \ref{sec:semspace}. 

We assume for simplicity that the context is defined at the sentence level. First, each context for a target word $w_t$ is represented by a \textit{context vector} of the form $\frac{1}{n}\sum_{i=1}^n\ket{w_i}$, where $\ket{w_i}$ is the semantic vector of some other word $w_i \neq w_t$ in the same context. Next, we apply hierarchical agglomerative clustering on this set of vectors in order to discover the latent senses of $w_t$. Ideally, the contexts of $w_t$ will vary according to the specific meaning in which this word has been used. Table \ref{tbl:wsi} provides a visualization of the outcome of this process for the ambiguous word `vessel'. Each meaning is visualized as a list of the most dominant words in the corresponding cluster, ranked by their TF-IDF values.

\begin{table}[t]
\begin{center}
\begin{tabular}{l}
\hline
\textbf{Meaning 1:} 24070 contexts \\
\hline
\begin{minipage}{8cm}
\begin{flushleft}
port owner cargo fleet sailing ferry craft Navy merchant cruise navigation officer metre voyage authority deck coast launch fishery island charter Harbour pottery radio trip pay River Agency Scotland sell duty visit fish insurance skipper Roman sink War shore sail town Coastguard assistance Maritime registration call rescue bank Museum captain incident customer States yacht mooring barge comply landing Ireland sherd money Scottish tow tug maritime wreck board visitor tanker freight purchase lifeboat 
\end{flushleft}
\end{minipage}
\\
\hline\hline
\textbf{Meaning 2:} 5930 contexts \\
\hline
\begin{minipage}{8cm}
\begin{flushleft}
clot complication haemorrhage lymph stem VEGF Vitamin glucose penis endothelium retinopathy spasm antibody clotting AMD coagulation marrow lesion angina blindness medication graft vitamin vasoconstriction virus proliferation Ginkgo diabetic ventricle thickening tablet anaemia thrombus Vein leukocyte scleroderma stimulation degeneration homocysteine Raynaud breathe mediator Biloba Diabetes LDL metabolism Gene infiltrate atheroma arthritis lymphocyte lobe C's histamine melanoma gut dysfunction vitro triglyceride infarction lipoprotein 
\end{flushleft}
\end{minipage}
\\
\hline
\end{tabular}
\end{center}

\vspace{0.2cm}
\caption{Derived Meanings for Word `Vessel'.}
\label{tbl:wsi}
\end{table}

We take the centroid of each cluster as the vectorial representation of the corresponding sense/meaning. Thus, each word $w$ is initially represented by a tuple $(\ket{w}, S_w)$, where $\ket{w}$ is the ambiguous semantic vector of the word as created by the usual distributional practice, and $S_w$ is a set of \textit{sense vectors} (that is, centroids of context vectors clusters) produced by the above procedure. 

Note that our approach takes place at the vector level (as opposed to tensors of higher order), so it provides a natural way to create sets of meaning vectors for ``atomic'' words of the language, that is, for nouns. It turns out that the generalization of this to tensors of higher order is straightforward, since the clustering step has already equipped us with a number of sets consisting of context vectors, each one of which stands in one-to-one correspondence with a set of contexts reflecting a different semantic usage of the higher-order word. One then can use, for example, the argument ``tensoring and summing'' procedure of \cite{GrefenSadr1} (briefly described in Sect. \ref{sec:frobenius}) in order to compute the meaning of the $i$th sense of a word of arity $n$ as:

\begin{equation}
\ket{word}_i = \sum\limits_{c \in C_i} \bigotimes\limits_{k=1}^{n} \ket{arg_{k,c}}
\end{equation}

\noindent where $C_i$ is the set of contexts associated with the $i$th sense, and $arg_{k,c}$ denotes the $k$th argument of the target word in context $c$. Of course, more advanced statistical methods could be also used for learning the sense tensors from the provided partitioning of the contexts, as long as these methods respect the multi-linear nature of the model. This completes the word sense induction step.

\subsection{Creating Density Matrices}

We have now managed to equip each word with a set of sense vectors (or higher-order tensors, depending on its grammatical type). Assigning a probability to each sense is trivial and can be directly derived by the number of times the target word occurs under a specific sense divided by the total occurrences of the word in the training corpus. This creates a statistical ensemble of state vectors and probabilities that can be used for computing a density matrix for the word according to Definition \ref{def:word}.

\end{document}